\documentclass[lettersize,journal]{IEEEtran}
\usepackage{amsmath,amsfonts}
\usepackage{algorithmic}
\usepackage{algorithm}
\usepackage{array}
\usepackage{textcomp}
\usepackage{stfloats}
\usepackage{url}
\usepackage{verbatim}
\usepackage{graphicx}
\usepackage{cite}
\usepackage{multirow}
\usepackage{subfigure}
\usepackage{booktabs}
\usepackage{color} 

\begin{document}

\title{No-Reference Quality Assessment for 3D Colored Point Cloud and Mesh Models}

\author{Zicheng Zhang, Wei Sun, Xiongkuo Min, \emph{Member, IEEE,} Tao Wang, \\ Wei Lu, and Guangtao Zhai, \emph{Senior Member, IEEE} 
\IEEEcompsocitemizethanks{\IEEEcompsocthanksitem Zicheng Zhang, Wei Sun, Xiongkuo Min, Tao Wang, Wei Lu, and Guangtao Zhai are with the Institute of Image Communication and Network Engineering, Shanghai Jiao Tong University, 200240 Shanghai, China. E-mail:\{zzc1998,sunguwei,minxiongkuo,f1603011.wangtao,SJTU-Luwei,zhaiguangtao\}@sjtu.edu.cn.\protect}
\thanks{
Part of the work was presented at \cite{zhang2021}. 

\textsuperscript{1}The code is available at https://github.com/zzc-1998/NR-3DQA.}\\}



\maketitle
\thispagestyle{plain}
\pagestyle{plain}
\begin{abstract}
To improve the viewer's Quality of Experience (QoE) and optimize computer graphics applications, 3D model quality assessment (3D-QA) has become an important task in the multimedia area. Point cloud and mesh are the two most widely used digital representation formats of 3D models, the visual quality of which is quite sensitive to lossy operations like simplification and compression. Therefore, many related studies such as point cloud quality assessment (PCQA) and mesh quality assessment (MQA) have been carried out to measure the visual quality of distorted 3D models. However, most previous studies utilize full-reference (FR) metrics, which indicates they can not predict the quality level in the absence of the reference 3D model. Furthermore, few 3D-QA metrics consider color information, which significantly restricts their effectiveness and scope of application. In this paper, we propose a no-reference (NR) quality assessment metric for colored 3D models represented by both point cloud and mesh. First, we project the 3D models from 3D space into quality-related geometry and color feature domains. Then, the 3D natural scene statistics (3D-NSS) and entropy are utilized to extract quality-aware features. Finally, a support vector regression (SVR) model is employed to regress the quality-aware features into visual quality scores. Our method is validated on the colored point cloud quality assessment database (SJTU-PCQA), the Waterloo point cloud assessment database (WPC), and the colored mesh quality assessment database (CMDM). The experimental results show that the proposed method outperforms most compared NR 3D-QA metrics with competitive computational resources and greatly reduces the performance gap with the state-of-the-art FR 3D-QA metrics. The code of the proposed model is publicly available now to facilitate further research\textsuperscript{1}.  
\end{abstract}



\section{Introduction}\label{sec:introduction}
\IEEEPARstart{N}OWADAYS, with the rapid development of computer graphics, the digital representation of 3D models has been widely studied and used in a wide range of application scenarios such as virtual reality (VR), medical 3D reconstruction, and video post-production \cite{application}, etc. Among the digital representation forms of 3D models, point cloud and mesh are the most widely used formats in practical. A point cloud based object is a set of points in space, in which each point is described with geometry coordinates and sometimes with other attributes such as color and surface normals. Mesh is more complicated because it is a collection of vertices, edges, and faces which together define the shape of a 3D model. Except for geometry information, the 3D mesh may also contain other appearance attributes, such as color and material. Both point cloud and mesh are able to vividly display exquisite models and complex scenes. However, since point cloud and mesh record the omnidirectional details of objects and scenes, lossless 3D models usually need large storage space and very high transmission bandwidth in practical applications. 
Hence, a variety of 3D processing algorithms such as simplification and compression, etc. have been proposed to satisfy the specific needs, which inevitably cause damage to the visual quality of 3D models \cite{li2020occupancy,mekuria2016design}. Additionally, in some 3D scanning model APIs like Apple 3D object capture \cite{apple} and Intel Lidar Camera Realsense \cite{intel}, some slight disturbance such as blur, noise, etc. may be introduced to the constructed 3D models.

Therefore, to improve users' Quality of Experience (QoE) in 3D fields and optimize 3D compression and reconstruction systems, it is of great significance to develop quality metrics for point cloud quality assessment (PCQA) and mesh quality assessment (MQA). However, different from the format of 2D media like images and videos where pixels are distributed in the fixed grid, the points in 3D models are distributed irregularly in the space, which brings a huge challenge for 3D-QA tasks. For example, the neighborhood of pixels in 2D media can be easily obtained, thus the local features are available by analyzing the relationship between the pixel and its neighborhood. However, the neighborhood for points in 3D models is very ambiguous, which makes it difficult to conduct local feature analysis. Furthermore, the geometrical shape of 2D media is fixed, which enables us to easily scale or crop the 2D media for feature extraction while similar operations are very hard to define for 3D models. Although some 3D-QA databases \cite{sjtu-pcqa} \cite{database} \cite{pcqa_database1} \cite{pcqa_database2} \cite{su2021perceptual} have been proposed to push forward the development of 3D-QA algorithms, the difficulty of collecting 3D models and conducting subjective experiments greatly limit the size of the 3D-QA databases, which may also restrict relevant research, especially for NR 3D-QA metrics. 

\subsection{Previous 3D-QA Works}
Quality assessment can be divided into subjective quality assessment and objective quality assessment according to whether human observers are involved. It is known that, in subjective quality assessment, large numbers of people are required to assess the visual quality of 3D models and give their subjective scores. Although the mean opinion scores (MOSs) collected from human observers are straightforward and precise, they cannot be used in practical applications due to the huge time consumption and high expense  \cite{sun2021blind}. Therefore, objective quality metrics are urgently needed to predict the visual quality of 3D models automatically. {The 3D objective quality assessment can then be divided into full-reference (FR), reduced-reference (RR), and no-reference (NR) 3D-QA methods. FR 3D-QA methods work by comparing the difference between the reference 3D models and distorted 3D models, RR 3D-QA methods employ part of the reference models' information for comparison, and NR 3D-QA methods only analyze the distorted 3D models to give the perceptual quality scores.} Considering the complexity of 3D models, in the literature, a large part of 3D-QA metrics are full-reference and they only take geometry features into consideration \cite{p2point, p2plane, m1,ff2_roughness, p2mesh,angular,pcqa2,dame}. When it comes to 3D models with color information, limited works have been proposed \cite{pcqa3,tian-color,guo-color,pcqm,liu2021reduced}. Clearly, the NR 3D-QA for colored models has fallen behind. In this section, we briefly review the development of 3D-QA and introduce the mainstream methods designed for 3D-QA tasks. 

\subsubsection{The Development of PCQA}
The earliest FR-PCQA metrics usually focus on the geometry aspect at the point level, such as p2point \cite{p2point}, p2plane \cite{p2plane}, and p2mesh \cite{p2mesh}. The p2point estimates the levels of distortion by computing the distance vector between the corresponding points. The p2plane further projects the distance vector on the normal orientation for evaluating the quality loss. The p2mesh first reconstructs the point cloud to mesh and then measures the distance from points to the reconstructed surface to predict the quality level, however, it greatly depends on the reconstruction algorithms and lacks stability. Since the point-level difference is difficult to reflect the complex structural distortions, some studies further consider other structural characteristics for PCQA. For example, Alexiou $et$ $al.$ \cite{angular} adopt the angular difference of point normals to estimate the degradations. Javaheri $et$ $al.$ \cite{pcqa2} utilize the generalized Hausdorff distance to reflect the distortions caused by compression operations.

In some situations, color information can not be ignored, which challenges the PCQA methods considering only geometry information. To incorporate the color information into PCQA models, Meynet $et$ $al.$ \cite{pcqm} propose a metric to use the weighted linear combination of curvature and color inf ormation to evaluate the visual quality of distorted point clouds. Inspired by SSIM \cite{ssim}, Alexiou $et$ $al.$ \cite{pcqa3} compute the similarity of four types of features, including geometry, normal vectors, curvature, and color information. What's more, some studies  \cite{sjtu-pcqa} \cite{pcqa_database2} try to predict the quality level by evaluating the projected 2D images from 3D models. The advantage is that image quality assessment (IQA) metrics have been well developed while the disadvantage is that there is inevitable information loss during the projection and the projected images are easily affected by the projected angles and viewpoints.

Few NR-PCQA metrics have been developed so far, which include a learning-based approach \cite{pcqa-large-scale} using two modified PointNet \cite{Qi_2017_CVPR} as feature extraction backbone and also a learning-based method \cite{liu2021pqa} using multi-view projection. 

\subsubsection{The Development of MQA}
The MQA metrics can be categorized into two types: model-based metrics \cite{m1} \cite{ff2_roughness} \cite{dame} \cite{nr-svr} \cite{nr-cnn} which operate directly on the 3D models, and IQA-based metrics \cite{tian-color} \cite{guo-color} \cite{nr-cnncmp} \cite{2d1} \cite{2d2} which operate on the rendering snapshots of 3D models.
Supported by the vast amount of previous research on FR-IQA methods \cite{ms-ssim,fsim,vif}, many FR-MQA models have been proposed using similar manners, which usually compute local features at the vertex level and then pool the features into a quality value. For example, MSDM2 \cite{m1} uses the differences of structure (captured via curvature statistics) computed on local neighborhoods to predict the quality level. DAME \cite{dame} measures the differences in dihedral angles between the reference and the distorted meshes to evaluate the quality loss. FMPD \cite{ff2_roughness} estimates the local roughness difference derived from Gaussian curvature to assess the quality of the distorted mesh. However, these metrics only take geometry information into consideration.

Furthermore, to analyze the influence of color information, some color-involved FR-MQA metrics have been carried out. Tian $et$ $al.$ \cite{tian-color} introduce a global distance over texture image using Mean Squared Error (MSE) to quantify the effect of color information. Guo $et$ $al.$ \cite{guo-color} exploit SSIM to calculate the texture image distance as the color information features. Nehmé $et$ $al.$ \cite{database} introduce a metric to incorporate perceptually relevant curvature-based and color-based features to evaluate the visual quality of colored meshes.

Recently, thanks to the effectiveness of machine learning technologies, some learning-based NR-MQA metrics have been proposed. Abouelaziz $et$ $al.$ \cite{nr-svr} extract features using dihedral angle models and train a support vector machine for feature regression. Later, Abouelaziz $et$ $al.$ \cite{nr-cnn} scale the curvature and dihedral angle into 2D patches and utilize the convolution neural network (CNN) for training. They further introduce a CNN framework with saliency views rendered from 3D meshes \cite{nr-cnncmp}. However, the NR methods mentioned above only extract geometry features and may fail to accurately predict the scores of colored meshes.

\subsection{Our Approach}
Generally speaking, a large part of the metrics mentioned above are full-reference metrics, which can make full use of the relationship between the reference and distorted 3D models and give relatively accurate results. However, the disadvantage of FR 3D-QA metrics is also obvious. They are not able to work in the absence of reference 3D models. Unfortunately, in many application scenarios like 3D reconstruction, pristine reference is not always available.  {Therefore, inspired by the huge success of natural scene statistics (NSS) in IQA tasks \cite{brisque,nss1,nss2}, some researchers \cite{lin2019blind,abouelaziz2018blind,nr-svr} use similar ideas of extracting handcrafted features and estimating the statistical parameters with certain NSS distributions as quality-aware information. In this paper, we further push forward the application of NSS for both colored point cloud and mesh models, and we summarize such a concept as 3D-NSS.} Specifically, NSS is a discipline within the field of perception, which is dependent on the premise that the perceptual system is designed to interpret natural scenes \cite{nss2}. Through observations of feature distributions, we find that the reference features obey certain NSS distributions and different types of distortions change the appearance of such feature distributions. Therefore, we believe that 3D-NSS is effective to quantify the visual quality of 3D models in the presence of distortion. More details of 3D-NSS are discussed in Section \ref{sec:estimate nss parameters}.

Specifically, we first project the 3D models into quality-related geometry and color feature domains. Then, 3D-NSS and entropy are utilized to extract these color and geometry characteristics. Finally, the obtained features are integrated into a quality value through the support vector regression (SVR) model. In order to test the effectiveness of different types of features and different kinds of distribution models, we test the performance of various combinations of the statistical parameters to find the optimal combination. Further, we conduct the data-sensitivity experiment, the ablation study, and the computational efficiency experiment to demonstrate the effectiveness of our method. In-depth discussions are given as well.   

\begin{figure*}[t]
    \centering
    \includegraphics[width = 14cm]{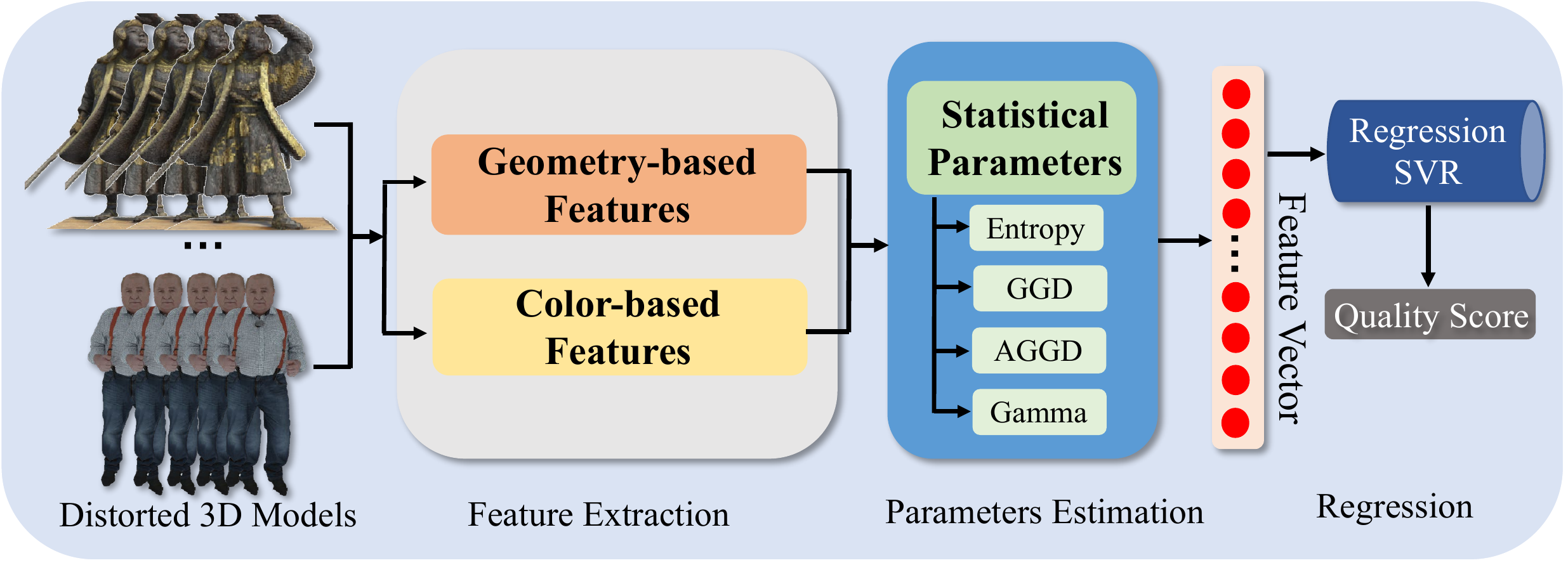}
    \caption{The framework of the proposed method. Geometry-based and color-based features are first extracted from the distorted 3D models. Then various statistical parameters are estimated from the extracted features to form the feature vector. Finally, the quality scores are given through the SVR regression module.}
    \label{fig:framework}
    \vspace{-0.5cm}
\end{figure*}


\subsection{Contributions of This Paper}
We summarize our contributions as follows:
\begin{itemize}
    \item {We push forward the development of NSS in the 3D-QA fields based on previous research \cite{lin2019blind,abouelaziz2018blind,nr-svr} and we systematically summarize the concept as 3D-NSS.} 
    \item We propose a no-reference quality assessment metric for both colored point cloud and mesh. We extract features not only from the geometry aspect, but also from the color aspect. Furthermore, it is the first method that can deal with both NR-PCQA and NR-MQA with color information. The proposed NR 3D-QA model follows a common NR framework, which means it is easy to modify our metric for performance improvement or meeting other specific needs.
    \item We deeply investigate the effectiveness of different types of features and different kinds of NSS models, which can provide useful guidelines for future research.
    \item Compared with the state-of-the-art methods, our method is more efficient in the computation process, which means the proposed method is potentially more capable of handling practical situations. The code of the proposed model is also released for promoting the development of NR 3D-QA.
\end{itemize}

The paper is organized as follows. Section \ref{sec:feature projection} describes the feature projection processes. Section \ref{sec:estimate nss parameters} describes the process of quantifying the distortions to statistical parameters. Section \ref{sec:experiment} presents the experiment setup and the experimental results. Section \ref{sec:conclusion} summarizes of this paper.

\section{Feature Projection}
\label{sec:feature projection}
The framework of the proposed method is illustrated in Fig. \ref{fig:framework}, which includes a feature extraction module, a parameters estimation module, and a regression module. Point cloud and mesh have similar quality-aware geometric properties as well as the attached color attributes. Therefore, we uniformly design the point cloud and mesh processing framework, and determine the specific projection domains based on their characteristics.
Before introducing the proposed model, we first define a distorted 3D object O as:
\begin{equation}
\begin{aligned}
& O \in \{P, M\},\\
& P=\{Points\},\\
& M=\{Vertices, Edges, Faces\},
\end{aligned}
\end{equation}
where $P$ and $M$ mean that the 3D object is represented by point cloud and mesh respectively, and the color information is attached to $Points$ in point cloud and $Vertices$ in mesh respectively.

\subsection{Geometry Feature Projection}
Geometry features usually have a strong correlation with human perception, which has been firmly proved in 3D-QA studies \cite{p2point, p2plane, m1,ff2_roughness, p2mesh,angular,pcqa2,dame}. Although the geometry features are computed in different ways for point cloud and mesh, they share similar characteristics for the visual quality of 3D models. In this section, we project the given 3D model into several quality-aware geometry feature domains:
\begin{equation}
\begin{aligned}
    F_{geo} & = \boldsymbol{\rm{Projection_{geo}}}(O),\\
    O & \in \{P, M\},
\end{aligned}
\end{equation}
where $F_{geo}$ indicates the set of geometry feature domains of the 3D model, $\boldsymbol{\rm{Projection_{geo}}}(\cdot)$ denotes the geometry projection function.

\subsubsection{  Point Cloud Geometry Feature Domains}
Considering that the point cloud lacks surfaces, we first need to get the neighborhood set for each point so we can further extract the geometry features. Given the point cloud $P=\left\{p_{i}\right\}_{i=1}^{N}$ , the neighborhood $P_{Nbi}$ of each vertex $p_{i}$ can be obtained utilizing the k-nearest neighbors (k-NN) algorithm:
\begin{equation}
\begin{aligned}
    & \quad \quad P_{Nb} = \mathop{\boldsymbol{\rm{KNN}}}\limits_{k=10}(P),\\
   Dist(p,q)\! &=\! \sqrt{\left(p_{x}\!-\!q_{x}\right)^{2}\!+\!\left(p_{y}\!-\!q_{y}\right)^{2}\!+\!\left(p_{z}\!-\!q_{z}\right)^{2}},
\end{aligned}
\end{equation}  
where $N$ indicates the number of points in the point cloud, $P_{Nb}$ is the set of all neighborhoods, $\boldsymbol{\rm{KNN}}(\cdot)$ denotes the k-NN algorithm function, the $k$ level is set as 10, and Euclidean distance $Dist(p,q)$ is exploited as the distance function. With the computed neighborhood $P_{Nbi}$, the corresponding covariance matrix $C_{i}$ for each point $p_{i}$ (represented by 3D geometry coordinates) can be denoted as:
\begin{equation}
\begin{aligned}
     C_{i} = &  \frac{1}{K} \sum_{j=1}^{K} ({p}_{n_{j}}-\hat{p})({p}_{n_{j}}-\hat{p})^{\top},\\
    &  \{{p}_{n_{1}}, \cdots ,{p_{n_{K}}} \} \in P_{Nbi},
\end{aligned}
\end{equation}
{where ${p}_{n_{j}}$ and $\hat{p}$ are vectors of dimension 3$\times$1, $C_{i}$ is a matrix of dimension 3$\times$3, $K$ stands for the size of neighborhood $P_{Nbi}$, and $\hat{p}$ represents the centroid of neighborhood $P_{Nbi}$.} Therefore, the eigenvector problem for the covariance matrix $C_{i}$ can be described as:
\begin{equation}
C_{i} \cdot {v}_{l}=\lambda_{l} \cdot {v}_{l}, l \in\{1,2,3\},
\end{equation}
where ($\lambda_{1}$, $\lambda_{2}$, $\lambda_{3}$) indicate the eigenvalues, (${v}_{1}$, ${v}_{2}$, ${v}_{3}$  represent the corresponding eigenvectors, and we assume that $\lambda_{1}>\lambda_{2}>\lambda_{3}$). Thus, a total of three eigenvalues are obtained for each point $p_{i}$ in the point cloud $P$.

In previous studies \cite{pcqa-curvature2} \cite{pc-eigenvalue1}, the eigenvalues mentioned above are used to calculate various geometry features for tasks like classification, simplification, and segmentation. These features have shown great ability to describe the complicated geometry structure information and achieved outstanding performance. Hence, we selected several geometry features that may be helpful for point cloud quality assessment: 
\begin{itemize}
    \item Curvature: Curvature is the amount by which a curve deviates from being a straight line, which is frequently used to describe roughness or smoothness.
    \item Anisotropy: Anisotropy is used to exhibit variations in geometrical properties for different directions.
    \item Linearity: Linearity is the property for estimating the similarity to a straight line.
    \item Planarity: Planarity is utilized to measure the similarity to a planar surface.
    \item Sphericity: Sphericity is the measure of how closely the shape of an object resembles that of a perfect sphere. 
\end{itemize}

Moreover, all the geometry features described above represent the local distribution patterns and are calculated at the point level, which indicates that each $p_{i}$ has its own curvature, anisotropy, linearity, planarity, and sphericity.
The formulations \cite{pc-eigenvalue2} for geometry features are denoted as:
\begin{align}
     \boldsymbol{Cur(p_{i})}&= \frac{\lambda_{3}}{\lambda_{1}+\lambda_{2}+\lambda_{3}}, \\
     \boldsymbol{Ani(p_{i})}&= \frac{\lambda_{1}-\lambda_{3}}{\lambda_{1}}, \\
    \boldsymbol{Lin(p_{i})}&= \frac{\lambda_{1}-\lambda_{2}}{\lambda_{1}}, \\
    \boldsymbol{Pla(p_{i})}&= \frac{\lambda_{2}-\lambda_{3}}{\lambda_{1}}, \\
    \boldsymbol{Sph(p_{i})}&= \frac{\lambda_{3}}{\lambda_{1}}, 
\end{align}
where $\lambda_{1}$, $\lambda_{2}$, and $\lambda_{3}$ refer to the corresponding eigenvalues respectively. Through the operations mentioned above, the point cloud is transformed into 5 geometry feature domains.

\begin{figure}[t]
    \centering
    \includegraphics[height = 3cm]{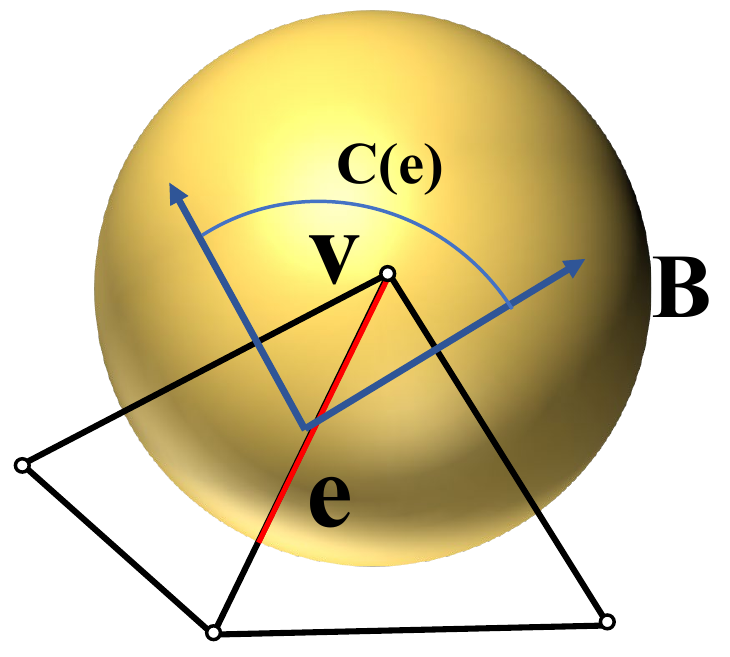}
    \caption{Example of the variables corresponding to the curvature calculation, where the red line indicates $|e \cap B|$ and $C(e)$ indicates the curvature contribution of $e$.}
    \label{fig:variables}
\end{figure}

\subsubsection{ Mesh Geometry Feature Domains}
\label{sec:mqa}
Compared with point cloud, the 3D mesh is more complicated because it is a collection of vertices, edges, and faces which together define the shape of a 3D model. Therefore, the feature extraction process is a bit different from that in the point cloud.

2-1) \textit{Curvature:}
Thanks to the edges and faces information contained in mesh, the computation of curvature is more precise than the point cloud. So far, various curvature definitions have been introduced for 3D meshes \cite{gaussian_curvature} \cite{average_curvature}, among which the average curvature is proven to be able to stably describe the local structural information of 3D meshes. Hence, a weighted average curvature method introduced in \cite{average_curvature} is adopted to measure the roughness of embedded surfaces in 3D meshes. Given the mesh $M=\left\{V,E,F\right\}$, the weighted average curvature is calculated by averaging the curvature contributions over a certain region $B$ around each vertex $v_{i}$:
\begin{equation}
Cur(v_{i})=\frac{1}{|B|} \sum_{e \in E} C(e)|e \cap B|,
\end{equation}
where $Cur(v_{i})$ represents the weighted average curvature for vertex $v$, region $B$ is a sphere region centered at $v$ and its radius is 1/100 of the bounding box of the 3D mesh, $|B|$ stands for the surface area of region $B$. $E$ is the set of edges that are linked to the vertex $v$, $C(e)$ indicates the curvature contribution of $e$ (the signed angle between the normals of the two oriented triangles incident to edge $e$), $|e \cap B|$ denotes the weight which is defined as the length of $e \cap B$ (always between 0 and $|e|$). An example of the variables corresponding to the curvature calculation is shown in Fig. \ref{fig:variables}.

2-2) \textit{Dihedral Angle:}
For a 3D mesh, the dihedral angle is the angle between the normals of two adjacent faces, which has been utilized in \cite{dihedral2} as an effective indicator for measuring the loss of the mesh simplification. Furthermore, \cite{dame} exploits oriented dihedral angles instead of the simple dot product of normals to better distinguish the convex and concave angles. Inspired by similar ideas, we adopt the dihedral angles set to describe the visual quality of 3D meshes. Assuming that the vertices for two adjacent faces $f_{1}$ and $f_{2}$ are \{$v1,v3,v4$\} and \{$v2,v3,v4$\}, we have:
\begin{equation}
Dih_{f_{1}, f_{2}}=\arccos \left(n_{1} \cdot n_{2}\right) \cdot \operatorname{sgn}\left(n_{1} \cdot\left(v_{2}-v_{1}\right)\right),
\end{equation}
where $Dih_{f_{1}, f_{2}}$ is the oriented dihedral angle between two adjacent faces $f_{1}$ and $f_{2}$, $n_{1}$ and $n_{2}$ are the normals of $f_{1}$ and $f_{2}$, $\operatorname{sgn}(\cdot)$ denotes the signum function which is used to decide the orientation of the dihedral angle. 

2-3) \textit{Face Area and Angle:}
Area and angle are two simple attributes of 3D mesh faces, which can be easily computed by making use of the coordinates of the face's vertices. In the mesh smoothing algorithm proposed in \cite{angle1}, attributes including face angle are used to predict the new location of the smoothed nodes. While in the 3D mesh encoding method introduced in \cite{angle2}, face angle is utilized to instruct the compression of 3D meshes, which indicates that face angle is related to the quality of 3D meshes. Therefore, to further measure the visual quality degradation of 3D meshes, the face area and angle are collected as feature sets. 

Finally, the mesh is projected into 4 geometry feature domains:
\begin{itemize}
    \item Curvature: The weighted average curvature is used to describe the geometry characteristics like roughness or smoothness.
    \item Dihedral Angle: Dihedral angle is employed as a useful descriptor for measuring the caused degradations.
    \item Face Area \& Angle: These two attributes are highly correlated with the effectiveness of lossy operations like compression.
\end{itemize}

\subsection{Color Feature Projection}
\label{pc:color}
Color is a significant aspect of visual quality assessment. For a colored point cloud, the color is directly determined by the color information of the point, while for a colored mesh, the color of the surface is generally rendered by the color information of the contained vertices. Besides, the color information in the 3D models is usually stored in the form of RGB channels. However, the RGB color space has been proven to have a poor correlation with human perception.
Therefore, we adopt the LAB color transformation as the color feature projection:
\begin{equation}
 \begin{aligned}
    F_{col} &= \boldsymbol{\rm{Projection_{col}}}(O), \\
    O & \in \{P, M\},
\end{aligned}   
\end{equation}
where $F_{col}$ indicates the set of color feature domains of the 3D model, $\boldsymbol{\rm{Projection_{col}}}(\cdot)$ stands for the color projection function, and $P$ and $M$ represent the distorted point cloud and mesh respectively. The detailed color transformation is formulated as:
\begin{equation}
\left[\begin{array}{c}
X \\
Y \\
Z
\end{array}\right]=\left[\begin{array}{ccc}
2.7688 & 1.7517 & 1.1301 \\
1.0000 & 4.5906 & 0.0601 \\
0 & 0.0565 & 5.5942
\end{array}\right]\left[\begin{array}{c}
R \\
G \\
B
\end{array}\right],
\end{equation}
\begin{equation}
\left\{
\begin{array}{lll}
L &=116 f\left(\frac{Y}{Y_{\mathrm{n}}}\right)-16, \\
A &=500\left(f\left(\frac{X}{X_{\mathrm{n}}}\right)-f\left(\frac{Y}{Y_{\mathrm{n}}}\right)\right), \\
B &=200\left(f\left(\frac{Y}{Y_{\mathrm{n}}}\right)-f\left(\frac{Z}{Z_{\mathrm{n}}}\right)\right),
\end{array}
\right.
\end{equation}
where $R,G,B$ represent the corresponding RGB color channels, $X,Y,Z$ stand for the corresponding XYZ color channels, $L,A,B$ denote the corresponding RGB color channels, and $X_{n},Y_{n},Z_{n}$ describe the specified white achromatic reference illuminant. The $f(\cdot)$ function is described as:
\begin{equation}
f(t)=\left\{\begin{array}{ll}
\sqrt[3]{t}, & \text { if } t>\delta^{3}, \\
\frac{t}{3 \delta^{2}}+\frac{4}{29}, & \text { otherwise },
\end{array}\right.
\end{equation}
where $\delta$ is set as $\frac{6}{29}$. Finally, the LAB color channels are computed as the color feature domains.

\begin{figure*}[!htp]
  \centering
  \subfigure[]{\includegraphics[width = 3.4cm]{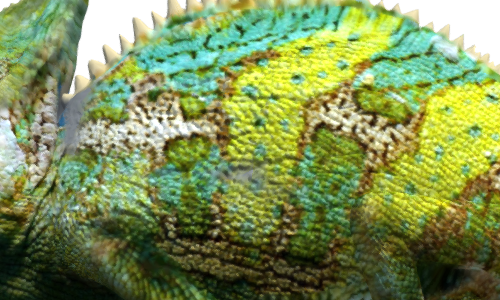}}
  \subfigure[]{\includegraphics[width = 3.4cm]{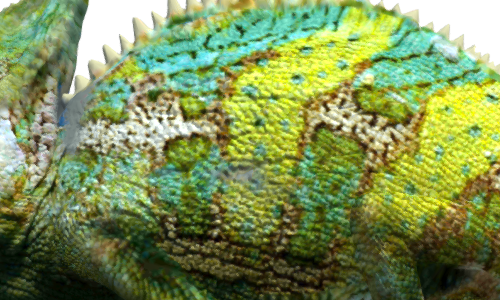}}
  \subfigure[]{\includegraphics[width = 3.4cm]{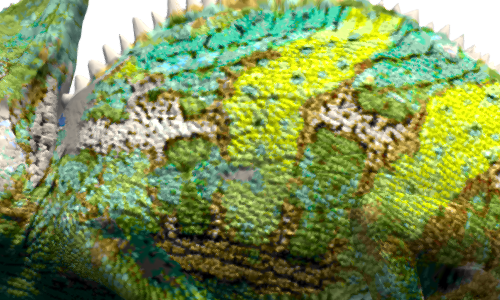}}
  \subfigure[]{\includegraphics[width = 3.4cm]{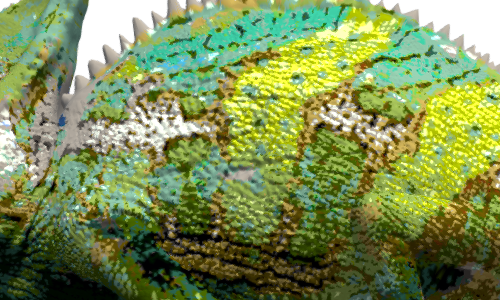}}
  \subfigure[]{\includegraphics[width = 3.4cm]{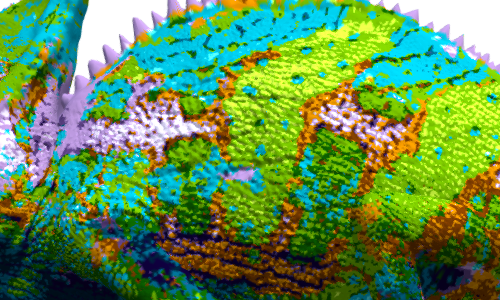}}
  \subfigure[]{\includegraphics[width = 3.4cm,height = 2.65cm]{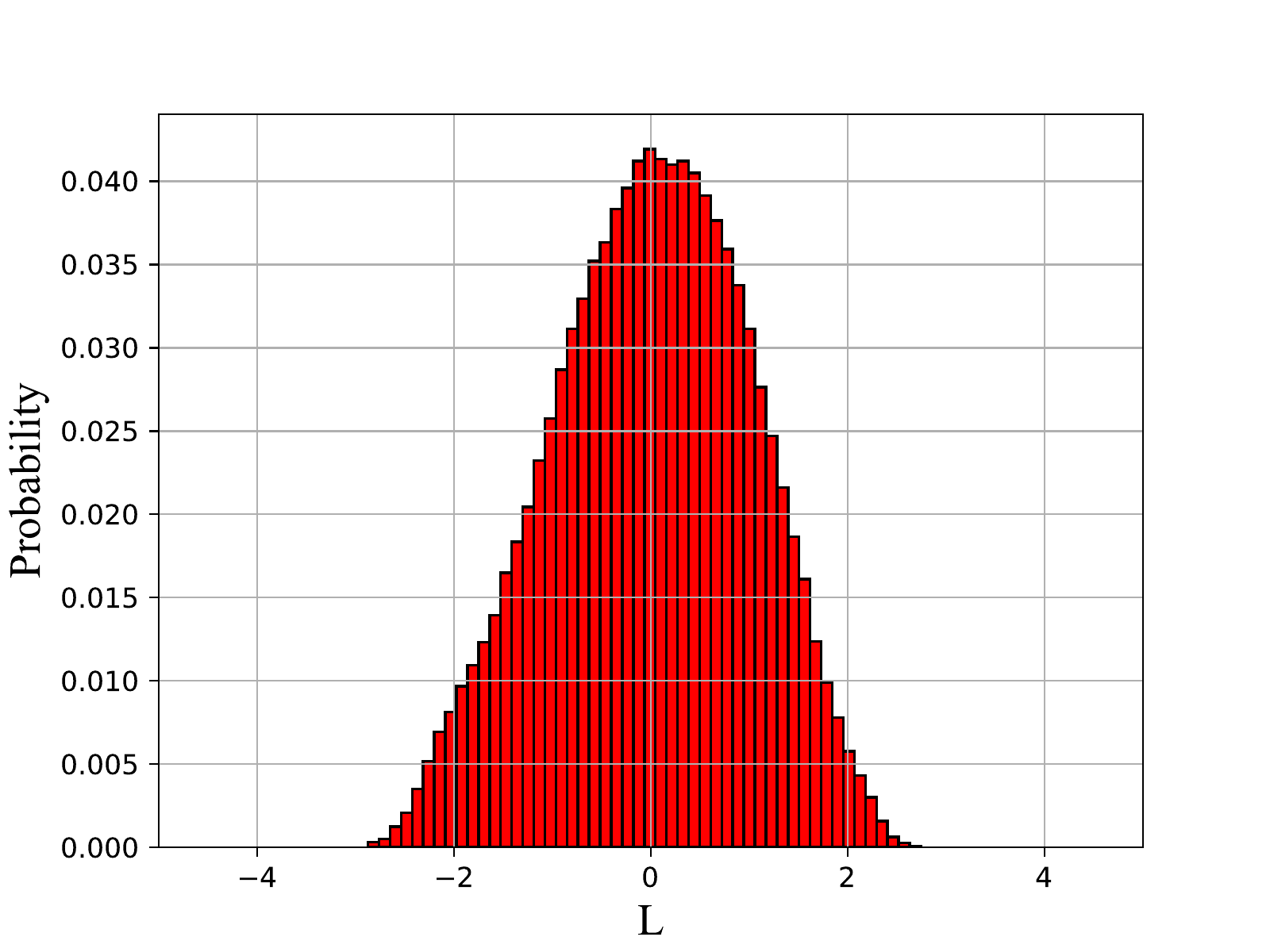}}
  \subfigure[]{\includegraphics[width = 3.4cm,height = 2.65cm]{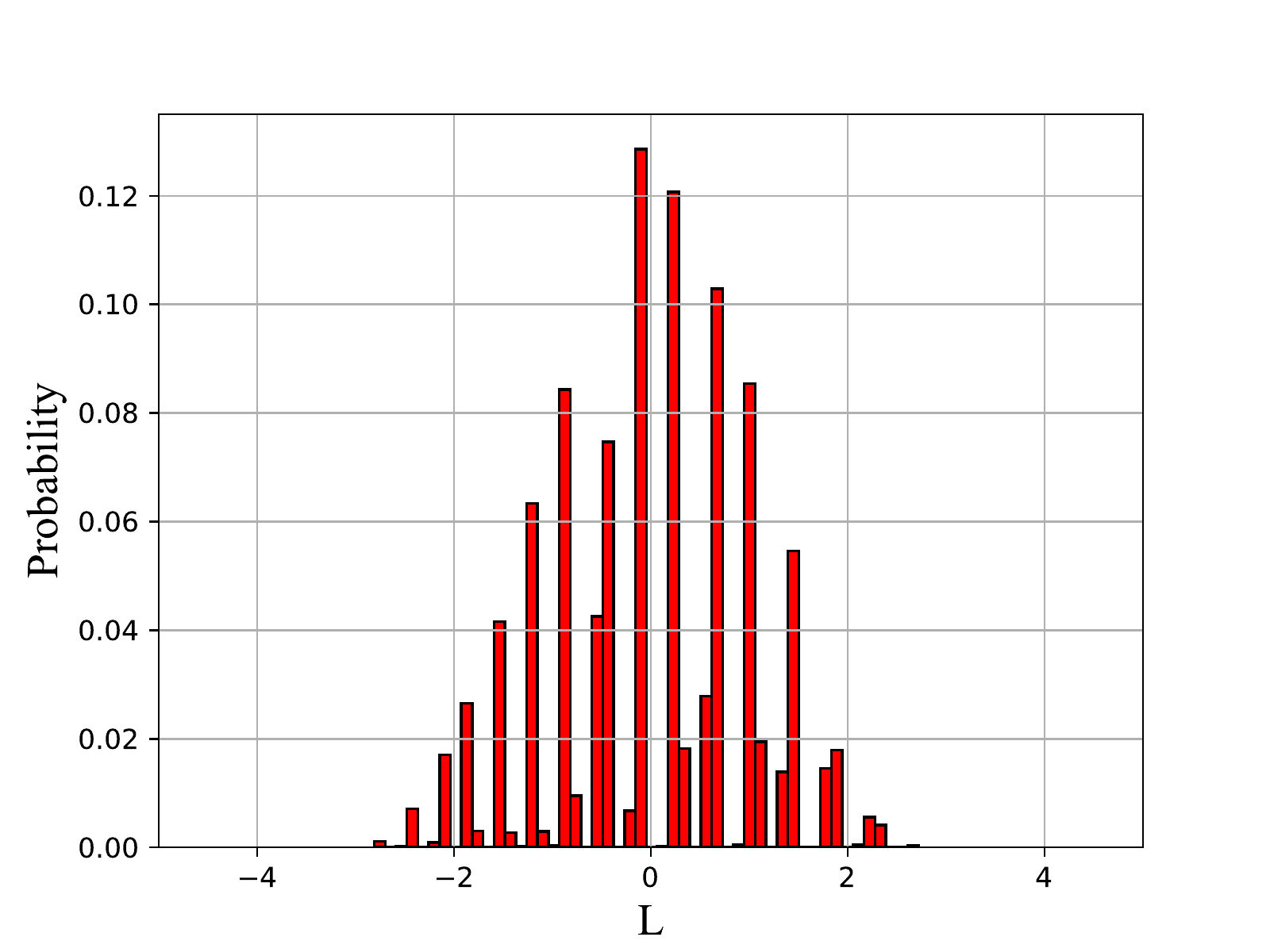}}
  \subfigure[]{\includegraphics[width = 3.4cm,height = 2.65cm]{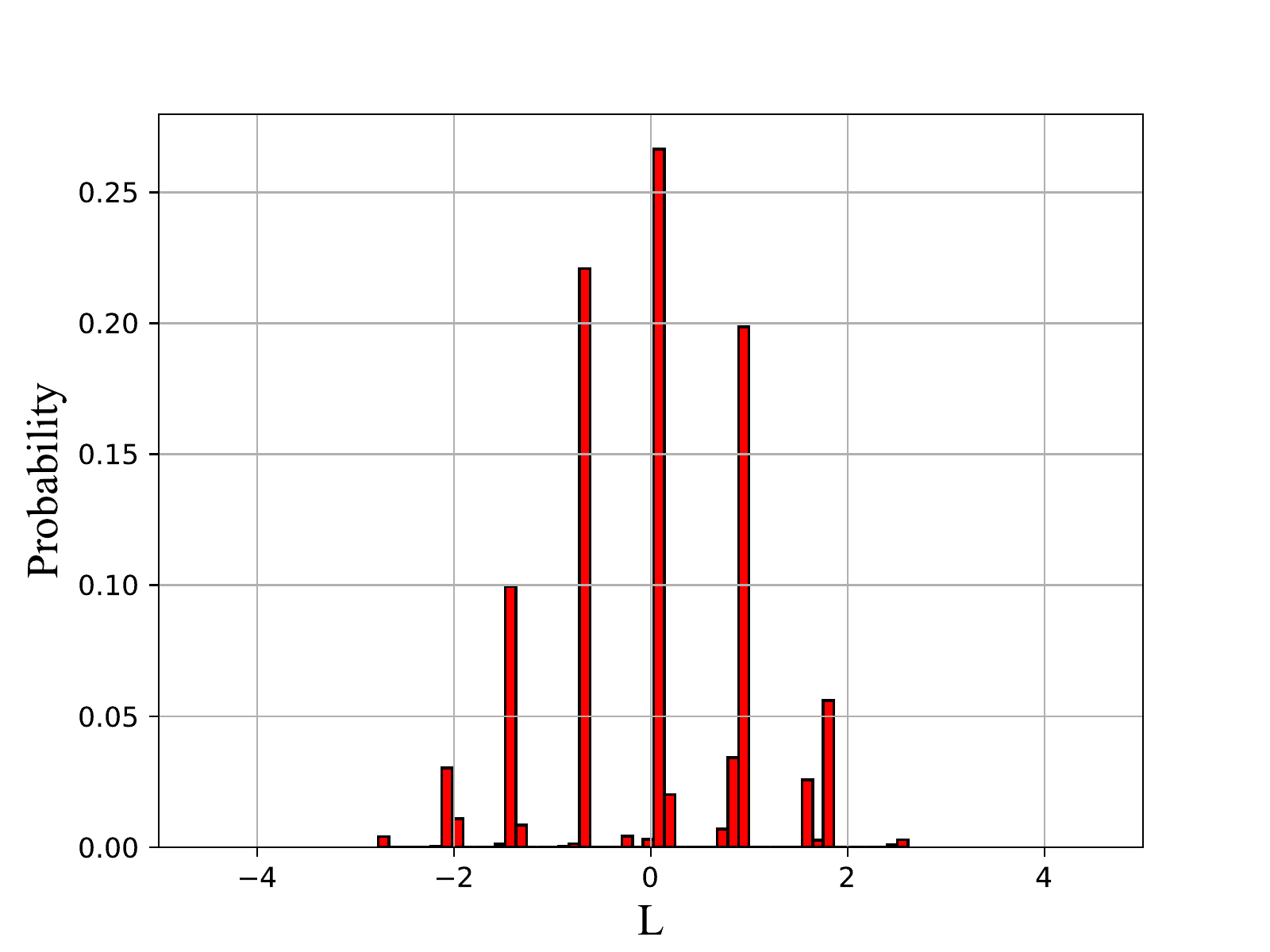}}
  \subfigure[]{\includegraphics[width = 3.4cm,height = 2.65cm]{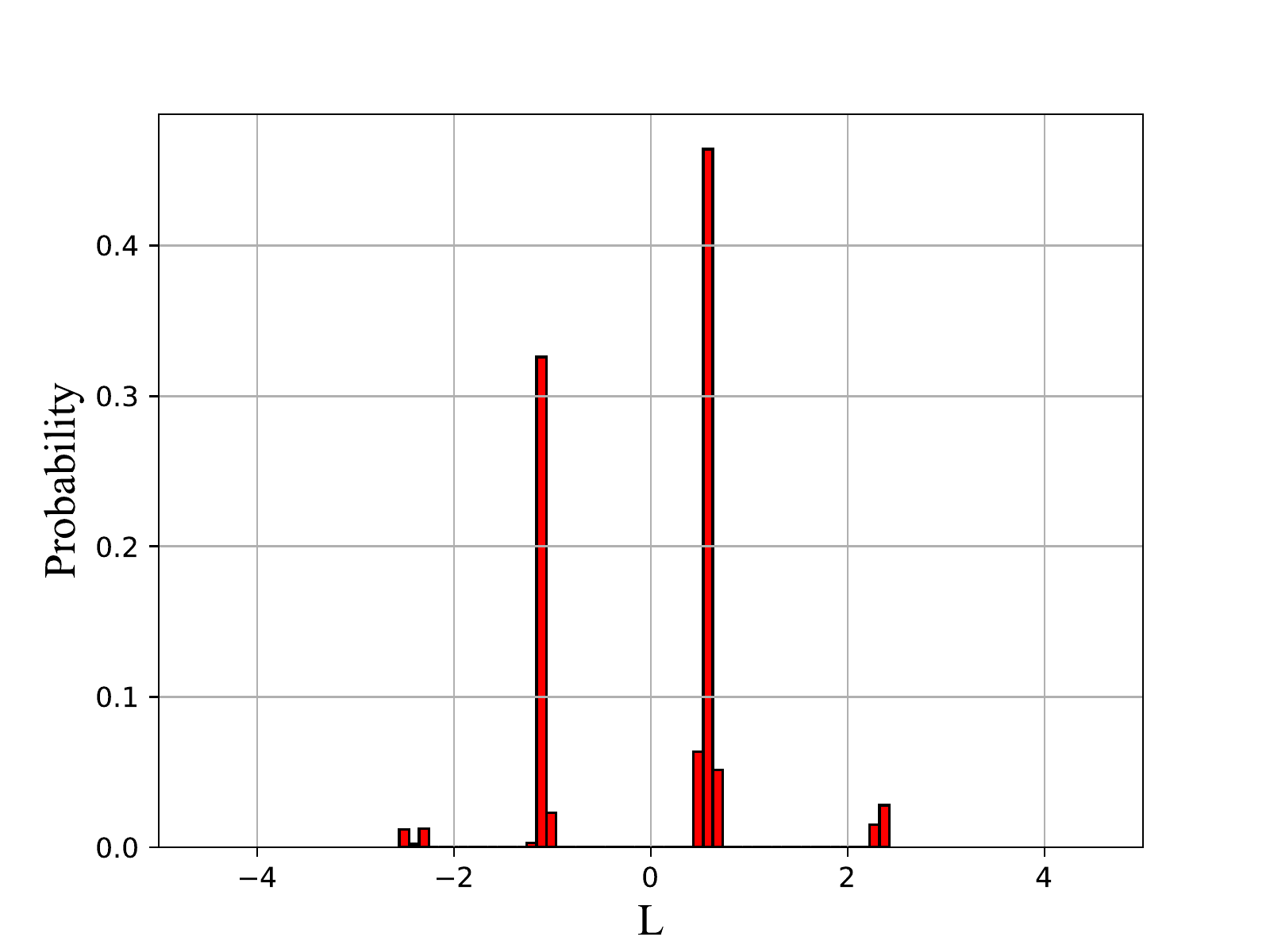}}
  \subfigure[]{\includegraphics[width = 3.4cm,height = 2.65cm]{Chameleon_QuantLAB_3L.pdf}}
  \subfigure[]{\includegraphics[width = 3.4cm,height = 2.65cm]{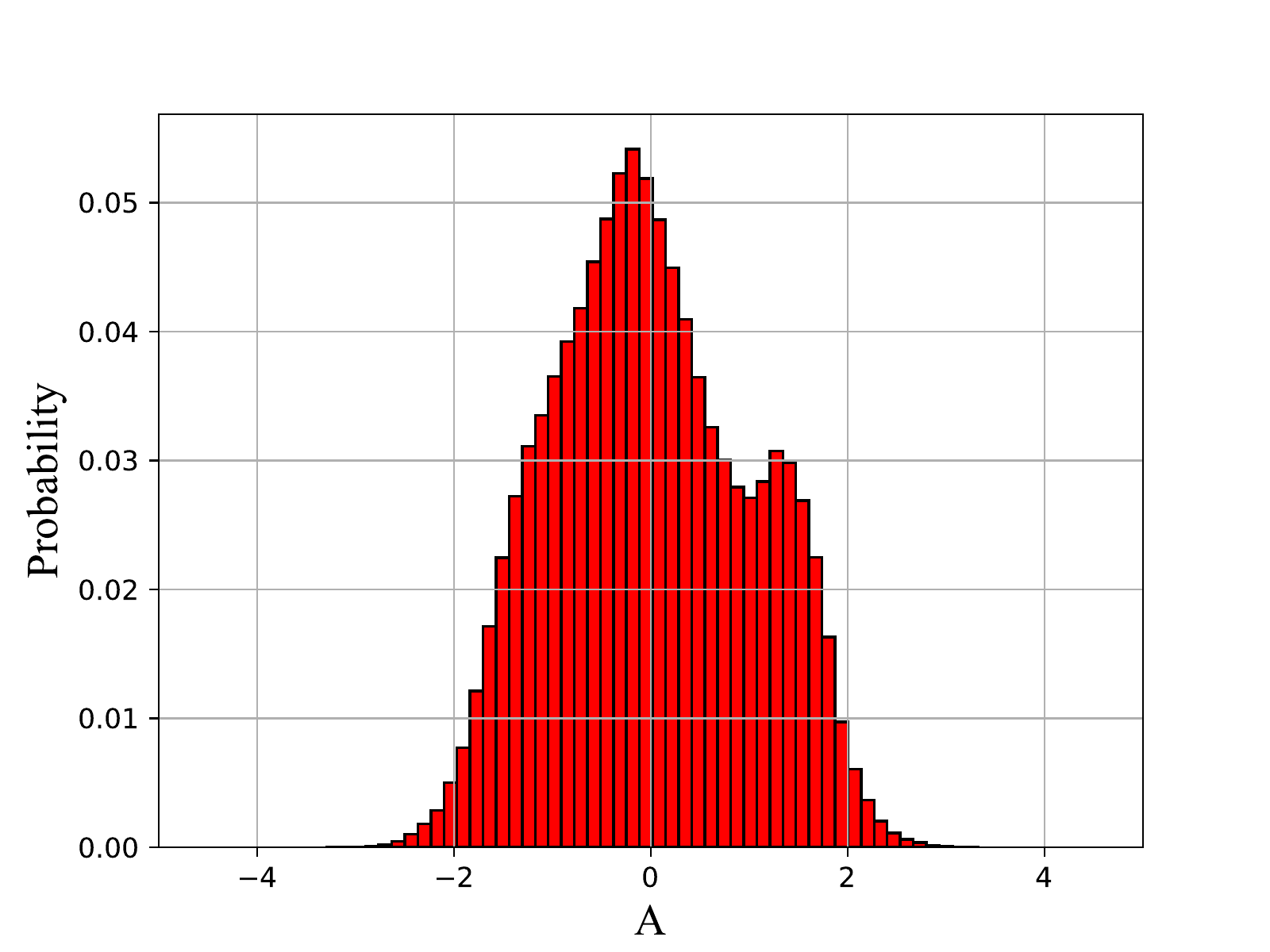}}
  \subfigure[]{\includegraphics[width = 3.4cm,height = 2.65cm]{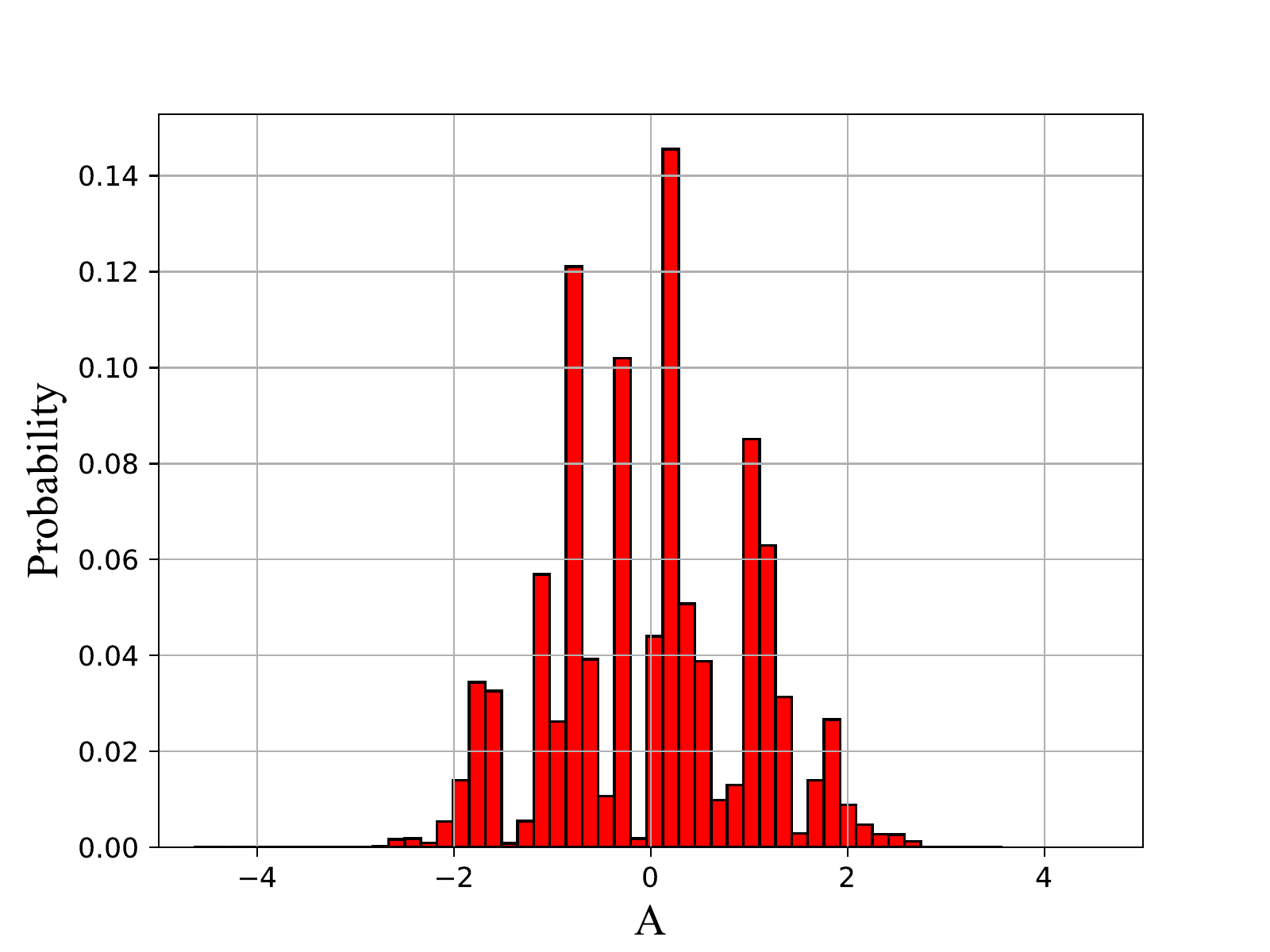}}
  \subfigure[]{\includegraphics[width = 3.4cm,height = 2.65cm]{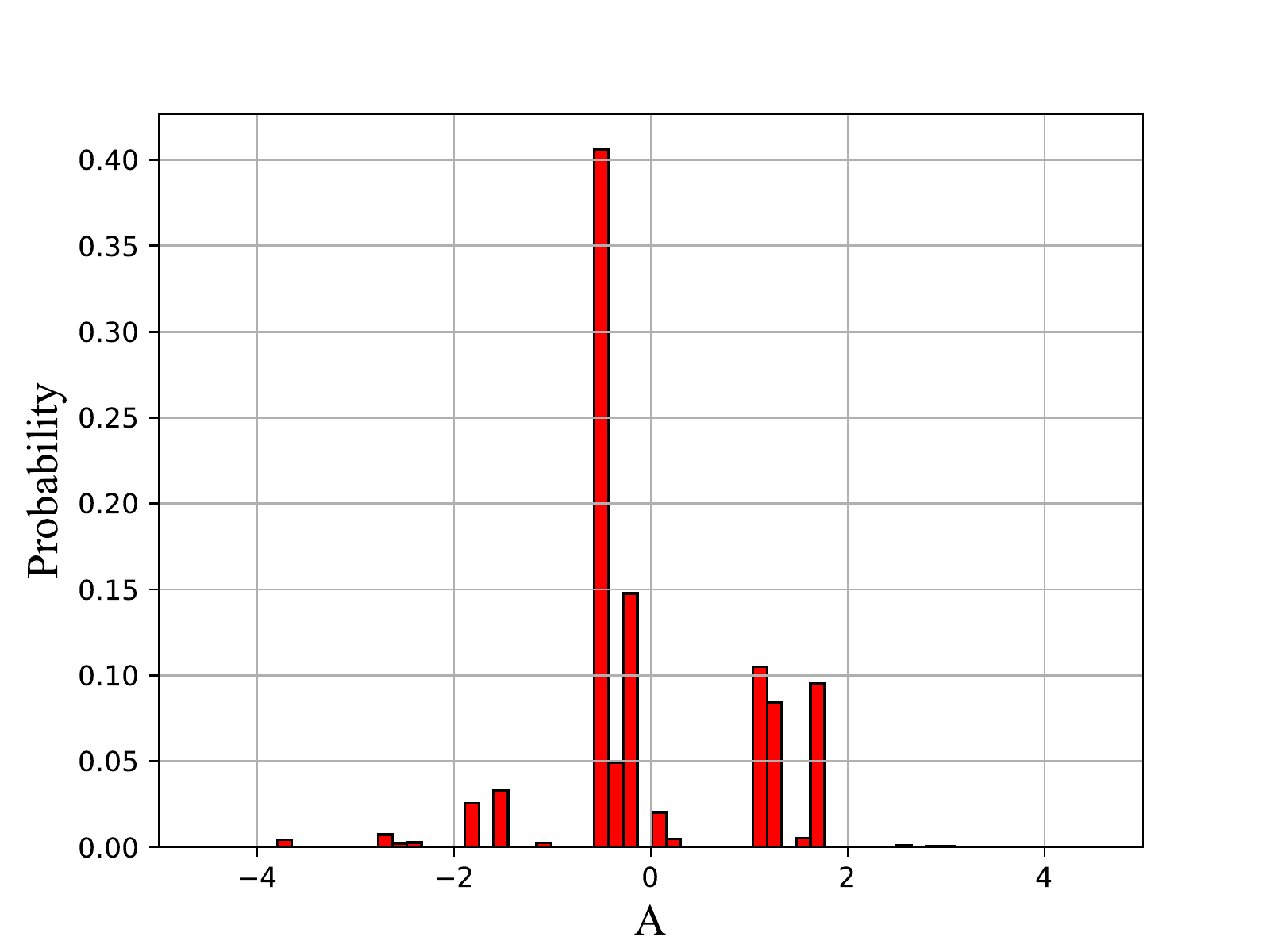}}
  \subfigure[]{\includegraphics[width = 3.4cm,height = 2.65cm]{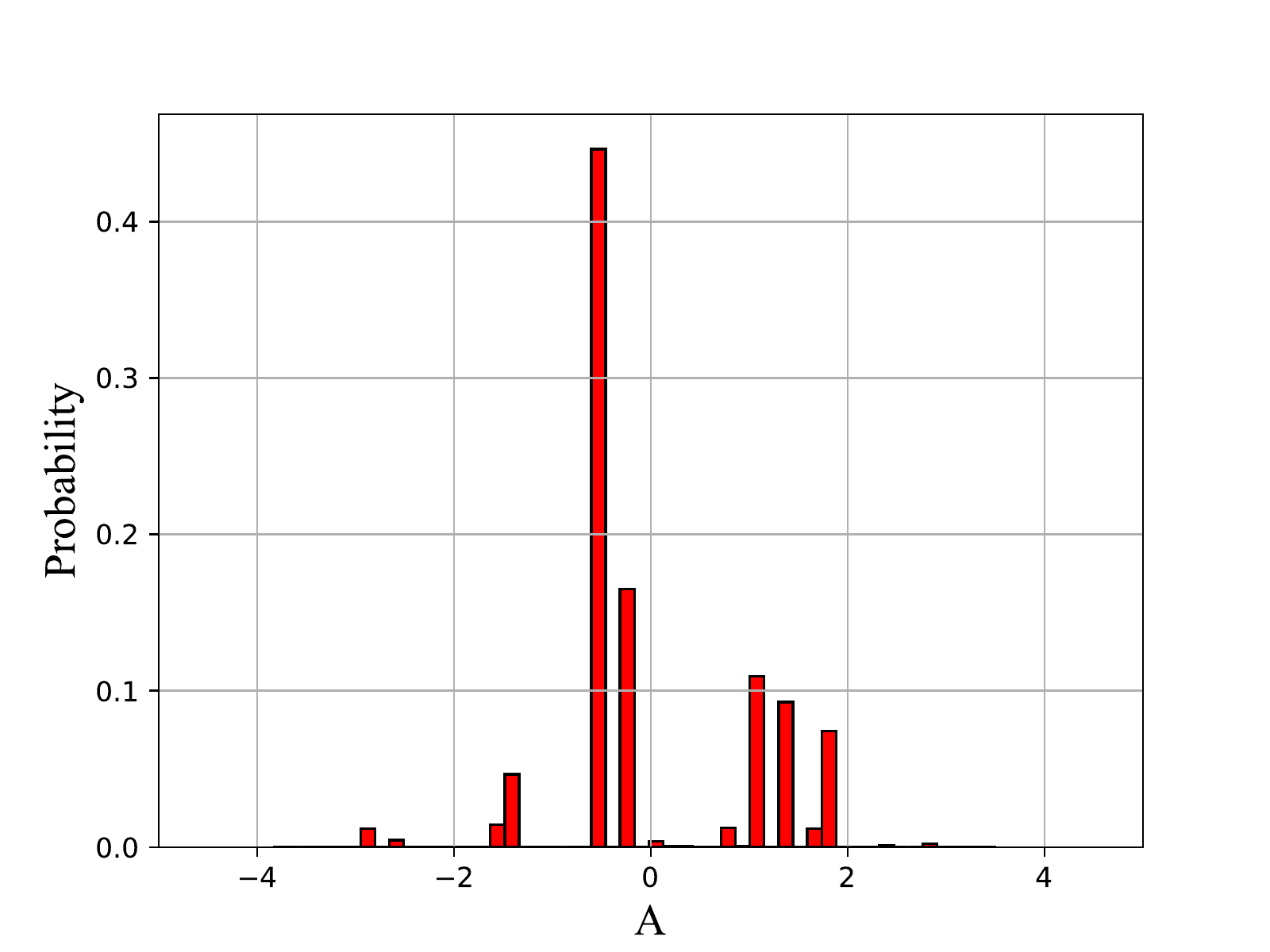}}
  \subfigure[]{\includegraphics[width = 3.4cm,height = 2.65cm]{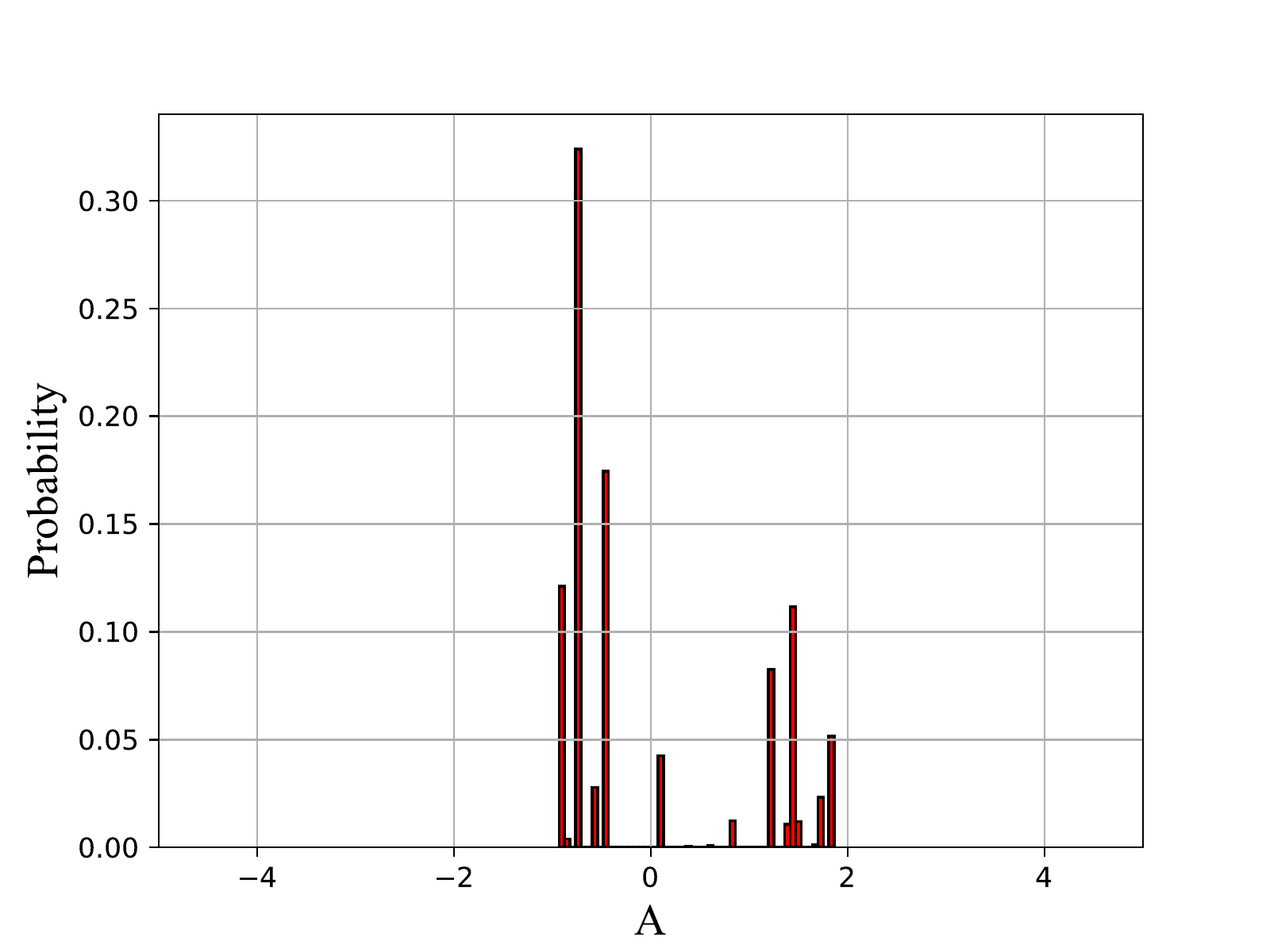}}
  \subfigure[]{\includegraphics[width = 3.4cm,height = 2.65cm]{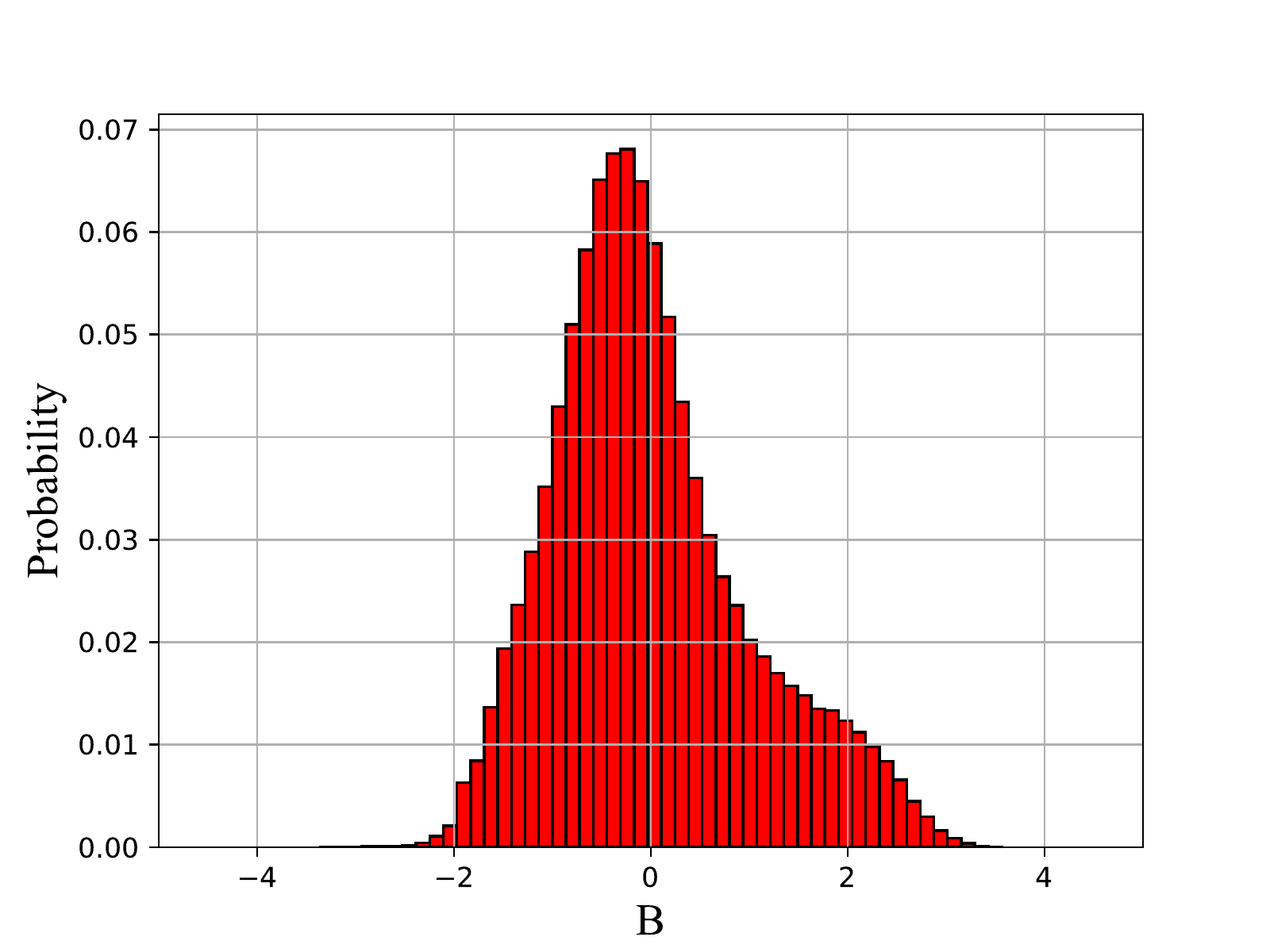}}
  \subfigure[]{\includegraphics[width = 3.4cm,height = 2.65cm]{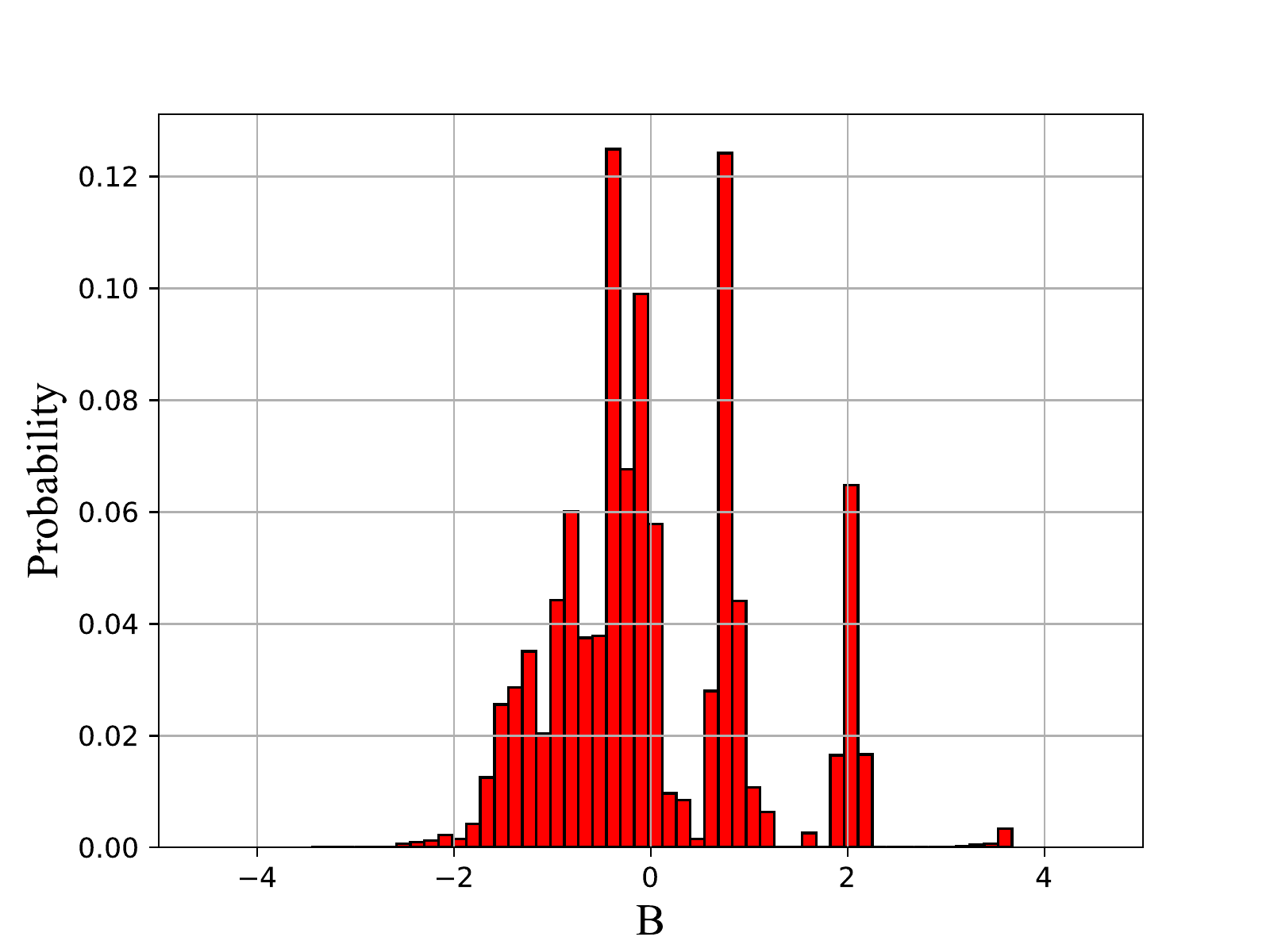}}
  \subfigure[]{\includegraphics[width = 3.4cm,height = 2.65cm]{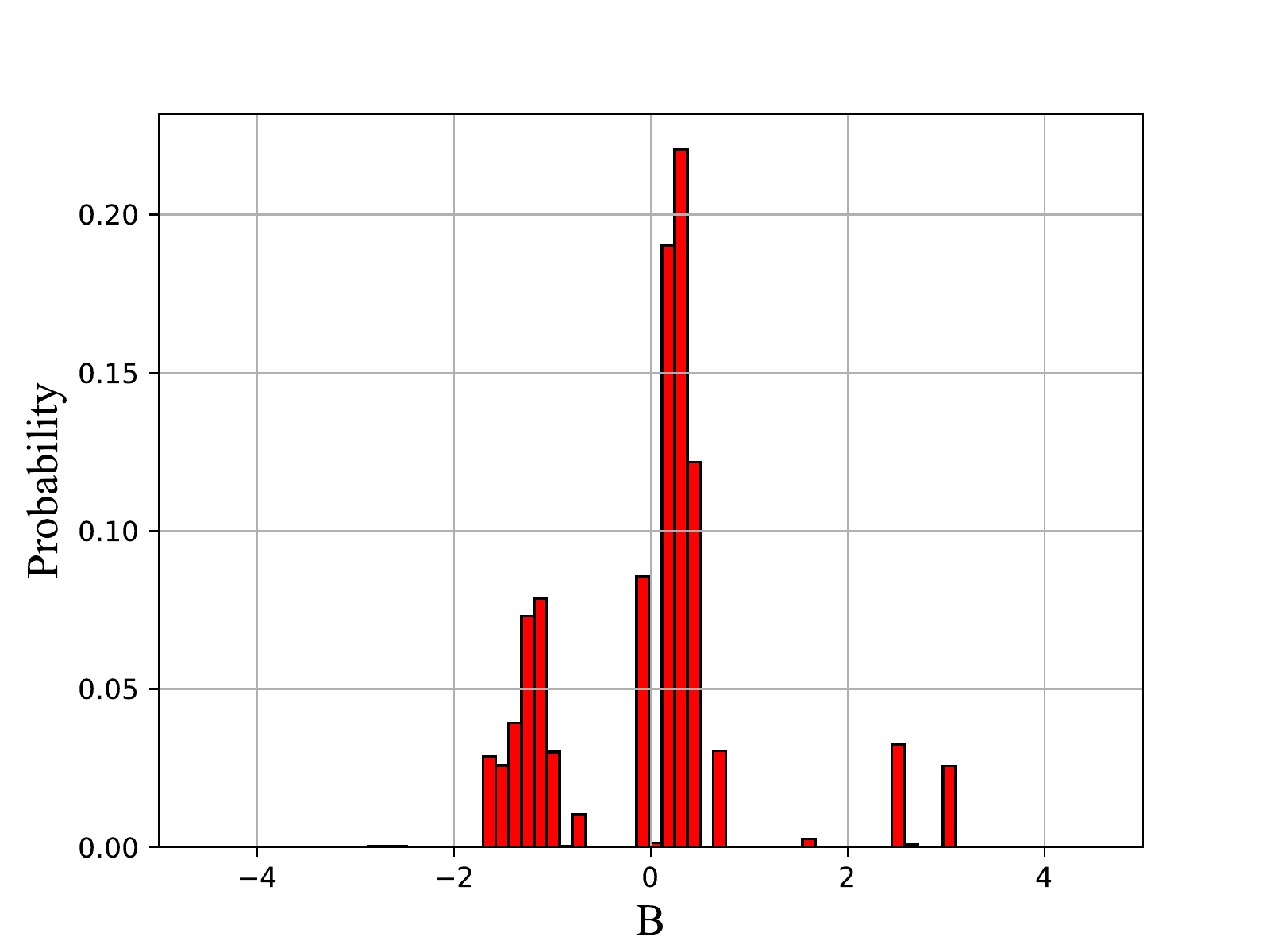}}
  \subfigure[]{\includegraphics[width = 3.4cm,height = 2.65cm]{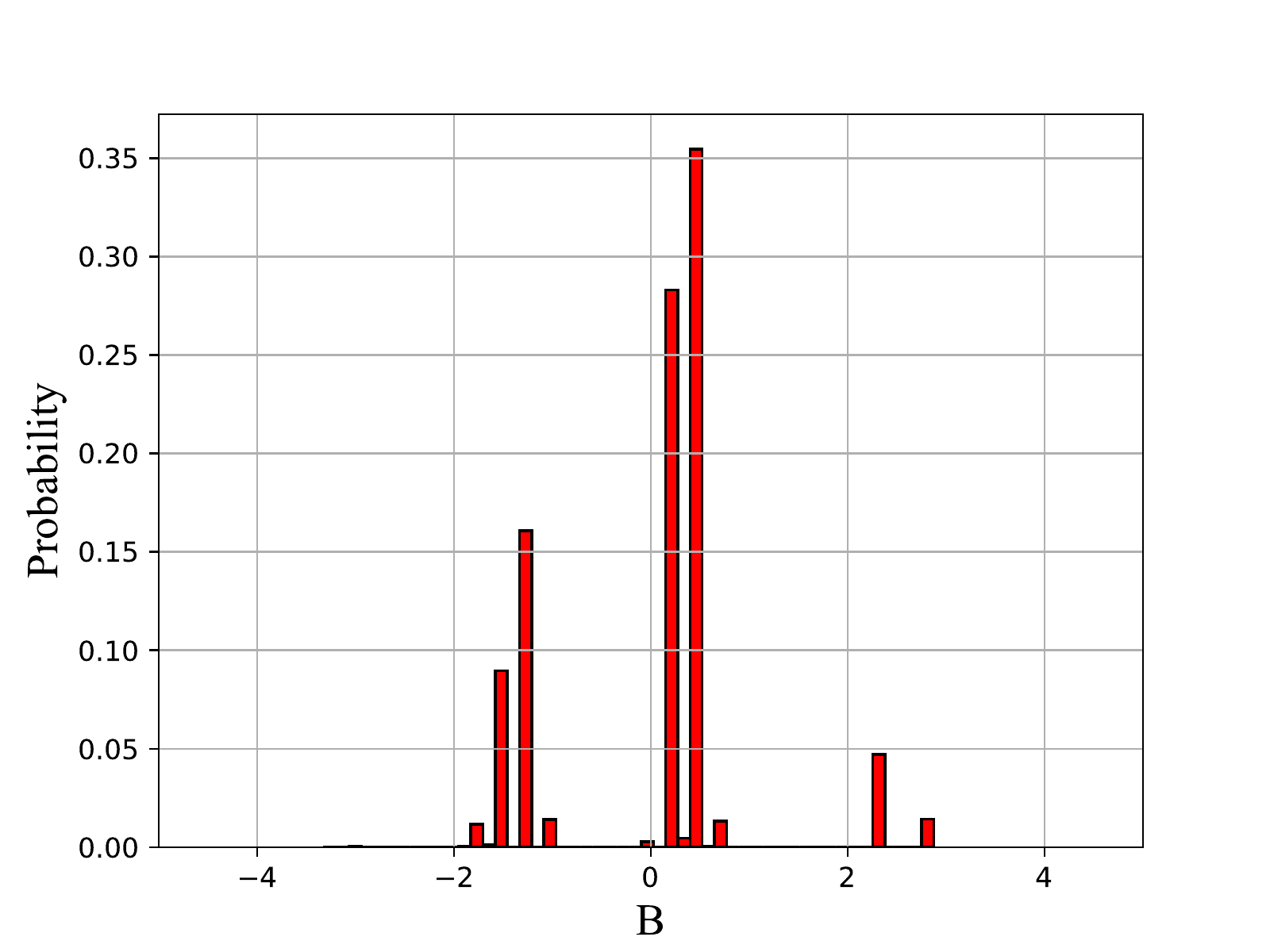}}
  \subfigure[]{\includegraphics[width = 3.4cm,height = 2.65cm]{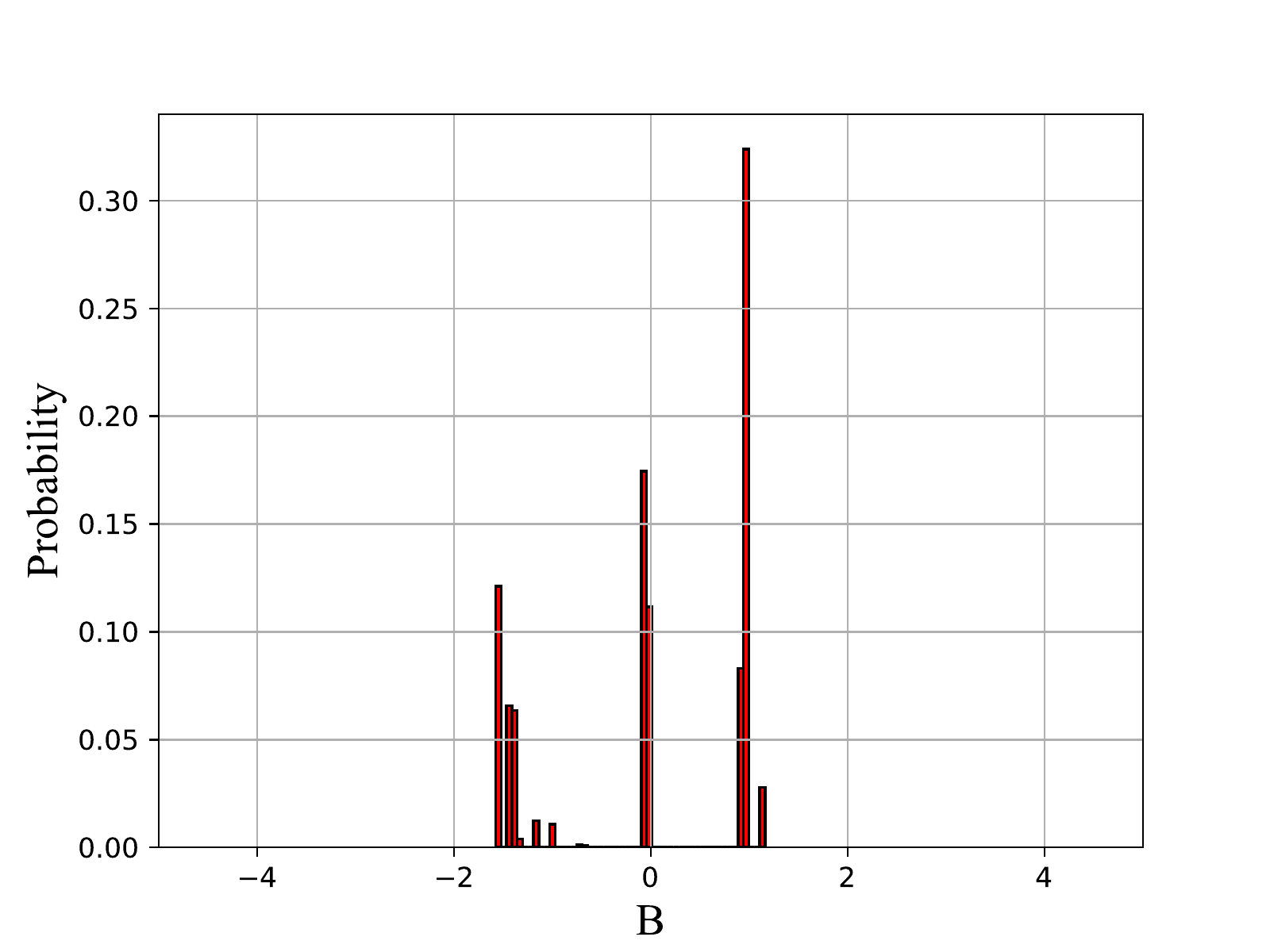}}
  \caption{A comparison example for color distortion in the CMDM database \cite{database}. (a) represents the snapshots of the reference 3D mesh while (b)-(e) stand for the snapshots of the meshes with 4 increasing levels of color information quantization. (f), (k), (p) represent the normalized probability distributions of LAB channels for (a) model, (g), (l), (q) represent the normalized probability distributions of LAB channels for (b) model, (h), (m), (r) represent the normalized probability distributions of LAB channels for (c) model,(i), (n), (s) represent the normalized probability distributions of LAB channels for (d) model, and (j), (o), (t) represented the normalized probability distributions of LAB channels for (e) model respectively.  }
  \label{fig:color}
\end{figure*}

\begin{figure*}[tbp]
\centering
\subfigure[]{
\begin{minipage}[t]{0.23\linewidth}
\centering
\includegraphics[width=3cm,height = 4.5cm]{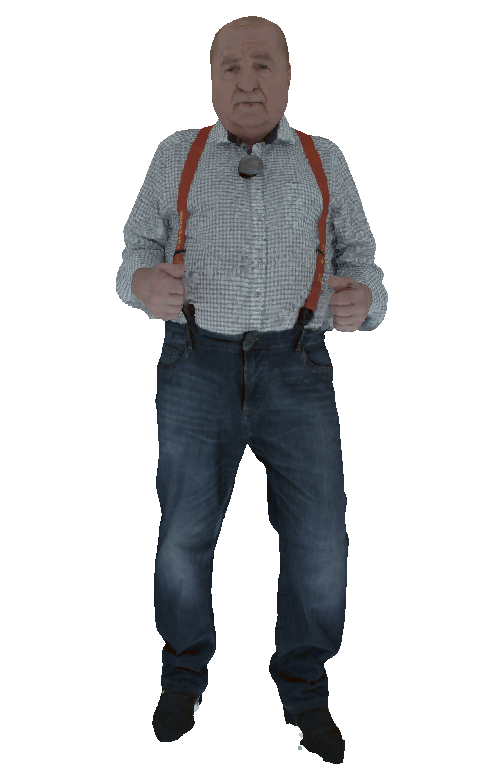}
\end{minipage}%
}%
\subfigure[]{
\begin{minipage}[t]{0.23\linewidth}
\centering
\includegraphics[width=3cm,height = 4.5cm]{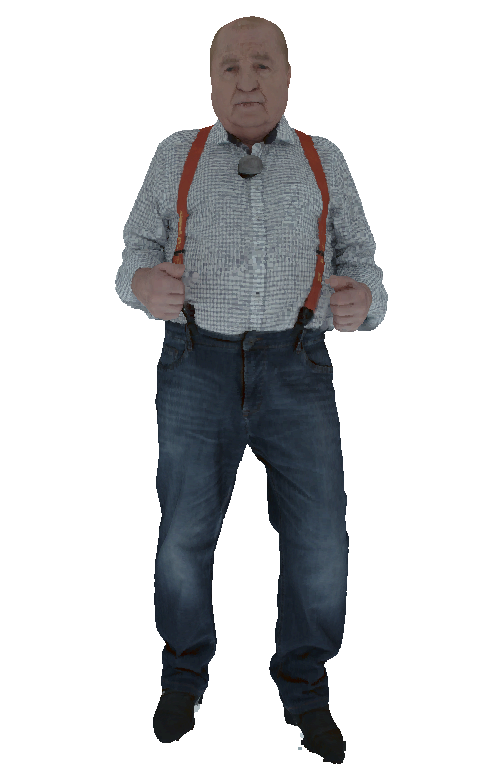}
\end{minipage}%
}%
\subfigure[]{
\begin{minipage}[t]{0.23\linewidth}
\centering
\includegraphics[width=3cm,height = 4.5cm]{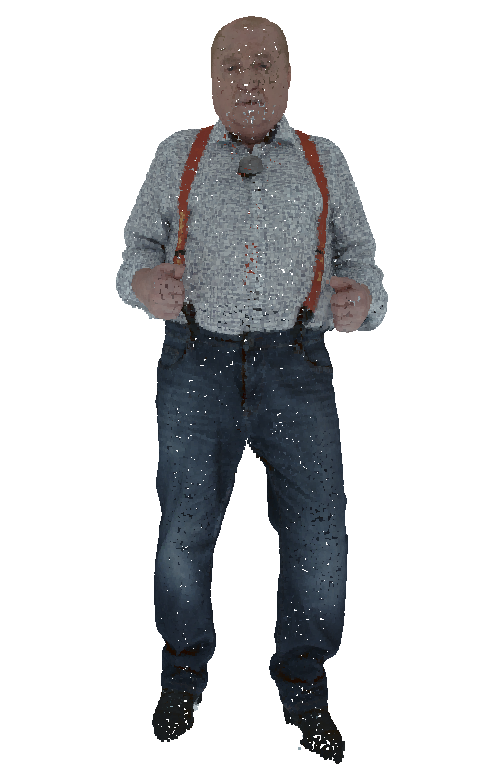}
\end{minipage}
}%
\subfigure[]{
\begin{minipage}[t]{0.23\linewidth}
\centering
\includegraphics[width=3cm,height = 4.5cm]{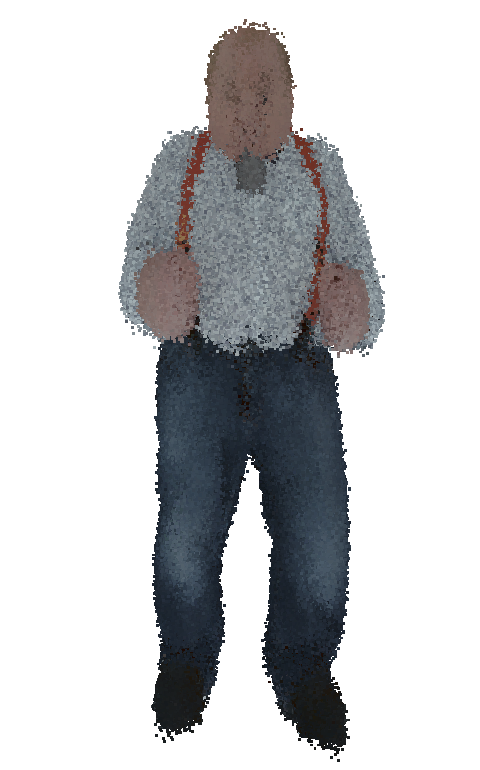}
\end{minipage}
}%

\subfigure[]{
\begin{minipage}[t]{0.23\linewidth}
\centering
\includegraphics[width=4.1cm]{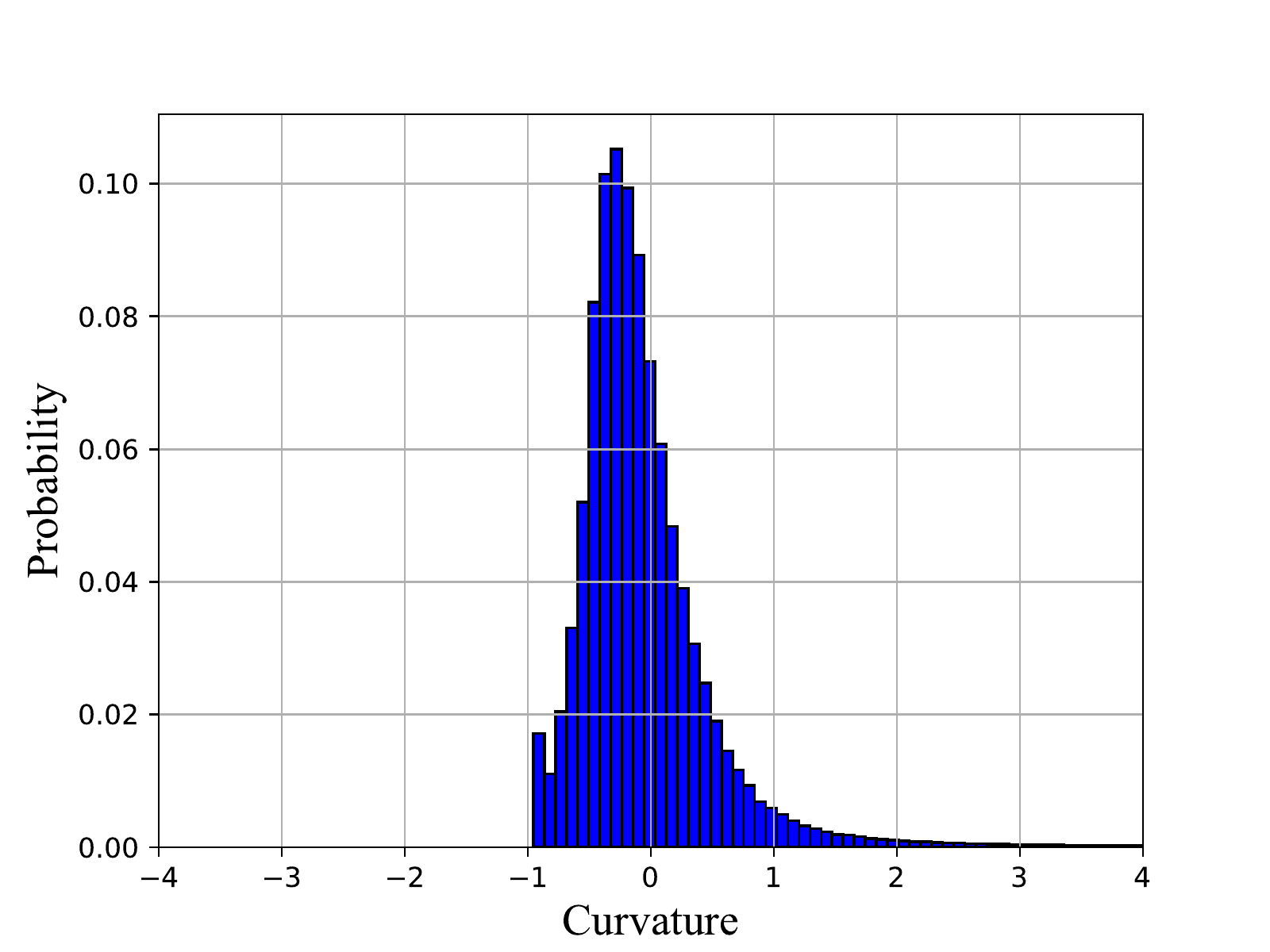}
\end{minipage}%
}%
\subfigure[]{
\begin{minipage}[t]{0.23\linewidth}
\centering
\includegraphics[width=4.1cm]{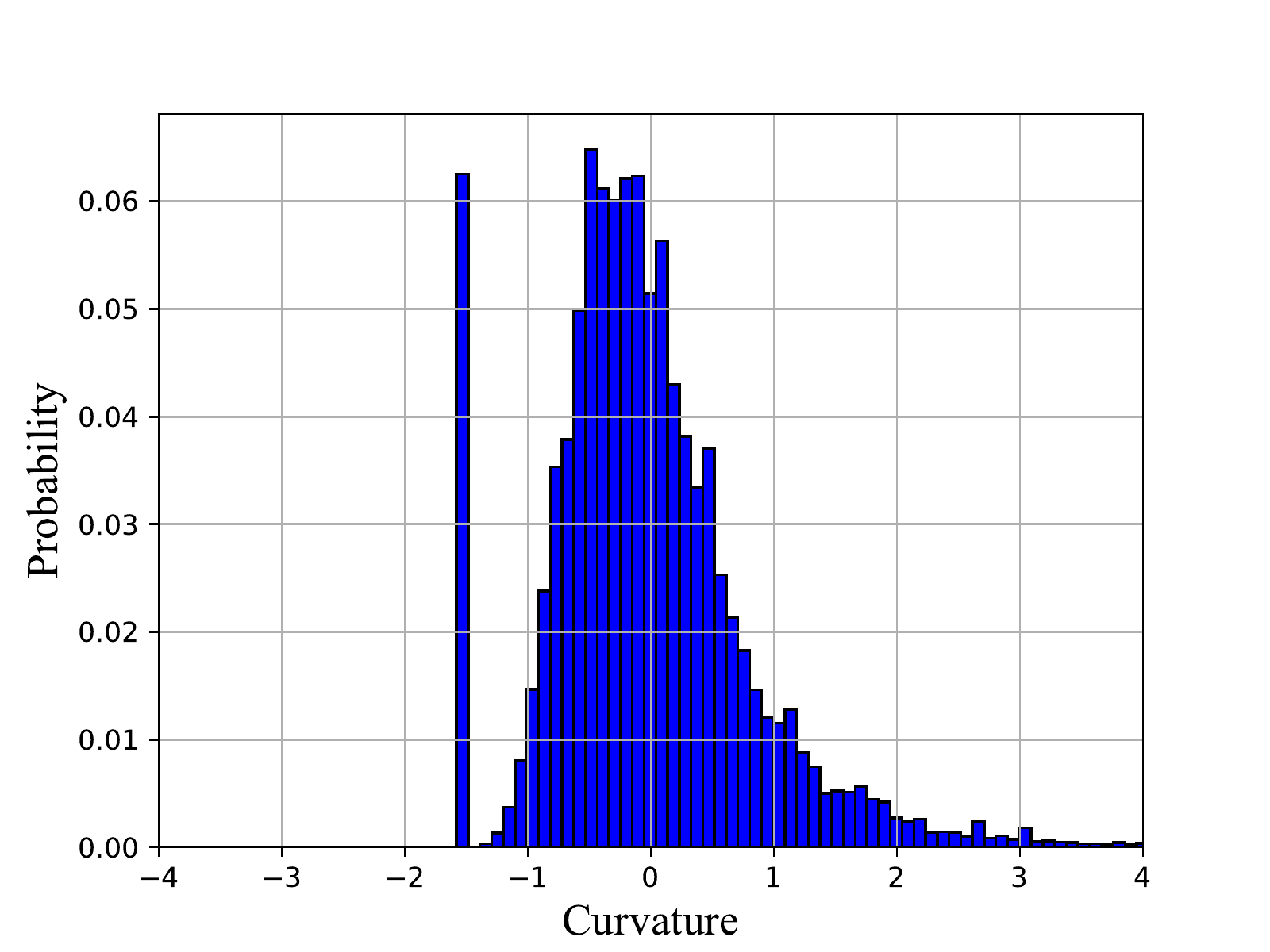}
\end{minipage}%
}%
\subfigure[]{
\begin{minipage}[t]{0.23\linewidth}
\centering
\includegraphics[width=4.1cm]{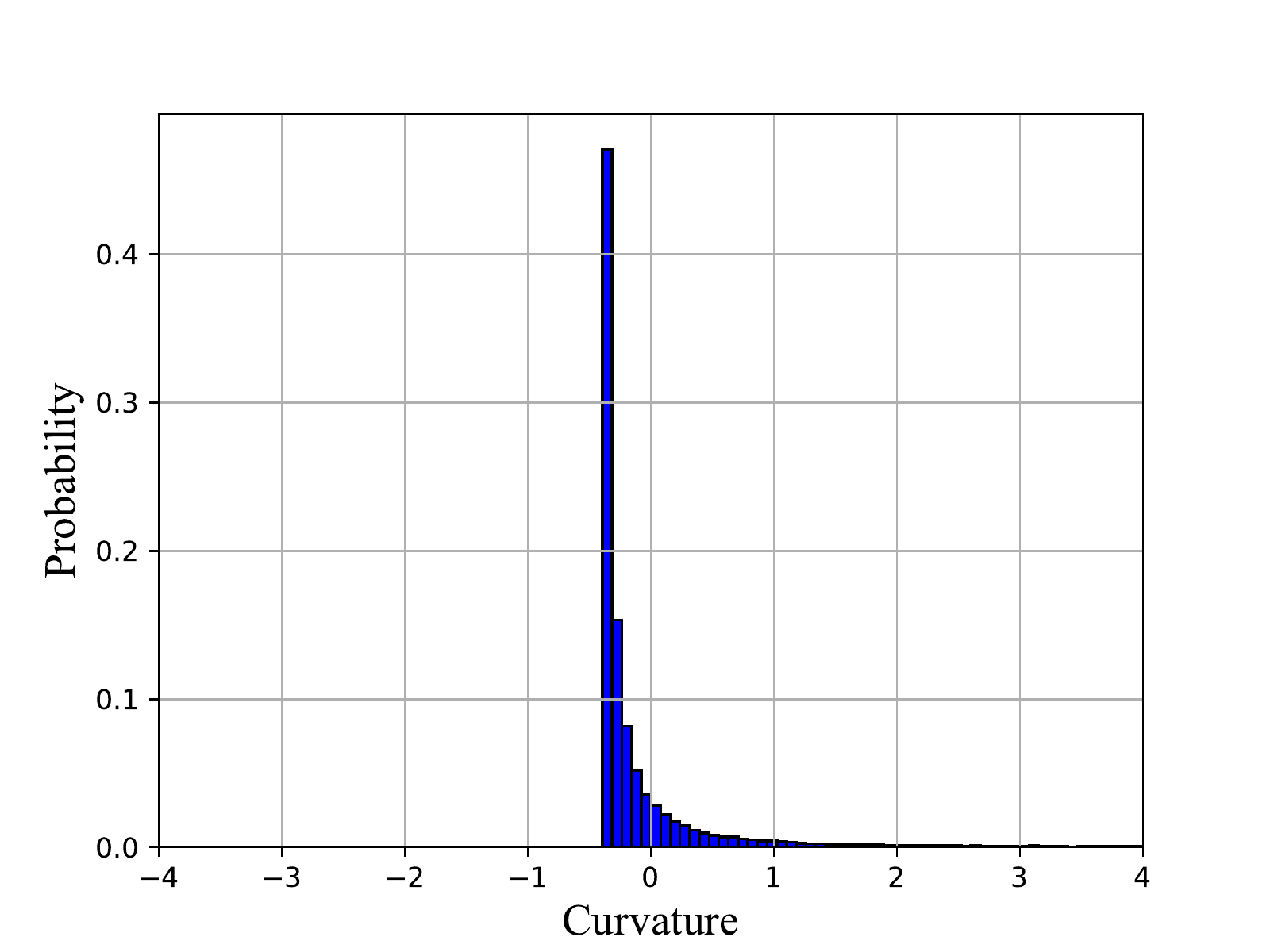}
\end{minipage}
}%
\subfigure[]{
\begin{minipage}[t]{0.23\linewidth}
\centering
\includegraphics[width=4.1cm]{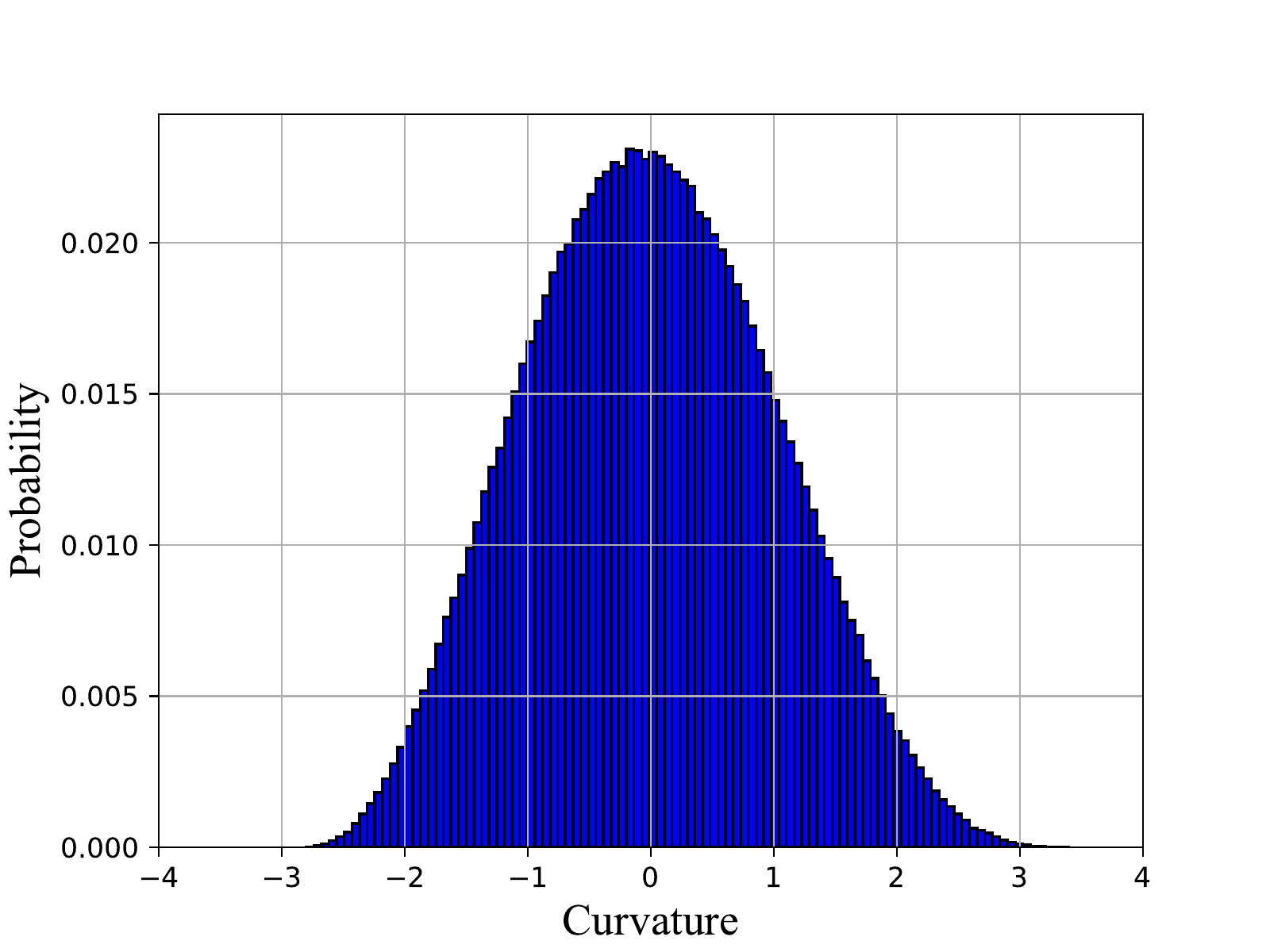}
\end{minipage}
}%

\subfigure[]{
\begin{minipage}[t]{0.23\linewidth}
\centering
\includegraphics[width=4.1cm]{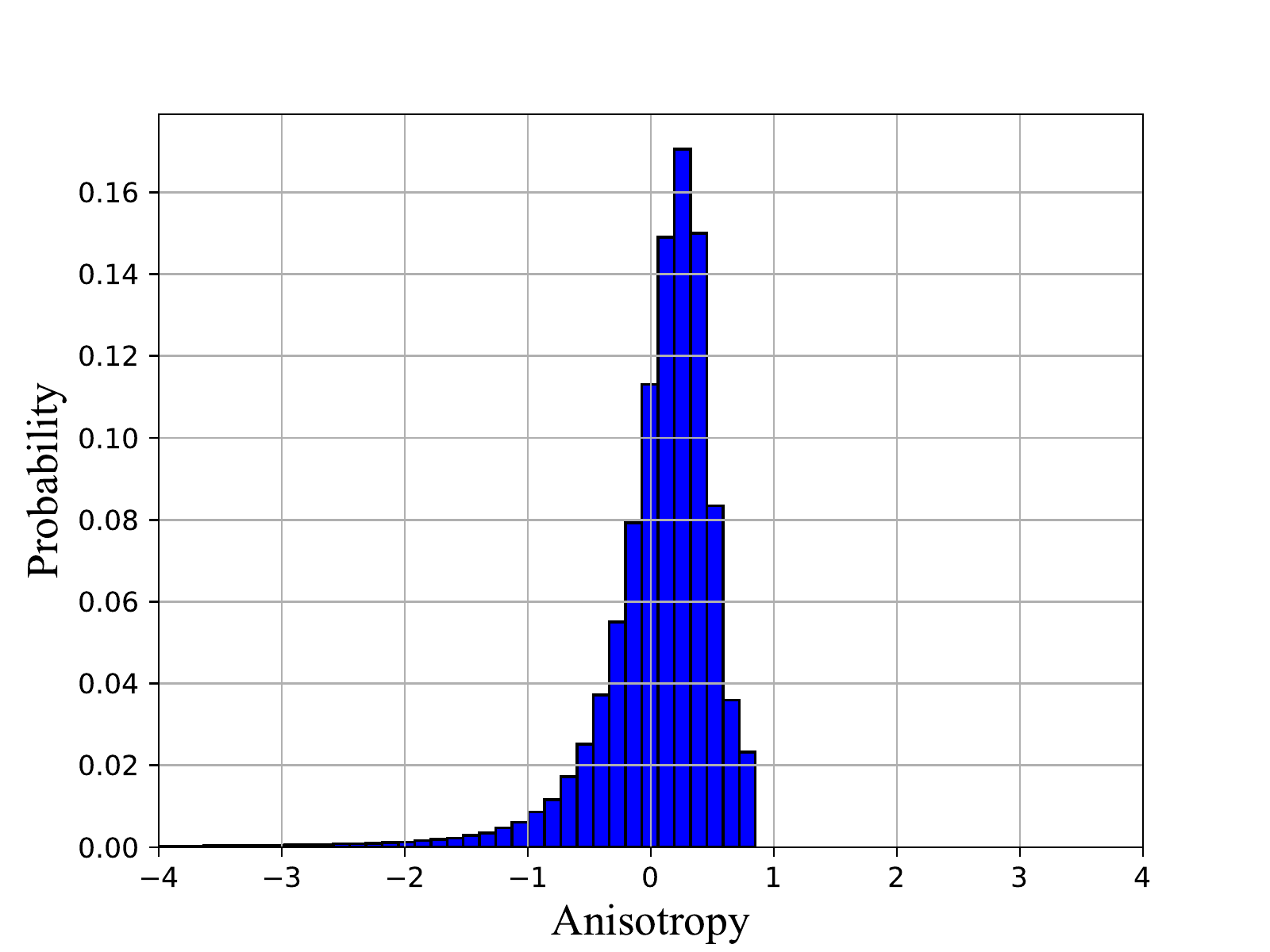}
\end{minipage}%
}%
\subfigure[]{
\begin{minipage}[t]{0.23\linewidth}
\centering
\includegraphics[width=4.1cm]{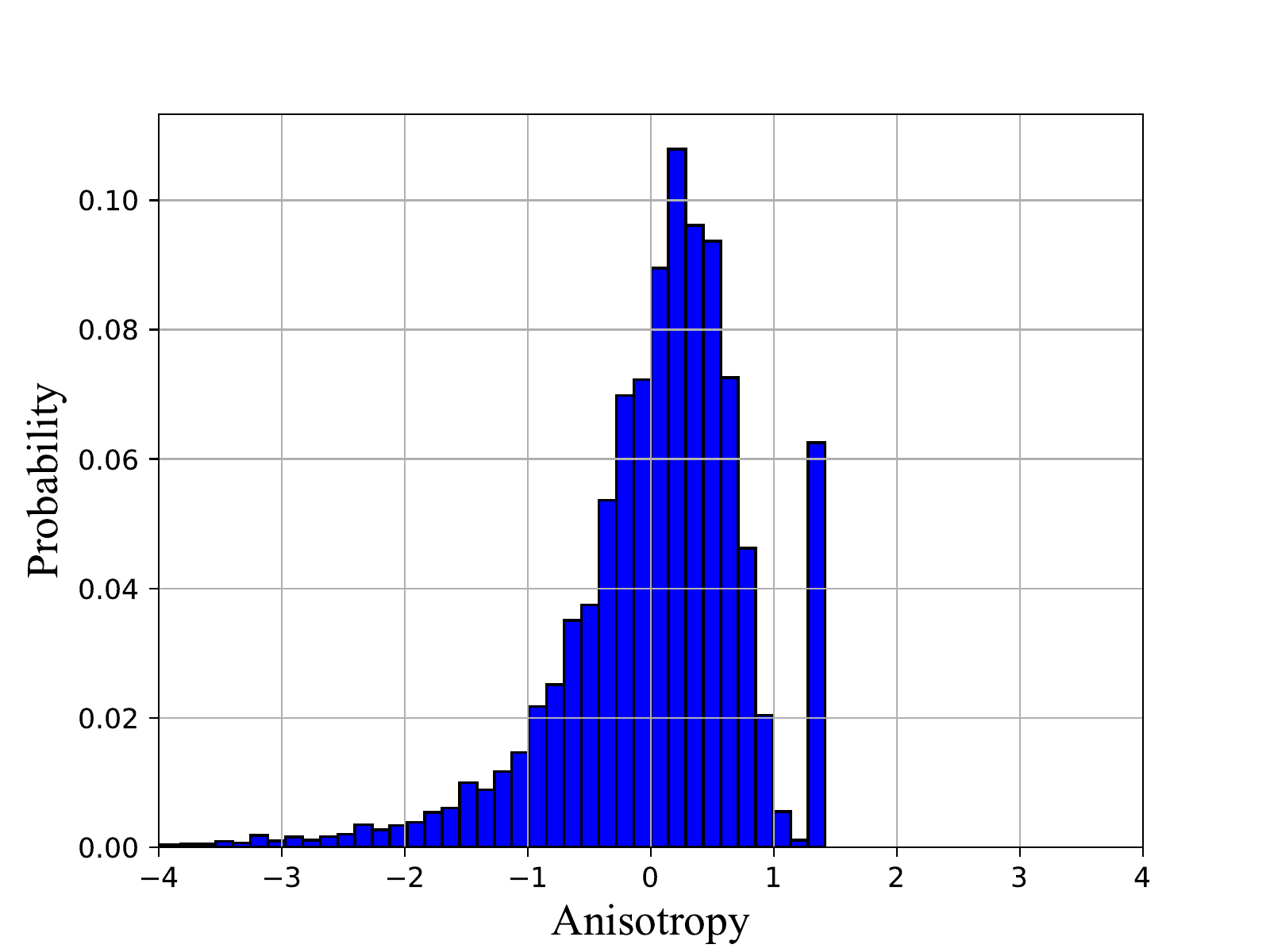}
\end{minipage}%
}%
\subfigure[]{
\begin{minipage}[t]{0.23\linewidth}
\centering
\includegraphics[width=4.1cm]{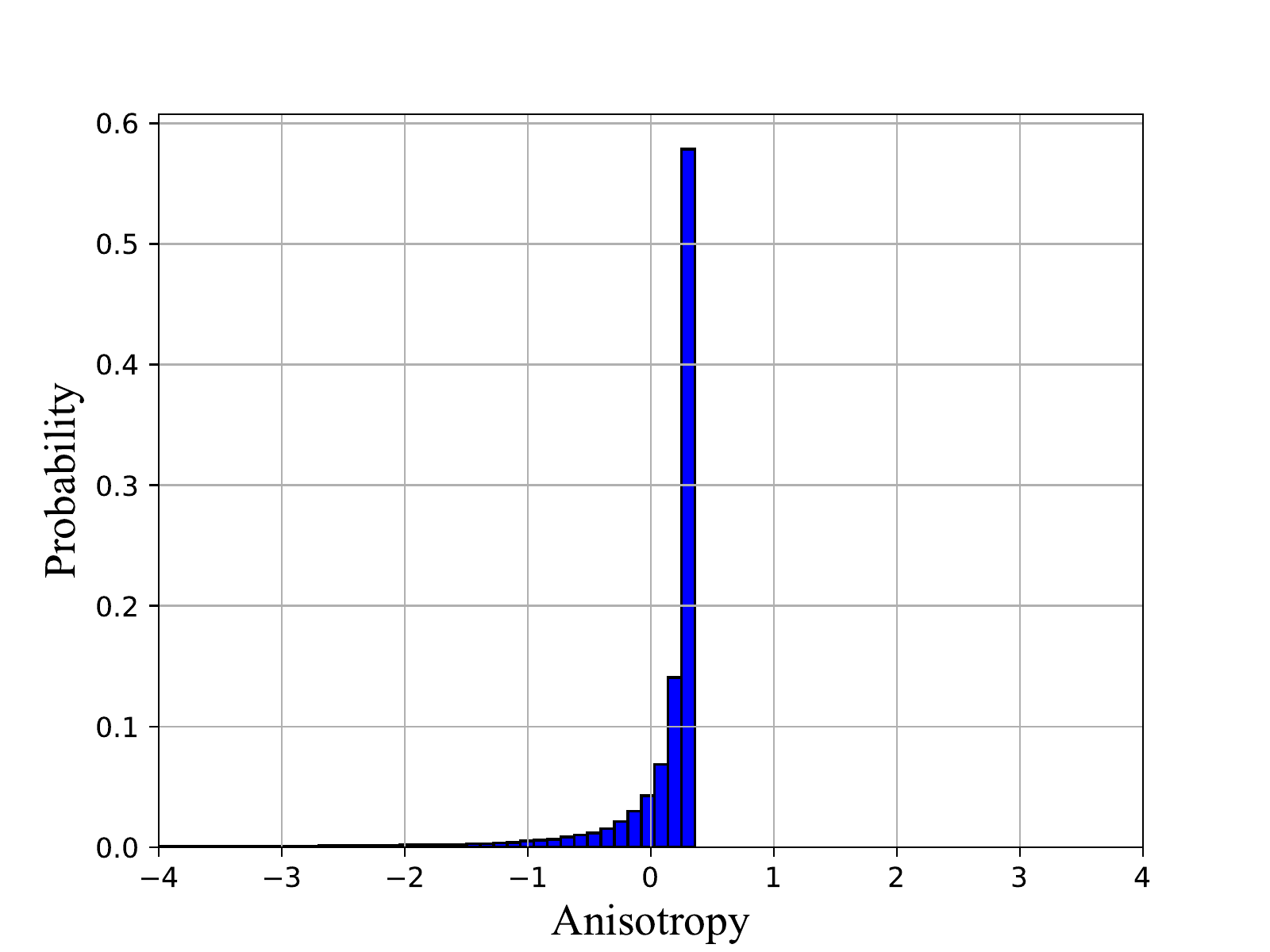}
\end{minipage}
}%
\subfigure[]{
\begin{minipage}[t]{0.23\linewidth}
\centering
\includegraphics[width=4.1cm]{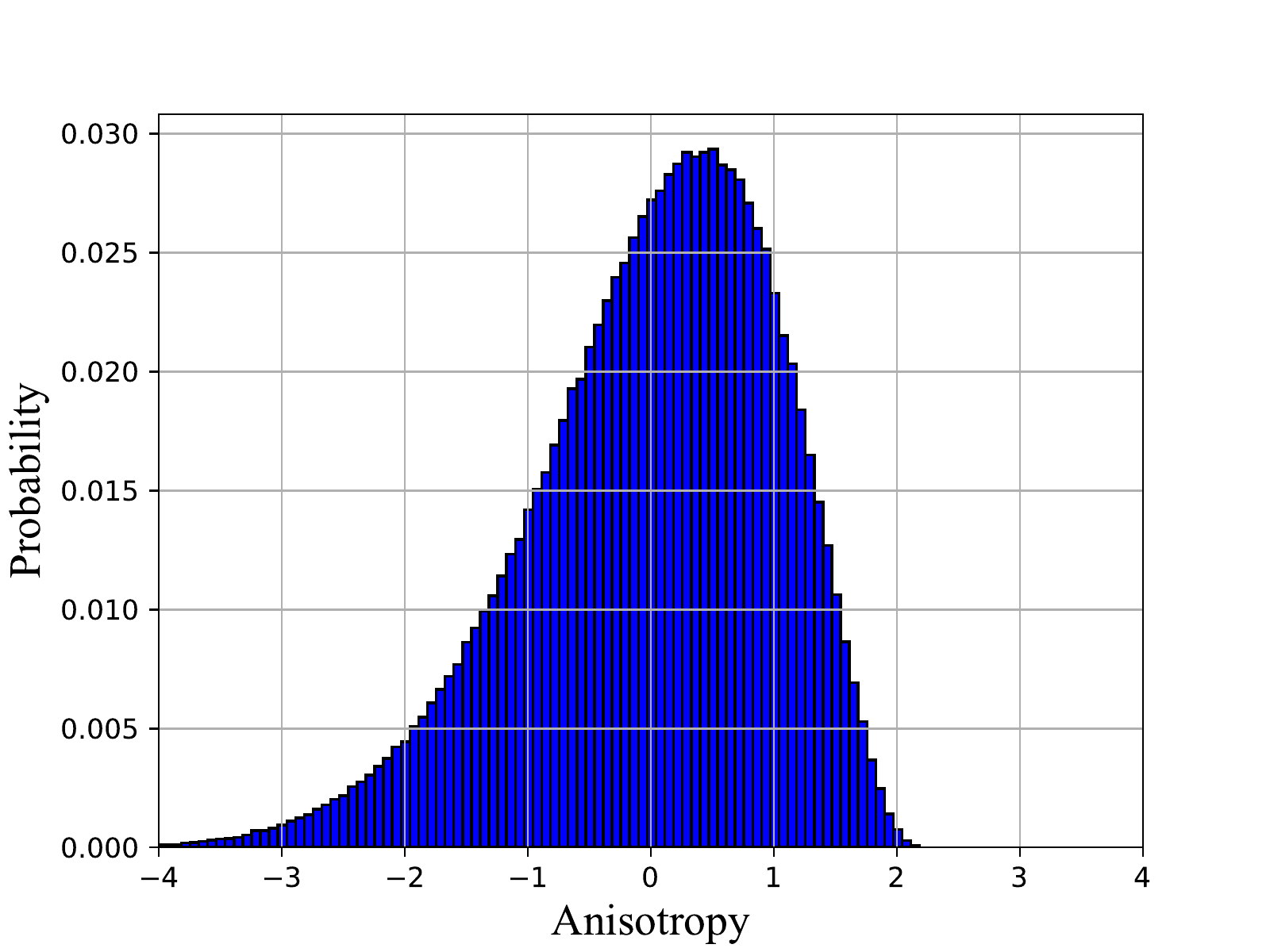}
\end{minipage}
}%

\subfigure[]{
\begin{minipage}[t]{0.23\linewidth}
\centering
\includegraphics[width=4.1cm]{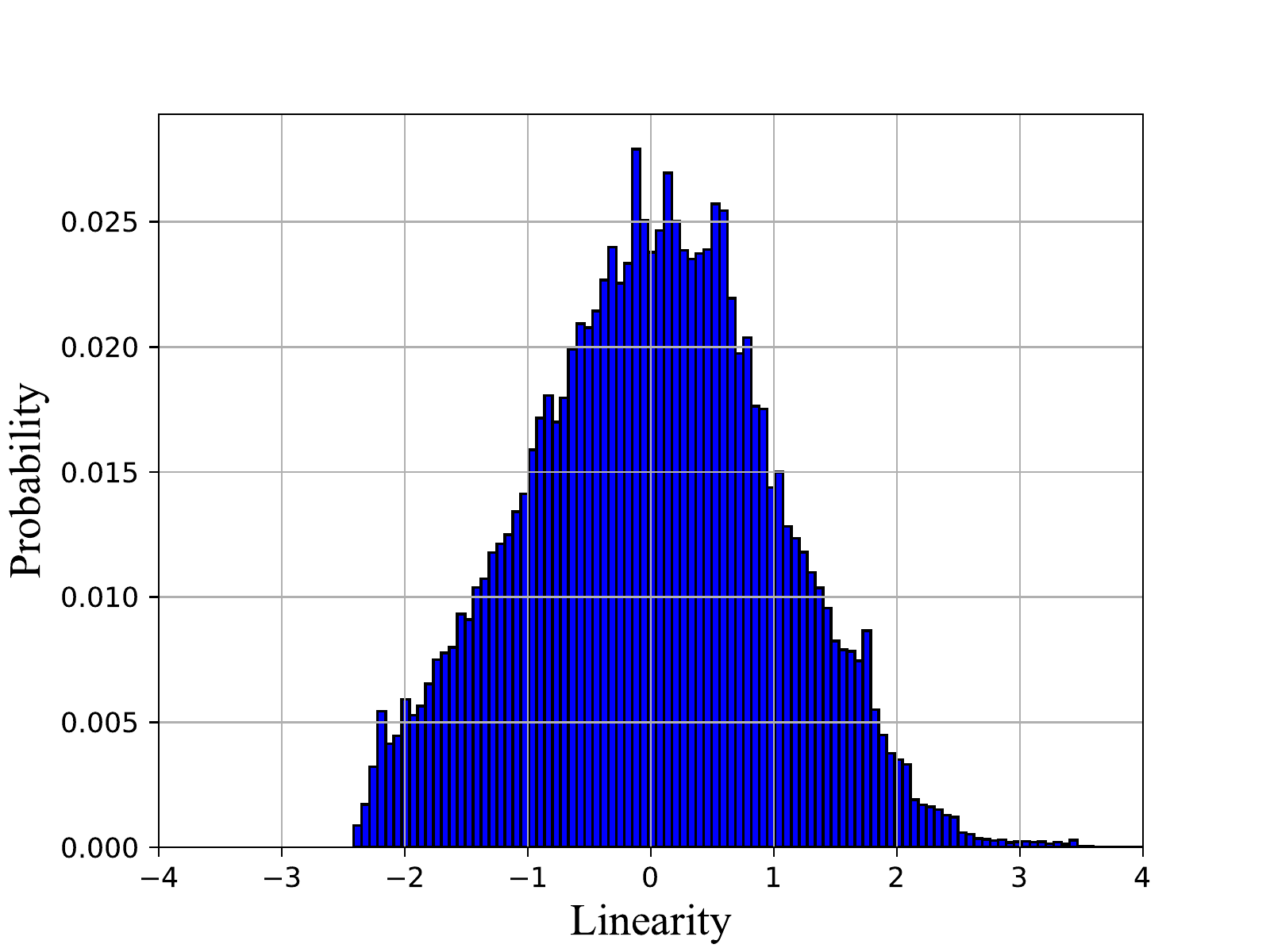}
\end{minipage}%
}%
\subfigure[]{
\begin{minipage}[t]{0.23\linewidth}
\centering
\includegraphics[width=4.1cm]{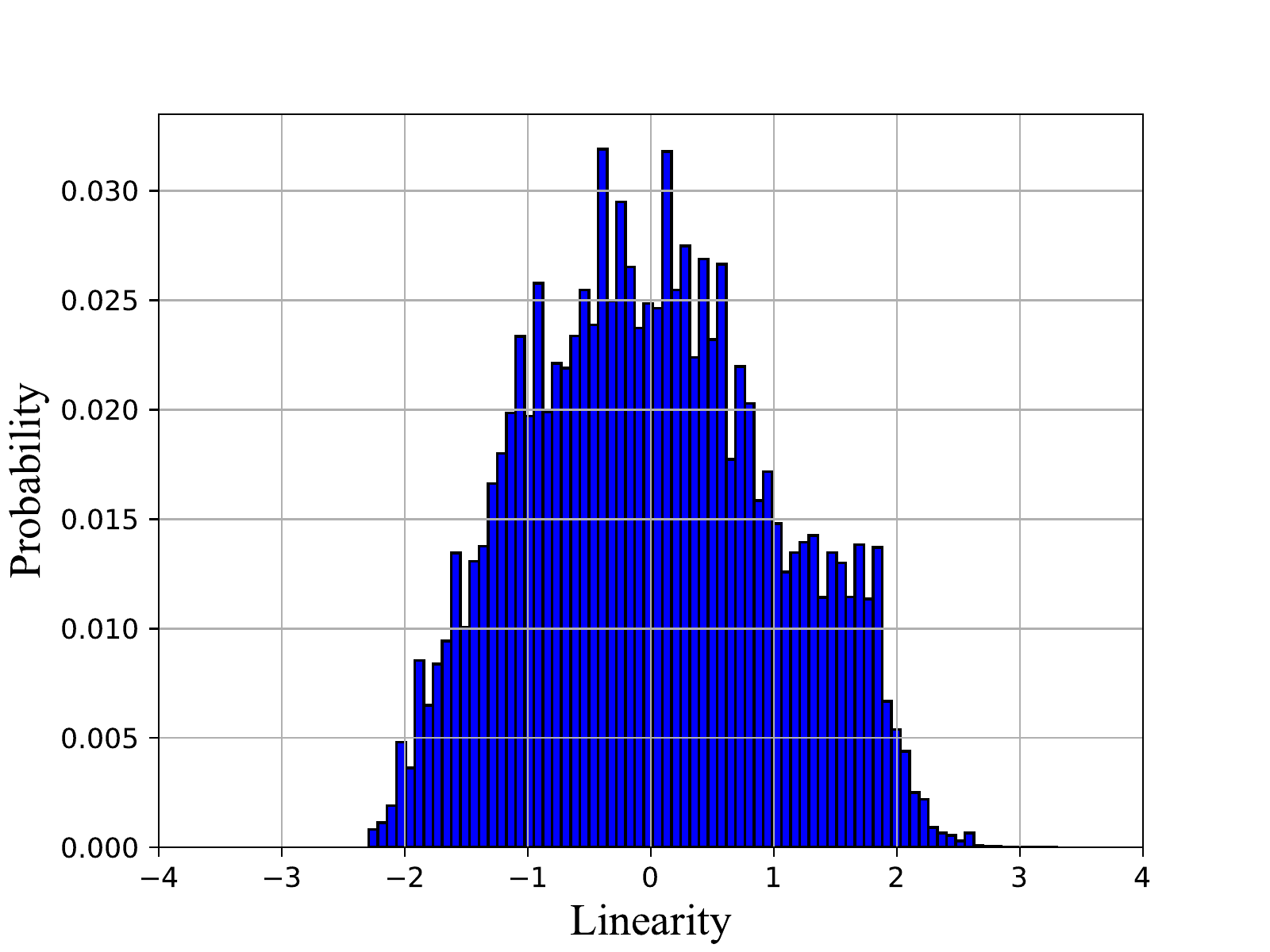}
\end{minipage}%
}%
\subfigure[]{
\begin{minipage}[t]{0.23\linewidth}
\centering
\includegraphics[width=4.1cm]{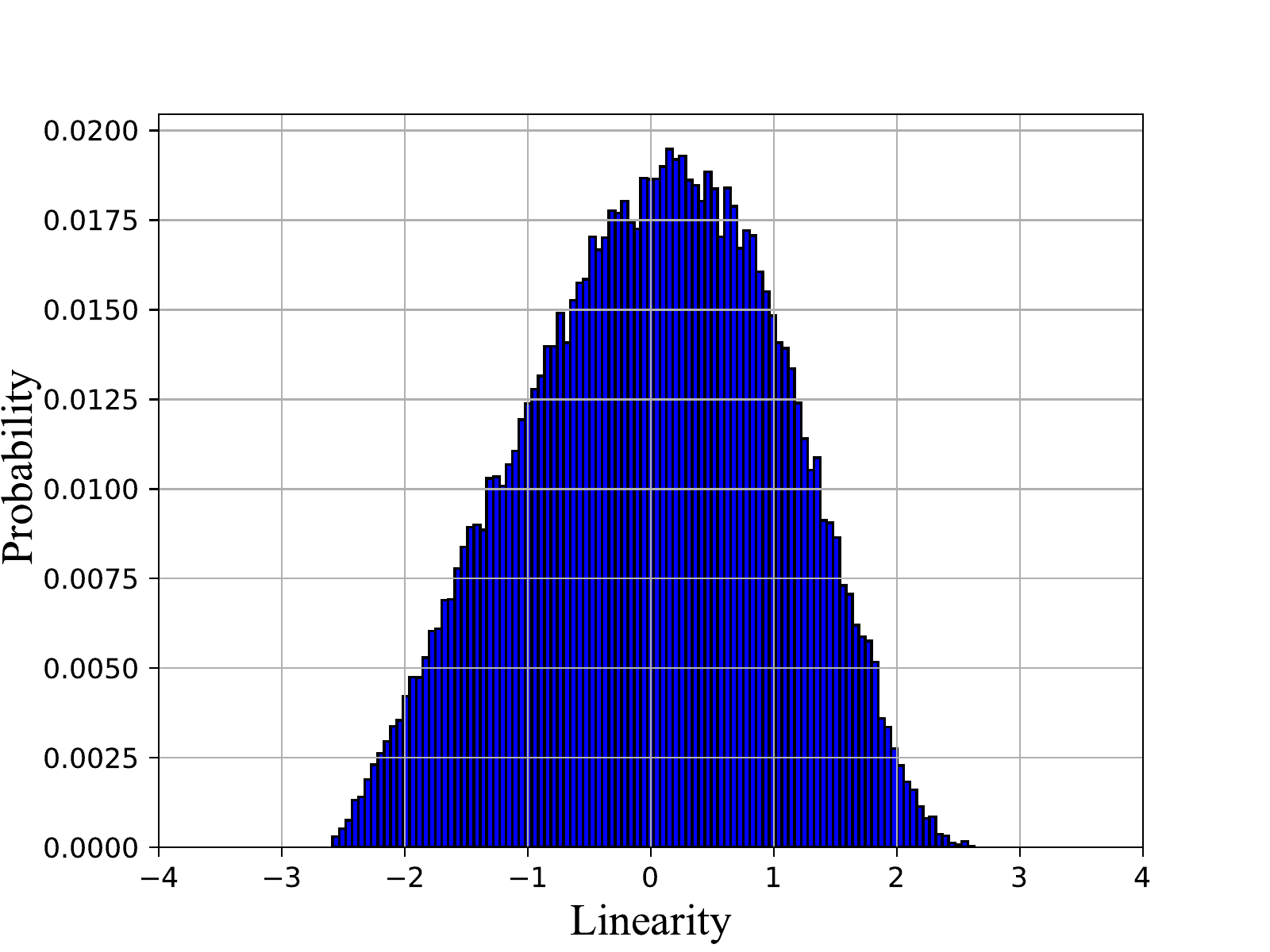}
\end{minipage}
}%
\subfigure[]{
\begin{minipage}[t]{0.23\linewidth}
\centering
\includegraphics[width=4.1cm]{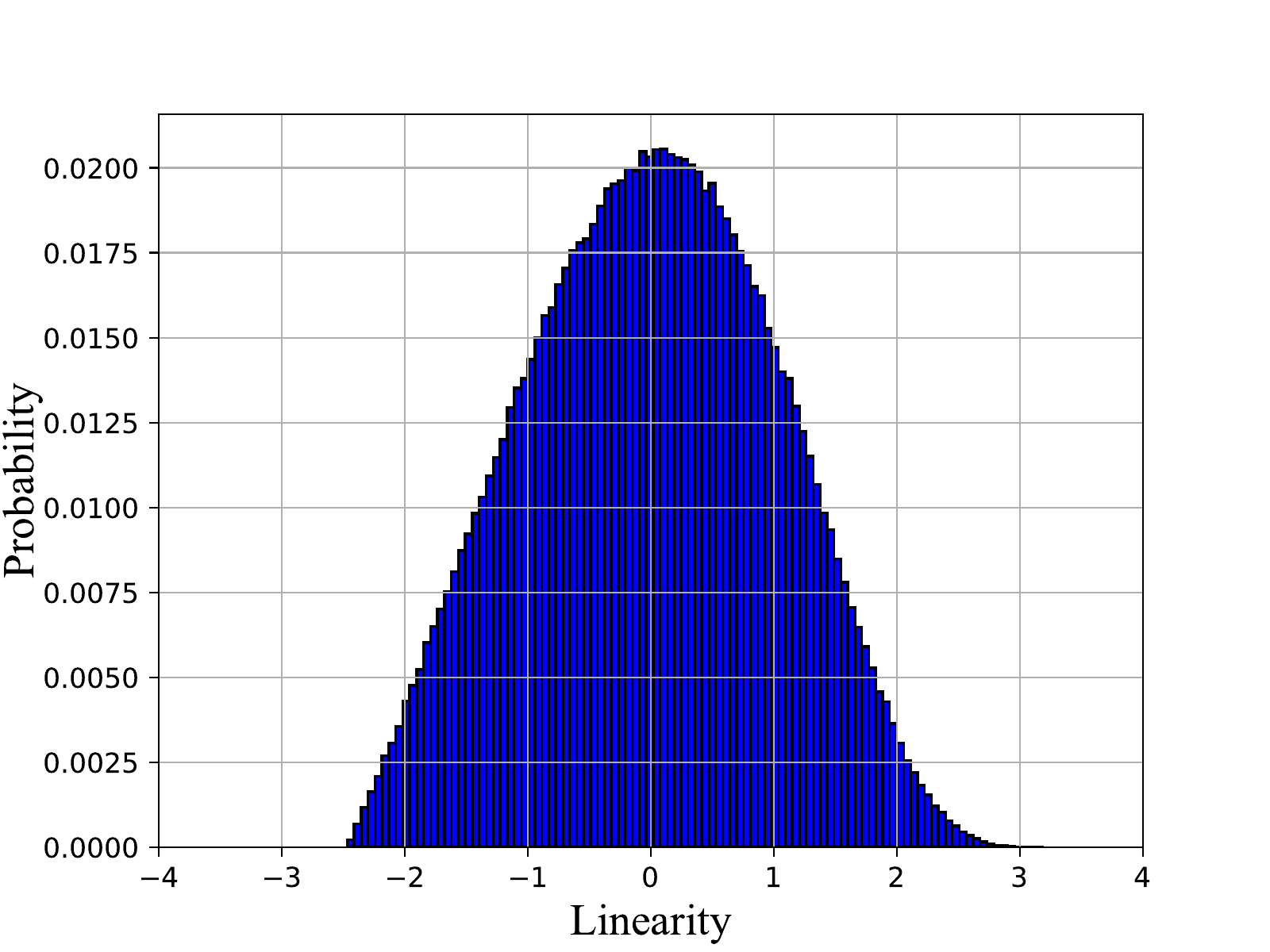}
\end{minipage}
}%

\subfigure[]{
\begin{minipage}[t]{0.23\linewidth}
\centering
\includegraphics[width=4.1cm]{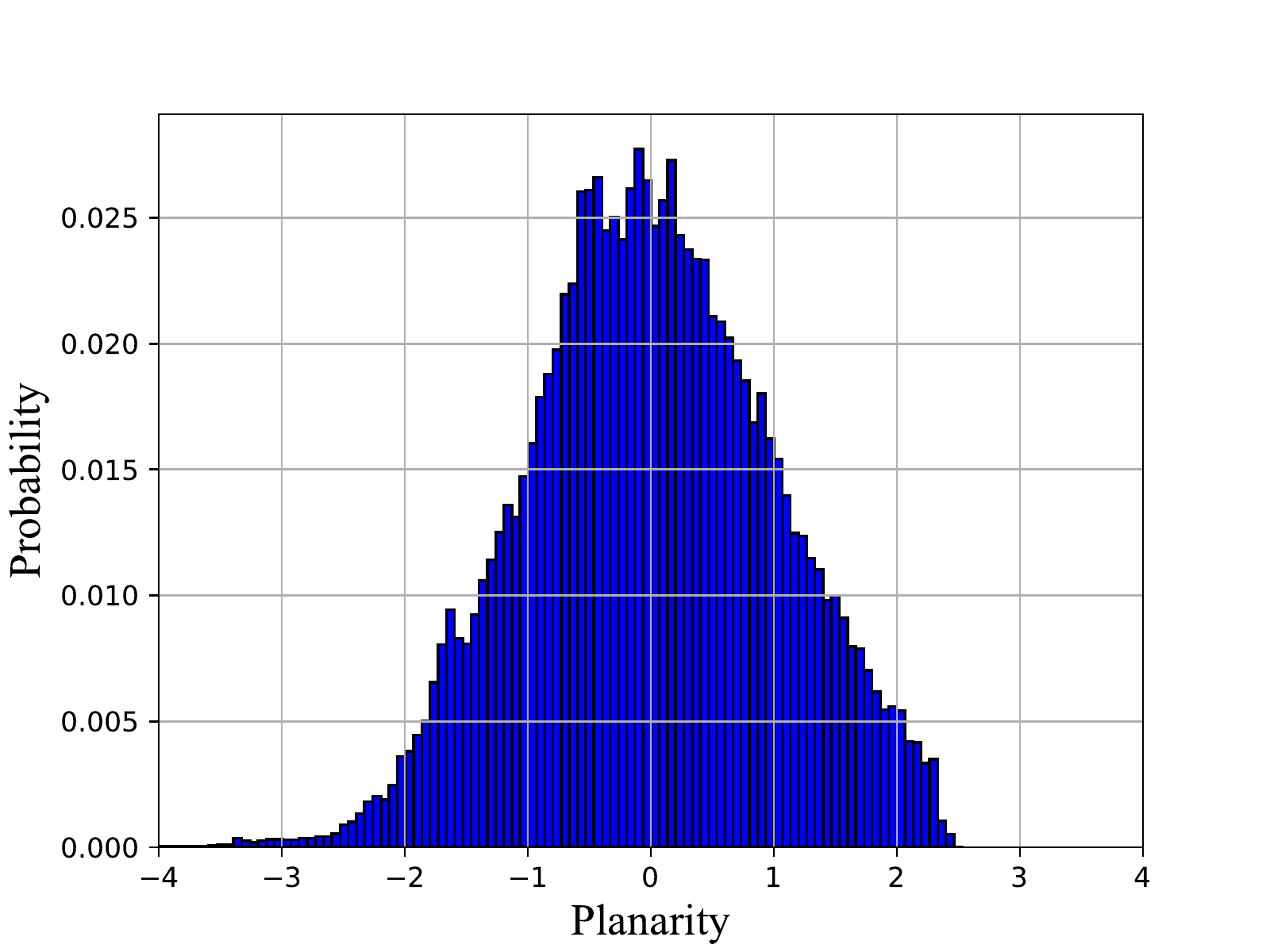}
\end{minipage}%
}%
\subfigure[]{
\begin{minipage}[t]{0.23\linewidth}
\centering
\includegraphics[width=4.1cm]{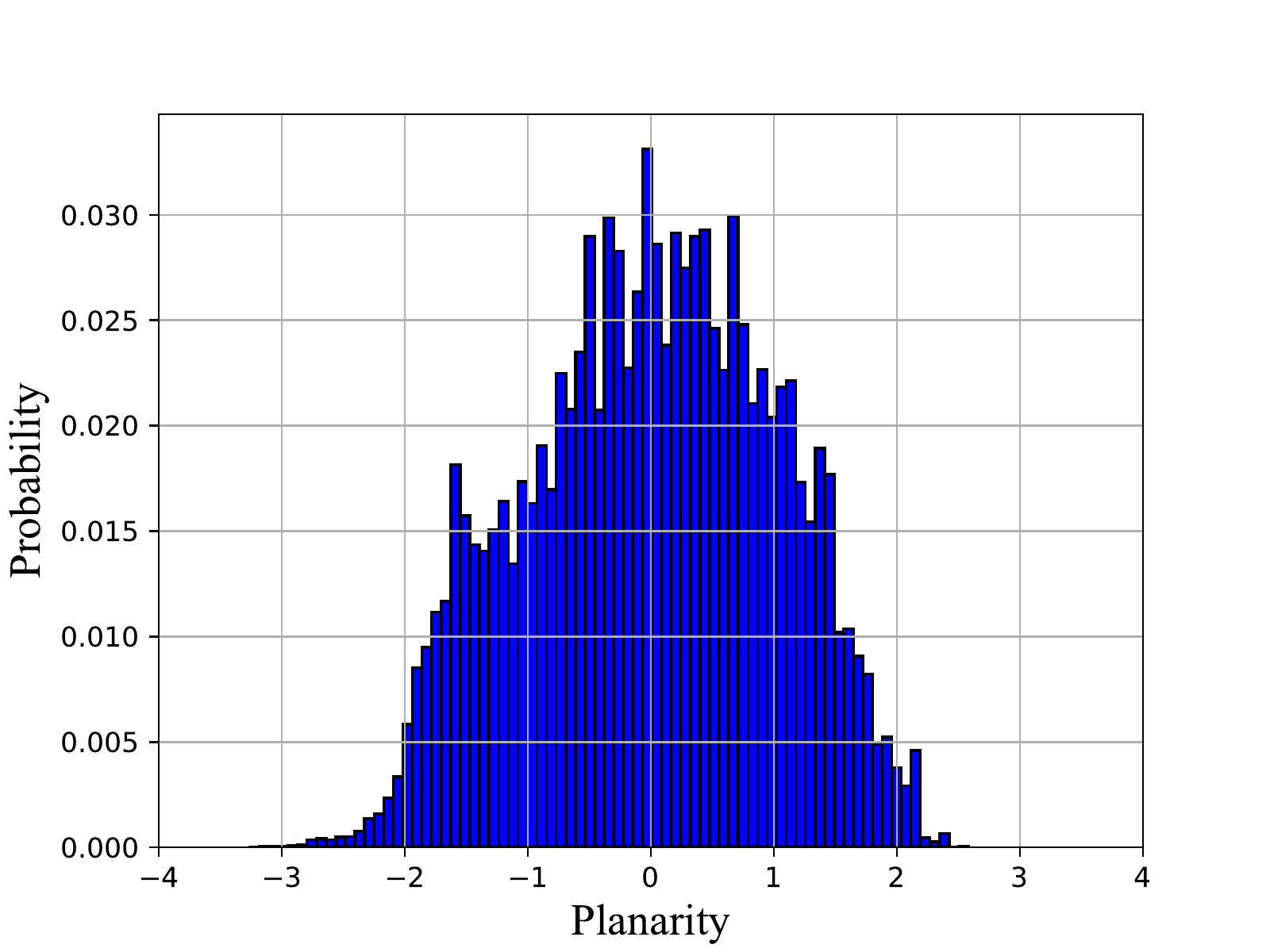}
\end{minipage}%
}%
\subfigure[]{
\begin{minipage}[t]{0.23\linewidth}
\centering
\includegraphics[width=4.1cm]{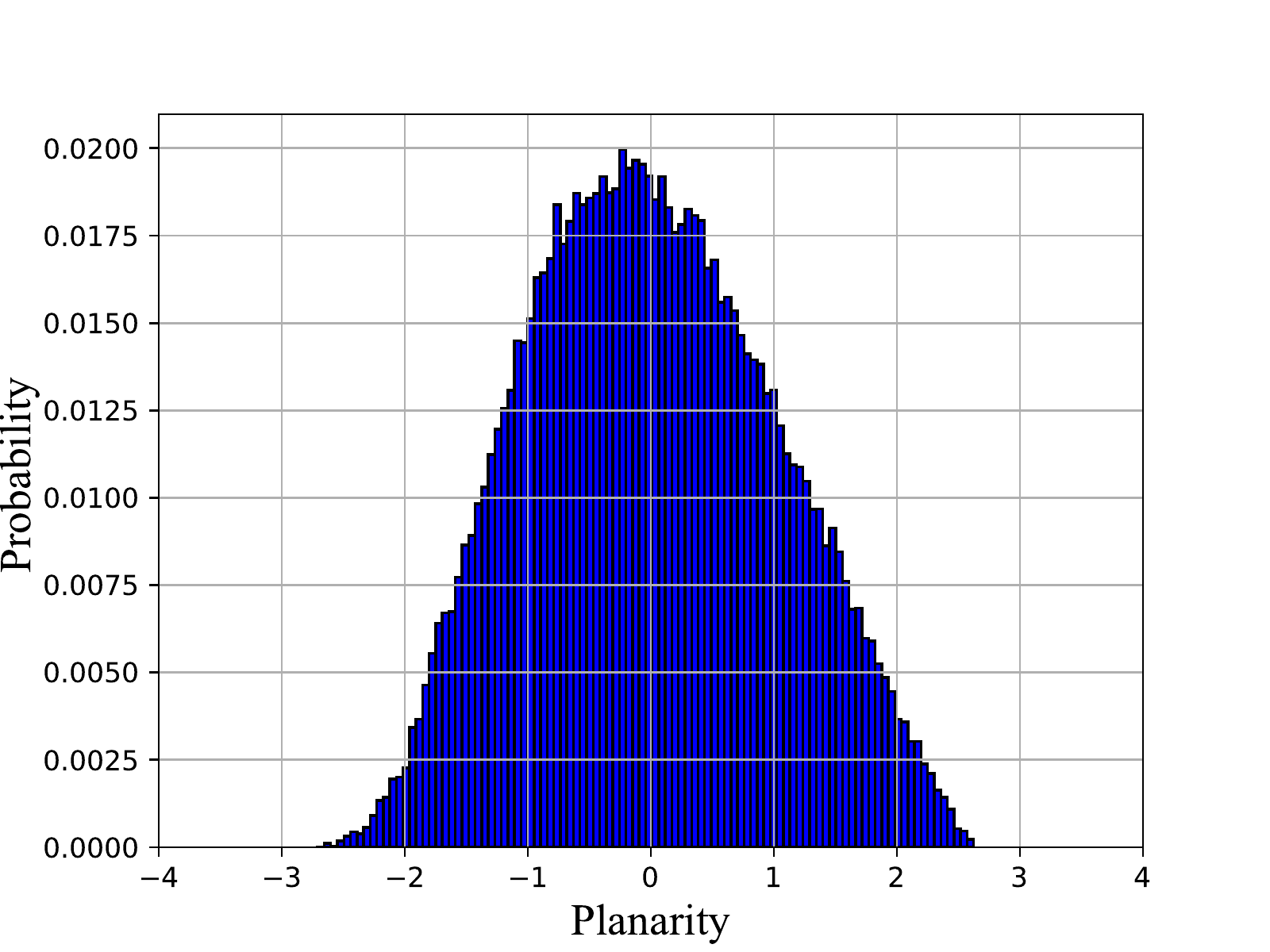}
\end{minipage}
}%
\subfigure[]{
\begin{minipage}[t]{0.23\linewidth}
\centering
\includegraphics[width=4.1cm]{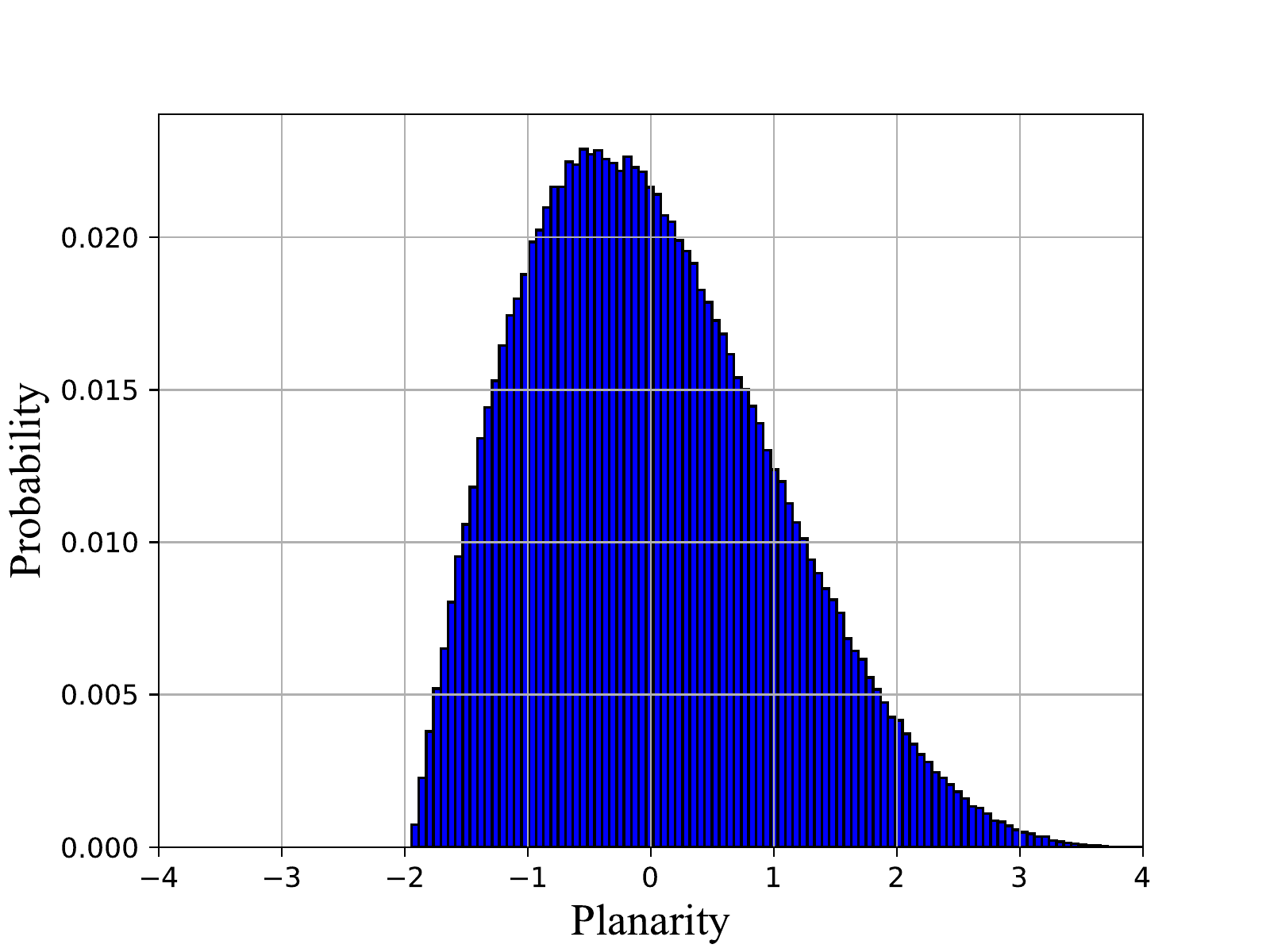}
\end{minipage}
}%
\centering
\caption{Examples of point cloud samples from SJTU-PCQA database \cite{sjtu-pcqa}. (a) is the reference point cloud sample, while (b), (c), and (d) are 3 distorted point cloud samples with different distortion types (compression, downsampling, geometry Gaussian noise). Some features' ($Cur,Ani, Lin, Pla$) normalized probability distributions are selected as examples. More specifically, (e), (i), (m), and (q) are the corresponding features'  normalized probability distributions of (a) model, (f), (j), (n), and (r) are the corresponding features' normalized probability distributions of (b) model, (g), (k), (o), and (s) are the corresponding features' normalized probability distributions of (c) model, (h), (l), (p), and (t) are the corresponding features' normalized probability distributions of (d) model respectively.  }
\label{fig:sjtu-pcqa}
\end{figure*}

\section{Estimating Statistical Parameters}
\label{sec:estimate nss parameters}
Through prior knowledge and observations of the corresponding feature distributions, we find that the characteristic statistical parameters of NSS models can be changed by the presence of distortion. Therefore, we choose entropy and several NSS models including the generalized Gaussian distribution (GGD), the general asymmetric generalized Gaussian distribution (AGGD), and the shape-rate Gamma distribution for parameters estimation to quantify the perceptual quality of 3D models.

\subsection{Basic Statistical Parameters}
For each set of features, we exploit the normalization operation as the pre-processing:
\begin{equation}
 \begin{aligned}
    & \hat{F}=\frac{F-mean(F)}{std(F)+C},\\
    & F \in \{F_{geo}, F_{col}\} ,
\end{aligned} 
\label{equ:mean} 
\end{equation}
where $F$ represents the feature domain, $mean(\cdot)$ is the average function, $std(\cdot)$ denotes the standard deviation function, and $C$ is a small constant to avoid instability.
Entropy is believed to be highly correlated with the quantization distortion. Fig. \ref{fig:color} presents the normalized probability distributions of LAB channels for color quantization distortion. It can be observed that with the increasing quantization levels, the corresponding distributions become significantly sparse, meaning the number of distinct colors is reduced.  
Considering that some simplification and compression algorithms usually introduce quantization operations to the 3D models, we decide to use entropy as one of the quality-aware features.
Then the entropy can be derived as:
\begin{equation}
   \begin{aligned}
    &E = \boldsymbol{\rm{Entropy}}(\hat{F}),\\
    &\hat{F} \in \{\hat{F_{geo}}, \hat{F_{col}}\}.
\end{aligned}
\label{equ:entropy} 
\end{equation}
where $\mathbf{Entropy}(\cdot)$ indicates the entropy function, $\hat{F_{geo}}$ and $\hat{F_{col}}$ are the normalized feature distributions for geometry and color respectively.

\subsection{GGD Parameters}
Fig. \ref{fig:sjtu-pcqa} shows one reference model and three distorted point cloud samples with  different distortion types  (compression,  downsampling,  geometry  Gaussian  noise). It can be observed that the reference curvature and anisotropy distributions shown in Fig. \ref{fig:sjtu-pcqa} (e, i) exhibit Gamma-like appearance while the reference linearity and planarity distributions shown in Fig. \ref{fig:sjtu-pcqa} (m, q) tend to be Gaussian-like. Obviously, with different distortion types, the shapes of the distributions are changed. For example, with closer inspections, we can see that compression adds more weight to the tail of curvature and anisotropy distributions shown in Fig. \ref{fig:sjtu-pcqa} (f, j) while downsampling makes curvature and anisotropy distributions shown in Fig. \ref{fig:sjtu-pcqa} (g, k) more centered. What's more, geometry Gaussian noise makes all the geometry distributions more smooth and adds more weight to the tail of curvature and anisotropy distributions shown in Fig. \ref{fig:sjtu-pcqa} (h, l) on both sides. 

The GGD distribution is effective for capturing a broad spectrum of statistics and reflecting the changes in distribution shapes, especially for the tail part \cite{sharifi1995estimation}.
To capture the corresponding characteristics of 3D model distributions, we propose to use the generalized Gaussian distribution (GGD):
\begin{equation}
\begin{aligned} 
&GGD(x;\alpha,\beta^2)=\frac{\alpha}{2 \beta \Gamma(1 / \alpha)} \exp \left(-\left(\frac{|x|}{\beta}\right)^{\alpha}\right),\\ \\
&\quad \quad \quad \quad \quad \quad  x \in \{{F_{geo}}, {F_{col}}\}, \label{equ:ggd}\end{aligned}
\end{equation}
where $\beta=\sigma \sqrt{\frac{\Gamma(1 / \alpha)}{\Gamma(3 / \alpha)}}$, $\Gamma(\alpha)=\int_{0}^{\infty} t^{\alpha-1} e^{-t} dt, \alpha>0$ is the gamma function, and the two estimated parameters ($\alpha,\beta^2$) indicate the shape and variance of the distribution. What's more, considering that the variance for normalized distribution is fixed as 1, we estimate the GGD parameters from distributions before normalization.  

\subsection{AGGD Parameters}
However, in some situations, the shape of the distribution is not symmetrical. We can see clearly from Fig. \ref{fig:sjtu-pcqa} that the reference planarity distribution represented by Fig. \ref{fig:sjtu-pcqa} (q) has a longer tail on the left while the planarity distribution represented by Fig. \ref{fig:sjtu-pcqa} (t) has a longer tail on the right. What's more, all types of distortions change weight of the tails for curvature and anisotropy distributions represented by Fig. \ref{fig:sjtu-pcqa} (f, g, h, j, k, l). Therefore, we utilize the general asymmetric generalized Gaussian distribution (AGGD) model to extract more detailed parameters from normalized feature distributions to describe the different spread extent in two directions:
\begin{equation}
\begin{aligned}
&AGGD\!\left(x ; v, \sigma_{l}^{2}, \sigma_{r}^{2}\right)\!=\!\left\{\begin{array}{ll}
\!\!\!\frac{v}{\left(\beta_{l}+\beta_{r}\right) \Gamma\left(\frac{1}{v}\right)} \exp \left(-\left(\frac{-x}{\beta_{l}}\right)^{v}\right)\! & \!\!\!x<0, \\
\!\!\!\frac{v}{\left(\beta_{l}+\beta_{r}\right) \Gamma\left(\frac{1}{v}\right)} \exp \left(-\left(\frac{x}{\beta_{r}}\right)^{v}\right)\! & \!\!\!x \geq 0,\\
\end{array}\right.\\
&\quad \quad \quad \quad \qquad \qquad x \in \{\hat{F_{geo}}, \hat{F_{col}}\},
\end{aligned}
\label{equ:aggd1}
\end{equation}
where $\beta_{l} =\sigma_{l} \sqrt{\frac{\Gamma\left(\frac{1}{v}\right)}{\Gamma\left(\frac{3}{v}\right)}},
\beta_{r} =\sigma_{r} \sqrt{\frac{\Gamma\left(\frac{1}{v}\right)}{\Gamma\left(\frac{3}{v}\right)}}, \eta=\left(\beta_{r}-\beta_{l}\right)$,  the parameter $v$ decides the shape of the distribution, $\sigma_{l}^{2}$ and $\sigma_{l}^{2}$ are scale parameters that refer to the spread extent on the left and right sides of the distribution respectively, $\eta$ is a parameter that can better fit the AGGD model stated in \cite{brisque}. Additionally, the AGGD can be recognized as the extension of GGD. When $\sigma_{l}^{2}= \sigma_{r}^{2}$, the AGGD turns into GGD. Finally, the four parameters ($\eta, v,\sigma_{l}^{2},\sigma_{l}^{2}$) are estimated to describe the  characteristics of asymmetric distributions.

\subsection{Gamma Parameters}
The reference curvature and anisotropy distributions shown in Fig. \ref{fig:sjtu-pcqa} (e, i) are similar to the shape of Gamma distribution. Compression, downsampling, and Gaussian noise all change the shape and scale of the corresponding distributions and such distributions still exhibit Gamma-like appearance. Therefore, we propose to use Gamma distribution parameters to quantify the distortions. The shape-rate Gamma model is formulated as:
\begin{equation}
\begin{aligned}
    Gamma(x ; \alpha, \beta)& =\frac{\beta^{\alpha} x^.{\alpha-1} e^{-\beta x}}{\Gamma(\alpha)} x>0,\\
    x & \in \{\hat{F_{geo}}, \hat{F_{col}}\},
\end{aligned}
    \label{equ:gamma}
\end{equation}
where $\alpha$ and $\beta$ stands for the shape and rate parameters and $\alpha, \beta>0$.

\begin{table*}[t]
\renewcommand\arraystretch{1.5}
\renewcommand\tabcolsep{3.0pt}
\setlength{\abovecaptionskip}{-5pt}
  \caption{Summary of features extracted in the proposed method}
  
  \label{tab:features}
  \begin{center}
  \begin{tabular}{cccccc}
    \toprule
    Format & Feature Domains & Feature ID & Feature Description  &  Computation\\
    \hline
    \multirow{4}{*}{Point cloud} 
    & ($Cur, Ani, Lin, Pla, Sph,L,A,B$) & $f_{p1}-f_{p16}$      & Mean and standard deviation for each feature domain    &  Eq.(\ref{equ:mean})  \\
    \cline{2-5}
    & ($Cur, Ani, Lin, Pla, Sph,L,A,B$) & $f_{p17}-f_{p24}$  & Entropy for each feature domain  &  Eq.(\ref{equ:entropy})    \\
    \cline{2-5}
    & ($Cur, Ani, Lin, Pla, Sph,L,A,B$) & $f_{p25}-f_{p40}$  &GGD ($\alpha,\beta^2$) for each feature domain  &  Eq.(\ref{equ:ggd})    \\
    \cline{2-5}
     & ($Cur, Ani, Lin, Pla, Sph,L,A,B$) & $f_{p41}-f_{p72}$  & AGGD ($\eta, v,\sigma_{l}^{2},\sigma_{l}^{2}$) for each normalized feature domain  &  Eq.(\ref{equ:aggd1})    \\
     \cline{2-5}
     & ($Cur, Ani, Lin, Pla, Sph,L,A,B$) & $f_{p73}-f_{p88}$  & Gamma ($\alpha,\beta$) for each normalized feature domain  &  Eq.(\ref{equ:gamma})    \\
     \hline
     
    \multirow{4}{*}{Mesh} 
    & ($Cur, Dih, Far, Fan, L, A, B$) & $f_{m1}-f_{m14}$      & Mean and standard deviation for each feature domain    &  Eq.(\ref{equ:mean})  \\
    \cline{2-5}
    & ($Cur, Dih, Far, Fan, L, A, B$) & $f_{m15}-f_{m21}$  & Entropy for each feature domain  &   Eq.(\ref{equ:entropy})    \\
    \cline{2-5}
    & ($Cur, Dih, Far, Fan, L, A, B$) & $f_{m22}-f_{m35}$  &GGD ($\alpha,\beta^2$) for each feature domain  &  Eq.(\ref{equ:ggd})    \\
    \cline{2-5}
     & ($Cur, Dih, Far, Fan, L, A, B$) & $f_{m36}-f_{m63}$  & AGGD ($\eta, v,\sigma_{l}^{2},\sigma_{l}^{2}$) for each normalized feature domain  &   Eq.(\ref{equ:aggd1})    \\
     \cline{2-5}
     & ($Cur, Dih, Far, Fan, L, A, B$) & $f_{m64}-f_{m77}$  & Gamma ($\alpha,\beta$) for each normalized feature domain  & Eq.(\ref{equ:gamma})    \\
    \bottomrule
  \end{tabular}
  \end{center}
\end{table*}

\subsection{Parameters Summary}
In summary, we collect the average, standard deviation, and entropy values in the 3D model feature domains as the fundamental features. Then we use 3 representative distribution models to estimate 8 statistical parameters, including GGD ($\alpha,\beta^2$), AGGD ($\eta, v,\sigma_{l}^{2},\sigma_{l}^{2}$), Gamma ($\alpha,\beta$), for all feature domains. Finally, considering that the colored point cloud has 8 feature domains ($Cur, Ani, Lin, Pla, Sph, L,A,B$) and the colored mesh has 7 feature domains ($Cur, Dih, Far, Fan, L, A, B$), 88 ($8\times11$) features are computed for a single colored point cloud and 77 ($7\times11$) features are computed for a single colored mesh respectively. The summary of features extracted in the proposed method is listed in Table \ref{tab:features}.

\section{Experiment Evaluation}
\label{sec:experiment}

\subsection{Regression Model}
After the feature extraction process, a feature vector is obtained to describe the characteristics of the 3D model. In our experiment, we propose to use the support vector machine regressor (SVR) as the regression model, which is a common and effective choice to handle high dimensional data in previous quality assessment research \cite{nr-svr} \cite{brisque}. We employ the standard normalization as the pre-processing to scale the features. Then the feature vector can be integrated into a quality score for evaluation. We employ the Python sklearn package \cite{sklearn} to implement the radial basis function (RBF) kernel SVR model with default settings.

\subsection{Experiment Setup}
\subsubsection{Experiment Setup for PCQA}
 To test the performance of the proposed method, we employ the subjective point cloud assessment database (SJTU-PCQA) \cite{sjtu-pcqa} and the Waterloo point cloud assessment database (WPC) proposed in \cite{su2021perceptual}. 
 
 The SJTU-PCQA database provides 420 point cloud samples distorted from 10 reference point clouds. Each reference point cloud is distorted with seven types of common distortions with six levels. Unfortunately, only 9 reference point clouds and their corresponding distorted point cloud samples are now available to the public, thus we can obtain 378 ($9\times6\times7$) point cloud samples for the experiment.
Since the proposed approach requires a training procedure to calibrate the SVR model and to avoid the influence of content overlap, we select 8 of the 9 groups of point clouds as training set and leave the rest 1 group as testing set. In order to ensure the validity of the results, we exhaustively list all the $C_{9}^8$=9 database separations for the experiment and use the average performance as the final experimental results. In addition, the MOSs collected in the SJTU-PCQA database are divided by 10 in the training process to scale the MOSs to [0,1]. The predicted scores are re-scaled for validation in the testing process. 

{The WPC database includes 20 high-quality source point clouds and creates 740 distorted point clouds using downsampling, Gaussian noise, and three types of compression. Specifically, we maintain the same training set and testing set split as stated in \cite{liu2021pqa}. Similarly, the MOSs provided in the WPC database are divided by 100 in the training process and the predicted scores are re-scaled for validation in the testing process.
The comparison PCQA metrics can be categorized into two types:}

\begin{itemize}
    \item  Image-based metrics: These metrics evaluate the quality of 3D models by assessing the quality of the corresponding 2D projections. Please refer to \cite{sjtu-pcqa} for detailed projection process and we use only RGB projections. The FR image-based metrics include PSNR, SSIM \cite{ssim}, and PB-PCQA\cite{sjtu-pcqa}. The NR image-based metrics include NIQE \cite{niqe}, BRISQUE \cite{brisque}, and PQA-net \cite{liu2021pqa}.

    \item Model-based metrics: Full-reference metrics operate directly from the 3D model, which include GraphSIM \cite{yang2020inferring}, PointSSIM \cite{pcqa3}, PCQM \cite{pcqm}, and ResCNN \cite{pcqa-large-scale}. Reduced-reference metric includes PCMRR \cite{viola2020reduced}. Specifically, ResCNN is a model-based deep-learning approach. 
\end{itemize}

\subsubsection{Experiment Setup for MQA}
The MQA method proposed in this paper is validated on the color mesh distortion measure (CMDM) database \cite{database}. The database is generated from 5 source models subjected to geometry and color distortions. Then the source models are corrupted with 4 types of distortions based on color and geometry and each type of distortion is adjusted with 4 different strengths. The $Aix$, $Chamleon$, $Fish$, and $Samurai$ models are selected for training while the $Ari$ model is chosen for testing. Specifically, each distorted model is provided with 5 subjective scores according to its viewpoints and animation types. For simplification, we use the average of the 5 subjective scores as the final quality score for the distorted model.

In the literature, few metrics are proposed to deal with the 3D-QA tasks of color meshes. In order to evaluate the performance of the proposed method, some image-based metrics that might be able to predict the visual quality of 3D meshes are utilized as competitors. Unfortunately, few no-reference quality assessment metrics for 3D meshes are open-sourced, thus we try to reproduce some of the metrics. Specifically, the metrics used for comparison can be divided into two types: 
\begin{itemize}
    \item  Image-based metrics: These metrics operate by evaluating the quality of 2D images rendered from the 3D colored meshes. The FR image-based metrics include PSNR and SSIM \cite{ssim}. The NR image-based metrics include NIQE \cite{niqe} and BRISQUE \cite{brisque}.
    \item Model-based metrics: The FR-MQA metric includes CMDM \cite{database}. The NR-MQA metrics designed especially for geometry-only 3D meshes include NR-SVR \cite{nr-svr}, NR-GRNN \cite{nr-grnn}, NR-CNN \cite{nr-cnn}.
\end{itemize}

\subsection{Evaluation Criterion}
Four mainstream consistency evaluation criteria are utilized to compare the correlation between the predicted scores and MOSs, which include Spearman Rank Correlation Coefficient (SRCC), Kendall’s Rank Correlation Coefficient (KRCC), Pearson Linear Correlation Coefficient (PLCC), and Root Mean Squared Error (RMSE).
An excellent model should obtain values of SRCC, KRCC, and PLCC close to 1, and the value of RMSE near 0.

\begin{table*}[t]
\renewcommand\arraystretch{1.4}
\renewcommand\tabcolsep{8pt}
\setlength{\abovecaptionskip}{-5pt}
  \caption{ Performance comparison with competitors on the SJTU-PCQA and WPC databases.}
  \vspace{-0.05cm}
  \begin{center}
  \begin{tabular}{c|c|c|c|cccc|cccc}
    \toprule
    \multirow{2}{*}{Ref} & \multirow{2}{*}{Type} &\multirow{2}{*}{Index} & \multirow{2}{*}{Metric} & \multicolumn{4}{c|}{SJTU-PCQA} & \multicolumn{4}{c}{WPC} \\ \cline{5-12}
     &&&& PLCC &  SRCC & KRCC & RMSE & PLCC &  SRCC & KRCC & RMSE \\
    \hline
    \multirow{6}{*}{FR} & \multirow{3}{*}{Image-based}
    &A&PSNR  & 0.2317 & 0.2422 & 0.1077 & 2.3124 & 0.4872 & 0.4235 & 0.3080 & 15.8133\\
    &&B&SSIM  & 0.3476 & 0.2987 & 0.1919  & 21770 & 0.4944 & 0.3878 &0.3234 & 15.7749 \\
    &&C &PB-PCQA  &0.6076 & 0.6020 & - & 1.8635  &-&-&-&- \\
    \cline{2-12}
    & \multirow{3}{*}{Model-based}
    &D&GraphSIM  &0.8449 & 0.8483 & 0.6448 & 1.5721 &0.6163 & 0.5831 & 0.4194 & 17.1939\\
    &&E &PointSSIM  &0.7136 & 0.6867 & 0.4964  & 1.7001 &0.4667 & 0.4542 & 0.3278  & 20.2733\\
    &&F &PCQM  &\textbf{0.8653} & \textbf{0.8544} & \textbf{0.6586}  & \textbf{1.2162} & \textbf{0.7499} & \textbf{0.7434} & \textbf{0.5601}  & \textbf{15.1639}\\
    \hline
    \multirow{1}{*}{RR} & \multirow{1}{*}{Model-based}
     &G&PCMRR  &0.6101 & 0.4816 & 0.3362  & 1.9342 &0.3433 & 0.3097 & 0.2082  & 21.5302\\
     \hline
    \multirow{5}{*}{NR} & \multirow{2}{*}{Image-based}
    &H&NIQE  &0.3764 & 0.2214 &0.1512  & 2.2671 &0.3957 &0.3887 &0.2551  & 22.5502\\
    
    & &I&BRISQUE  &0.2241 & 0.2051 & 0.1121 & 2.2428 &0.4176 & 0.3781 & 0.2444 & 22.5414\\
    & &J&PQA-Net  &-&-&-&- & \textbf{0.7000} & \textbf{0.6900} & \textbf{0.5100}  & \textbf{15.1800}\\
    \cline{2-12} & \multirow{2}{*}{Model-based}
    &K&ResCNN   & 0.5975 & 0.6187 & - & - &-&-&-&- \\
    & &L&Proposed  & \textbf{0.7382} & \textbf{0.7144} & \textbf{0.5174} & \textbf{1.7686} & 0.6514 & 0.6479 & 0.4417 & 16.5716\\
    \bottomrule
  \end{tabular}
  \end{center}
  \label{tab:pcqa}
  \vspace{-0.7cm}
\end{table*}

\begin{figure*}[!tbp]
\centering
\subfigure[\textit{}]{
\begin{minipage}[t]{0.15\linewidth}
\centering
\includegraphics[width = 2.6cm]{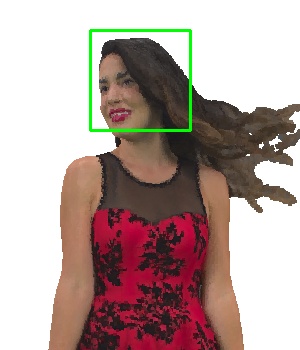}
\end{minipage}%
}%
\subfigure[\textit{}]{
\begin{minipage}[t]{0.16\linewidth}
\centering
\includegraphics[width = 2.6cm]{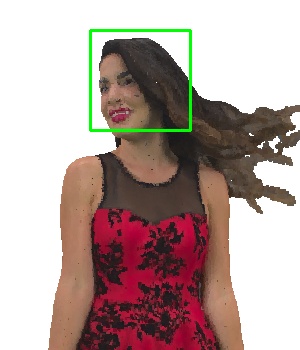}
\end{minipage}%
}%
\subfigure[\textit{}]{
\begin{minipage}[t]{0.16\linewidth}
\centering
\includegraphics[width = 2.6cm]{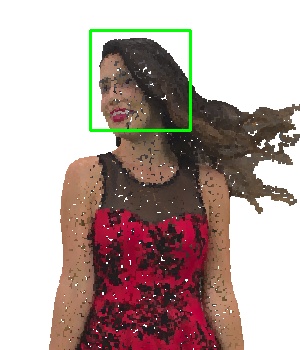}
\end{minipage}%
}%
\subfigure[\textit{}]{
\begin{minipage}[t]{0.16\linewidth}
\centering
\includegraphics[width =2.6cm]{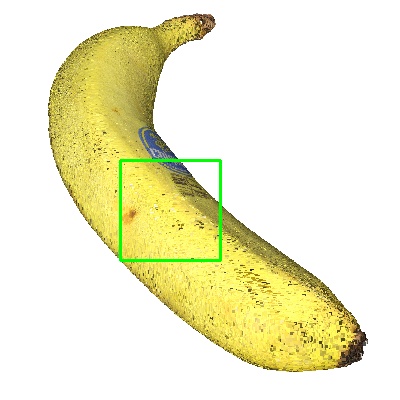}
\end{minipage}%
}%
\subfigure[\textit{}]{
\begin{minipage}[t]{0.16\linewidth}
\centering
\includegraphics[width = 2.6cm]{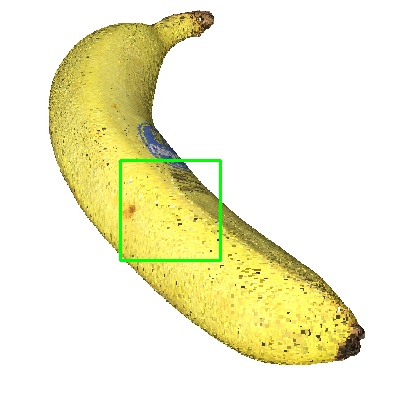}
\end{minipage}%
}%
\subfigure[\textit{}]{
\begin{minipage}[t]{0.16\linewidth}
\centering
\includegraphics[width =2.6cm]{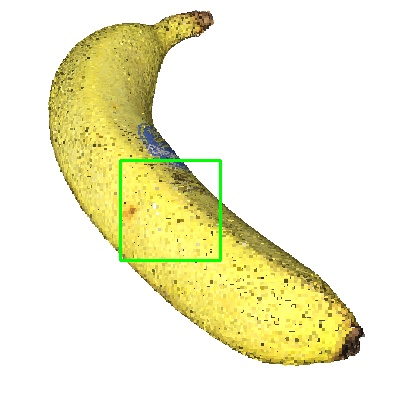}
\end{minipage}%
}%

\caption{Illustration of point clouds with consistently increasing downsampling distortion levels from the SJTU-PCQA \cite{sjtu-pcqa} and the WPC \cite{su2021perceptual} databases. (a), (b), and (c) are the snapshots of \textit{rednandblack\_15}, \textit{rednandblack\_16}, and \textit{rednandblack\_17} from the SJTU-PCQA database. (d), (e), and (f) are the snapshots of \textit{banana\_level\_9}, \textit{banana\_level\_8}, and \textit{banana\_level\_7} from the WPC database respectively. }
\label{redandblack}
\end{figure*}

\subsection{Performance Discussion}
\subsubsection{The performance of PCQA validated on the SJTU-PCQA and WPC Databases}
{ The experimental results for PCQA on the SJTU-PCQA database and the WPC database are shown in Table \ref{tab:pcqa}. The best performance results of the NR-PCQA methods along with the best performance results of all methods are marked in bold. We can see clearly that the FR-PCQA metric PCQM outperforms other PCQA metrics. With the assistance of reference information, FR-PCQA methods tend to be more effective at predicting the quality levels of colored point clouds than NR-PCQA methods. The proposed method achieves first place on the SJTU-PCQA database and second place on the WPC database among the compared NR-PCQA metrics respectively. It's worth mentioning that the PQA-Net extracts features from the rendering views with deep-learning networks. The proposed method depends on handcrafted features and remains competitive considering the computational resource consumption of deep-learning networks.

\begin{table}[!t]
\setlength{\abovecaptionskip}{-5pt}
\renewcommand\arraystretch{1.5}
\renewcommand\tabcolsep{3.4pt}
  \caption{Performance comparison with competitors on the CMDM database.}
  \vspace{-0.05cm}
  \begin{center}
  \begin{tabular}{c|c|c|c|cccc}
    \toprule
    Ref & Type &Index & Metric & PLCC  &  SRCC & KRCC &RMSE\\
    \hline
    \multirow{3}{*}{FR} & \multirow{2}{*}{Image-based} 
    &A&PSNR & 0.7672 & 0.7735 & 0.7280 & 0.8832\\
    &&B&SSIM & 0.7944 & 0.7817 &0.7000 & 0.9656\\
    \cline{2-8} & \multirow{1}{*}{Model-based} 
    &C&CMDM & \textbf{0.9130} & \textbf{0.9000} & - & -\\
    \hline
    \multirow{6}{*}{NR}  & \multirow{2}{*}{Image-based} 
    &D&NIQE  &0.4059 & 0.4768 & 0.3000 & 1.3352\\
    & &E&BRISQUE  & 0.5786 &0.4882 & 0.3598 & 1.2237\\
    \cline{2-8} & \multirow{4}{*}{Model-based} 
    &F&NR-SVR  &0.6082 & 0.4489 & 0.3420 & 1.3147\\
    & &G&NR-GRNN  & 0.6599 & 0.6948 & 0.5130 & 1.1121\\
    & &H&NR-CNN  &0.5204 & 0.5022 & 0.3420 & 1.2804\\
   &  &I&Proposed  & \textbf{0.8626} & \textbf{0.8754} & \textbf{0.7222} & \textbf{0.6062}\\
    \bottomrule
  \end{tabular}
  \end{center}
  \label{tab:mqa}
  \vspace{-0.7cm}
\end{table}

With closer observations, we can make several more detailed analyses: 1-1) Image-based metrics using simple IQA models (such as PSNR and SSIM) are less effective. This is because the rendering views captured from various viewpoints are quite different in content. To get more quality information, more snapshots have to be sampled, which may confuse such IQA models because they are not capable of dealing with multiple pairs of images with dissimilar contents. Thanks to the strong learning ability of CNN, PQA-Net overcomes the difficulty and achieves high performance by utilizing the correlation of the rendering views. To conclude, we think that increasing the sample number of snapshots and optimizing learning models are the focus of image-based methods in the future. However, such methods inevitably require large amounts of computation resources and highly depend on the scale of the database.  1-2) Most model-based methods achieve lower performance on the WPC database than on the SJTU-PCQA database. We attempt to give the reasons. First, although the two databases manually introduce similar types of distortions, the WPC database employs some point cloud samples that are less sensitive to distortions in perceived visual quality.
For example, as can be seen in Fig. \ref{redandblack}, the \textit{rednandblack} sample from the SJTU-PCQA database appears to be more distinct under the influence of downsampling distortions. However, the \textit{banana} sample from the WPC database is less complex in geometry structure and limited in colorfulness, which makes the distortion difference less obvious. Second, the SJTU-PCQA database includes distorted point clouds with mixed distortions while the WPC database only adds one type of distortion to a single point cloud. It seems that point clouds with mixed distortions are more distinguishable in quality under close distortion levels. What's more, the WPC database has twice as many reference point clouds as the SJTU-PCQA database. The expanded scale and content diversity challenge the generalization ability of such methods as well.}


\begin{figure}[tbp]
\centering
\subfigure[\textit{}]{
\begin{minipage}[t]{0.45\linewidth}
\centering
\includegraphics[width = 4cm]{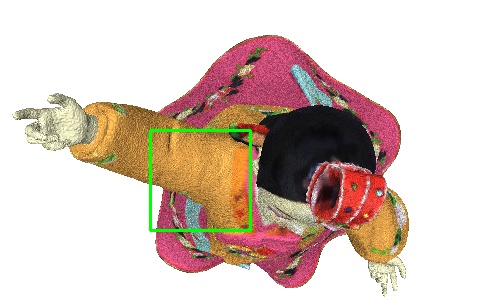}
\end{minipage}%
}%
\subfigure[\textit{}]{
\begin{minipage}[t]{0.45\linewidth}
\centering
\includegraphics[width = 4cm]{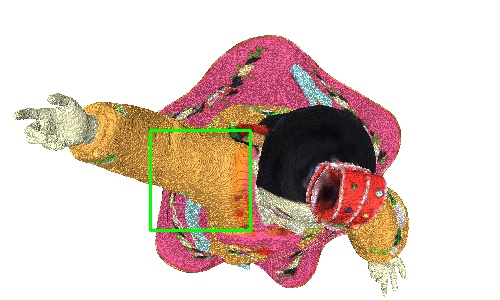}
\end{minipage}%
}%
\\
\subfigure[\textit{}]{
\begin{minipage}[t]{0.45\linewidth}
\centering
\includegraphics[width = 4cm]{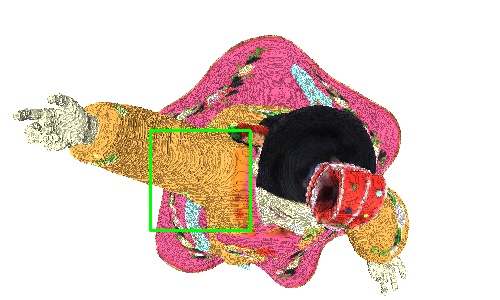}
\end{minipage}%
}%
\subfigure[\textit{}]{
\begin{minipage}[t]{0.45\linewidth}
\centering
\includegraphics[width = 4cm]{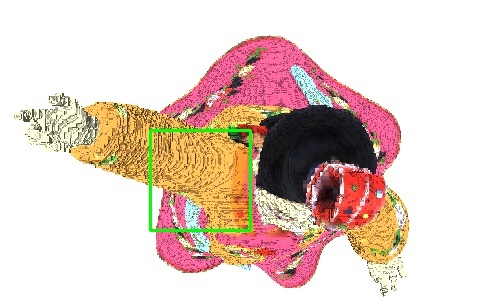}
\end{minipage}%
}%

\caption{Examples of meshes with increasing geometry quantization distortion levels from the CMDM database \cite{database}. (a), (b), (c), and (d) are the snapshots of \textit{Ari\_QuantGeom\_1}, \textit{Ari\_QuantGeom\_2}, \textit{Ari\_QuantGeom\_3}, and \textit{Ari\_QuantGeom\_4} respectively. }

\label{fig:ari}
\vspace{-0.5cm}
\end{figure}

\subsubsection{The performance of MQA validated on the CMDM Database}
{Table \ref{tab:mqa} presents the experimental results on the CMDM database, in which the top performance results of the NR-PCQA methods along with the best performance results of all methods are marked in bold. Surprisingly, the FR image-based metrics get remarkable performance results. It is because the CMDM database manually adjusts the employed 4 strengths of distortions to cover most range of visual quality levels (from imperceptible levels to high levels of impairment), which makes the distortion levels more coarse-grained. As illustrated in Fig. \ref{fig:ari}, the degradations caused by the 4 distortion strengths are quite obvious, thus enabling such simple methods to gain competitive performance. 

The NR model-based MQA methods such as NR-SVR, NR-GRNN, and NR-CNN have a stronger ability to predict the visual quality than the NR image-based MQA methods, but they do not take color information into consideration, thus resulting in lower performance than the proposed method. The FR-MQA metric CMDM achieves first place in PLCC and SRCC and the proposed method obtains second place in PLCC and SRCC with a slight gap. The reason is that the coarse-grained levels of distortions make it easier for the proposed method to learn the relationship between the corresponding 3D-NSS parameters and quality scores. Therefore, even with relatively little training data, the proposed method still obtains good performance.}

\begin{table*}[t]
\renewcommand\arraystretch{1.5}
\renewcommand\tabcolsep{5.5pt}
\setlength{\abovecaptionskip}{-5pt}
  \caption{ Performance of the ablation study. The best performance for each database is marked in bold.}
  
  \label{tab:ablation}
  \begin{center}
  \begin{tabular}{c|c|cccc|cccc|cccc}
    \toprule
    \multirow{2}{*}{Type}  & \multirow{2}{*}{Feature}  & \multicolumn{4}{c|}{SJTU-PCQA} & \multicolumn{4}{c|}{WPC} & \multicolumn{4}{c}{CMDM} \\
    \cline{3-14}
    &  &PLCC & SRCC & KRCC & RMSE & PLCC & SRCC & KRCC & RMSE & PLCC & SRCC & KRCC & RMSE \\
    \hline
    \multirow{5}{*}{Geometry} & F1 & 0.6406 & 0.6171 & 0.5283 & 1.8825 & 0.4280 & 0.4219 & 0.2801 & 19.5704 & 0.6214  &0.4162  &0.2852 & 0.9288 \\
    &F2 & 0.6082 & 0.5837 & 0.3963 & 1.8989 & 0.5042 & 0.4653 & 0.2992 & 18.7064 & 0.5872  &0.4355  &0.3022 & 1.0861 \\
    &F3 & 0.6259 & 0.5997 & 0.4166 & 1.9205 & 0.3874 & 0.3714 & 0.2350 & 20.4081 & 0.5656  &0.3333  &0.2481 & 1.0053 \\
    &F4 & 0.5187 & 0.4216 & 0.2713 & 1.9892 & 0.3889 & 0.3838 & 0.2761 & 21.2468 & 0.5358  &0.4163  &0.3096 & 1.0692 \\ 
    & (F1\~{}F4)  & 0.7314  & 0.6937 & 0.5067 & 1.8096 & 0.5889  & 0.5827 & 0.4186 & 17.8330 & 0.6561 & 0.5076 & 0.3954 & 0.9267 \\\hline
    \multirow{5}{*}{Color}
    &F5 & 0.1897 & 0.1250 & 0.0840 & 2.5758 & 0.1436 & 0.1327 & 0.0900 & 26.3404 & 0.1245  &0.1101  &0.0807 & 1.2254 \\
    &F6 & 0.2755 & 0.2170 & 0.1548 & 2.3927 & 0.1548 & 0.1009 & 0.0708 & 25.1412 & 0.0074  & 0.0291  & 0.0171 & 1.2582 \\
    &F7  & 0.0097 & 0.0069 & 0.0042 & 2.6471 & 0.1557 & 0.1836 & 0.1242 & 27.9339 & 0.1090  &0.0793  &0.0574 & 1.2556 \\
    &F8 & 0.0004 & 0.0012 & 0.0001 & 2.5288 & 0.1839 & 0.1488 & 0.0906 & 21.9701 & 0.1281  &0.1860  &0.0064 & 1.2617 \\
    &(F5\~{}F8)  & 0.2351 & 0.2408 & 0.1685 & 2.4511 & 0.2220 & 0.1857 & 0.1599 & 24.3489 & 0.2985 & 0.1820  & 0.2283 & 1.2469\\ \hline
    Basic & (F1+F5)  & \textbf{0.7522} & \textbf{0.7449} & \textbf{0.5454}  & \textbf{1.7004} & 0.5335 & 0.5121 & 0.3647  & 21.4647 & {0.8481} & {0.8530}  & {0.6955} & {0.6079}\\
    GGD & (F2+F6)  & 0.7173  & 0.6973 & 0.5035 & 1.7781 & 0.6172  & 0.5968 & 0.4353 & 17.1555 & 0.7819  & 0.7392 & 0.5821 & 0.7646\\
    AGGD & (F3+F7)  & 0.6106 & 0.6122 & 0.4337 & 2.0208 & 0.4268 & 0.4243 & 0.2921 & 20.9246 & 0.8342 & 0.7974 & 0.6456 & 0.6989\\
    Gamma & (F4+F8)  & 0.5841 & 0.4999 & 0.3504 & 2.0469 & 0.4584 & 0.4516 & 0.3177 & 19.8646 & 0.5982 & 0.5262 & 0.3951 & 0.9956\\ \hline
    
    
    \multirow{2}{*}{NSS} & 2D & 0.3604 & 0.2123 & 0.1613 &2.3125 & 0.4019 &0.3919 & 0.2337 & 22.5500 & 0.5616 &0.4772 & 0.2988 &1.2347\\
    &2D+3D &0.4122 & 0.3344 & 0.2951 &2.2005 & 0.4658 & 0.4452 & 0.2500 & 21.2028 & 0.6658 & 0.6652 & 0.3700 & 0.9981\\\hline
    All & F1\~{}F8 & 0.7382 & 0.7144 & 0.5174 & 1.7686 & \textbf{0.6514} & \textbf{0.6479} & \textbf{0.4417} & \textbf{16.5716} & \textbf{0.8626} & \textbf{0.8754} & \textbf{0.7222} & \textbf{0.6062}\\
    \bottomrule
  \end{tabular}
  \end{center}
  \vspace{-5pt}
\end{table*}

\subsection{Ablation Study}

{
To further test the effectiveness and contributions of different types of features (color and geometry) and different statistical parameters, we split the features into 8 groups: (1) F1: mean, standard deviation, and entropy values of geometry feature domains; (2) F2: GGD parameters of geometry feature domains; (3) F3: AGGD parameters of geometry feature domains; (4) F4: Gamma distribution parameters of geometry feature domains; (5) F5: mean, standard deviation, and entropy values of color feature domains; (6) F6: GGD parameters of color feature domains; (7) F7: AGGD parameters of color feature domains; (8) F8: Gamma distribution parameters of color feature domains. Then we can analyze the features' contributions by obtaining the performance of different combinations of feature groups. For example, the F2+F6 model only contains AGGD parameters of both geometry and color feature domains while the F1\~{}F4 model only consists of statistical parameters of geometry feature domains.

Besides, most 3D models are perceived by humans only through the rendered 2D views. To find out whether the information of 2D views would help improve the performance of 3D-NSS, we test the effectiveness of the combined 2D-NSS and 3D-NSS features as well. The 2D-NSS features are extracted by the 2D views of objects using BRISQUE \cite{brisque} and NIQE \cite{niqe} (The redundant features of BRISQUE and NIQE are removed ). The features extracted from different viewpoints are averaged to get the final 2D-NSS features. The proposed all groups' features in this paper are employed as the 3D-NSS features. We simply concatenate the 2D-NSS and 3D-NSS features as the quality-aware feature vector.

The performance results of the ablation study are shown in Table \ref{tab:ablation}, where the mean, standard deviation, and entropy values of geometry and color feature domains are represented as the basic type of features (F1+F5). It can be obviously observed that the geometry features contribute more to the final quality score on all three databases. It might be because all three databases introduce more types of geometry distortions than color distortions and the geometry information weighs more in the human perception of 3D models.

By analyzing the different performances of different models, we can make several observations. 1) On the SJTU-PCQA database, we can see that the mean, standard deviation, and entropy values obtain the best performance, even higher than the proposed all groups. While on the WPC database, the all groups model is significantly superior to the separate distribution models. We attempt to give our explanation as follows. The ability of simple parameters to describe the shape of distributions is limited. For example, some distributions with very different shapes may have the same mean and variance. When the diversity and number of samples are not saturated, simpler parameters may achieve better results (more complex parameters are troubled by the redundant information), but when the diversity and number of samples increase, these simpler parameters are less effective, which requires the help of more complex parameters to describe the distributions. Therefore, we suggest using the proposed all groups model for practical application since such model is more capable of describing and distinguishing different distributions. 2) It seems that the 2D-NSS is not effective for 3D-QA tasks. We try to give the reason for such phenomenon as well. Single rendering 2D views contain limited information, therefore, we have to use multiple viewpoints to cover more quality information. However, the contents of snapshots from various viewpoints are quite different, such as the top and bottom views of the 3D models. The contents difference may have a much greater influence on the distributions rather than the distortions. }

\begin{figure*}[htbp]
\centering
\subfigure[SJTU-PCQA]{
\begin{minipage}[t]{0.3\linewidth}
\centering
\includegraphics[width = 4.8cm]{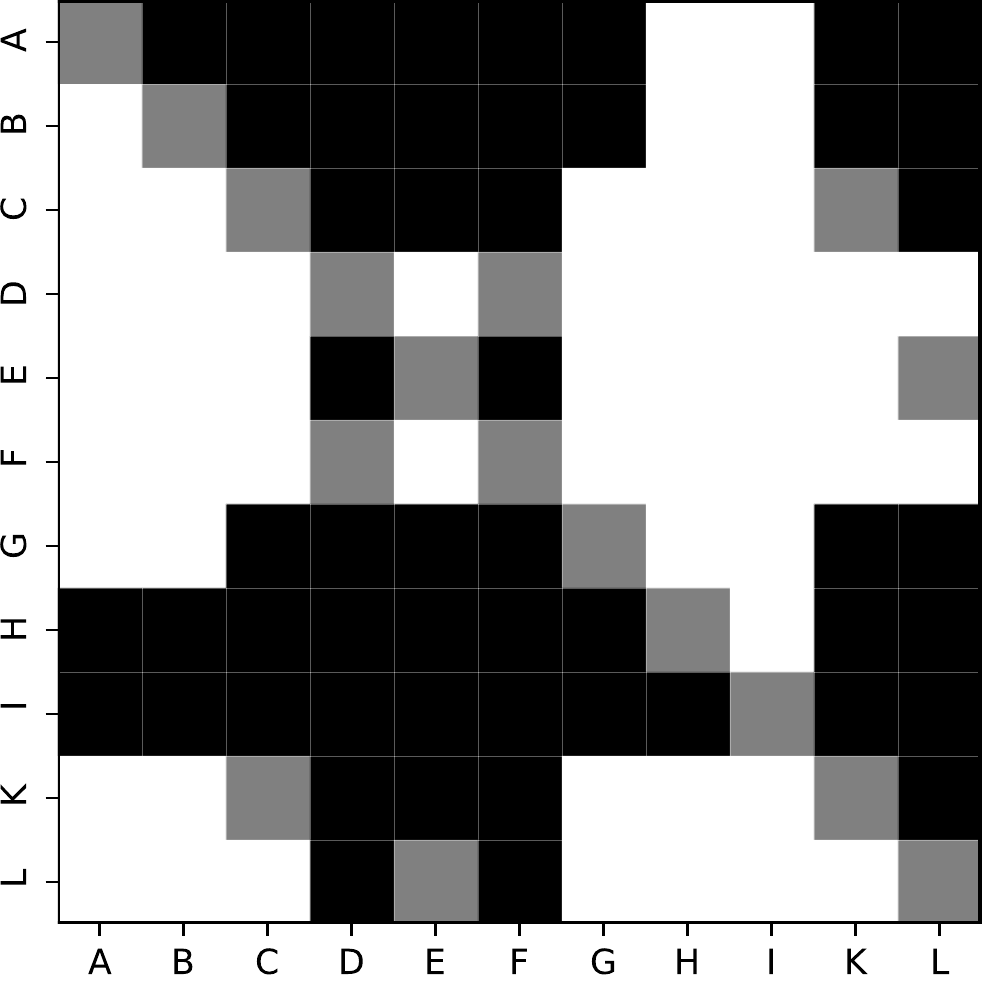}
\end{minipage}%
}%
\subfigure[WPC]{
\begin{minipage}[t]{0.33\linewidth}
\centering
\includegraphics[width = 4.8cm]{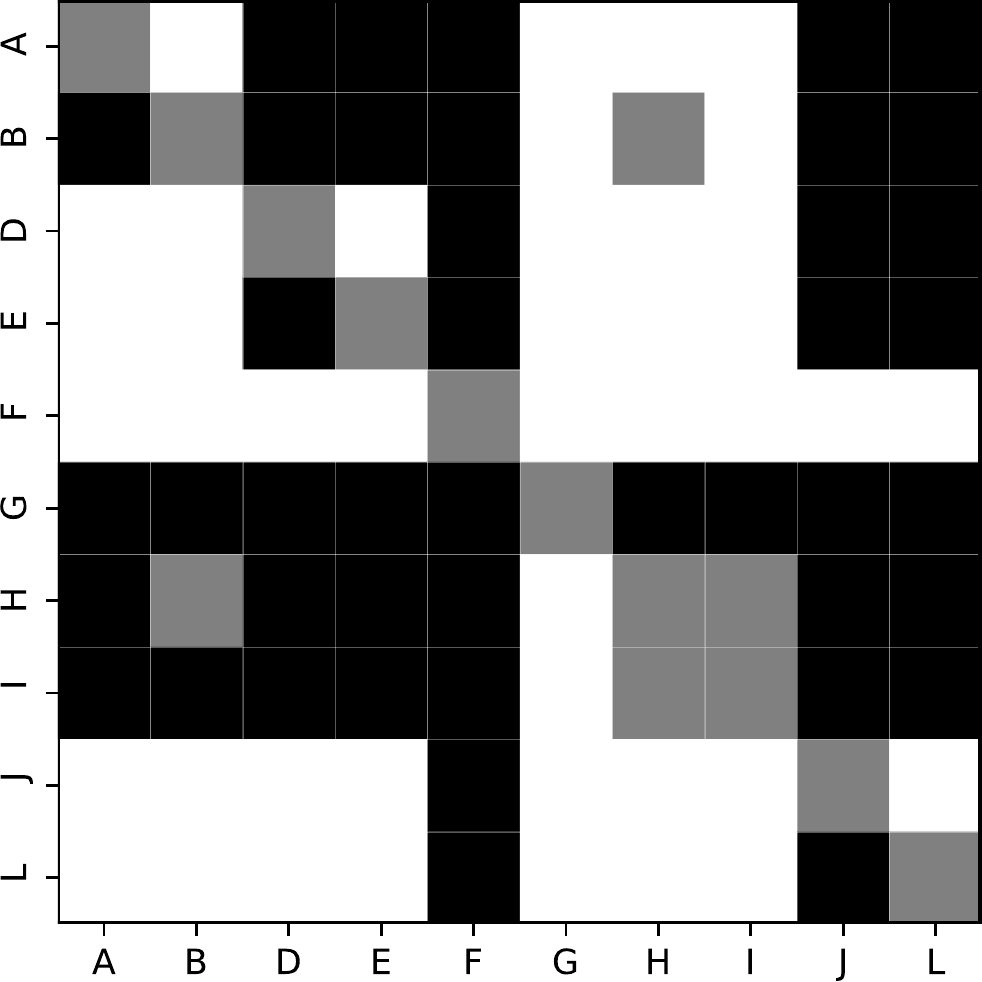}
\end{minipage}%
}%
\subfigure[CMDM]{
\begin{minipage}[t]{0.3\linewidth}
\centering
\includegraphics[width = 4.8cm]{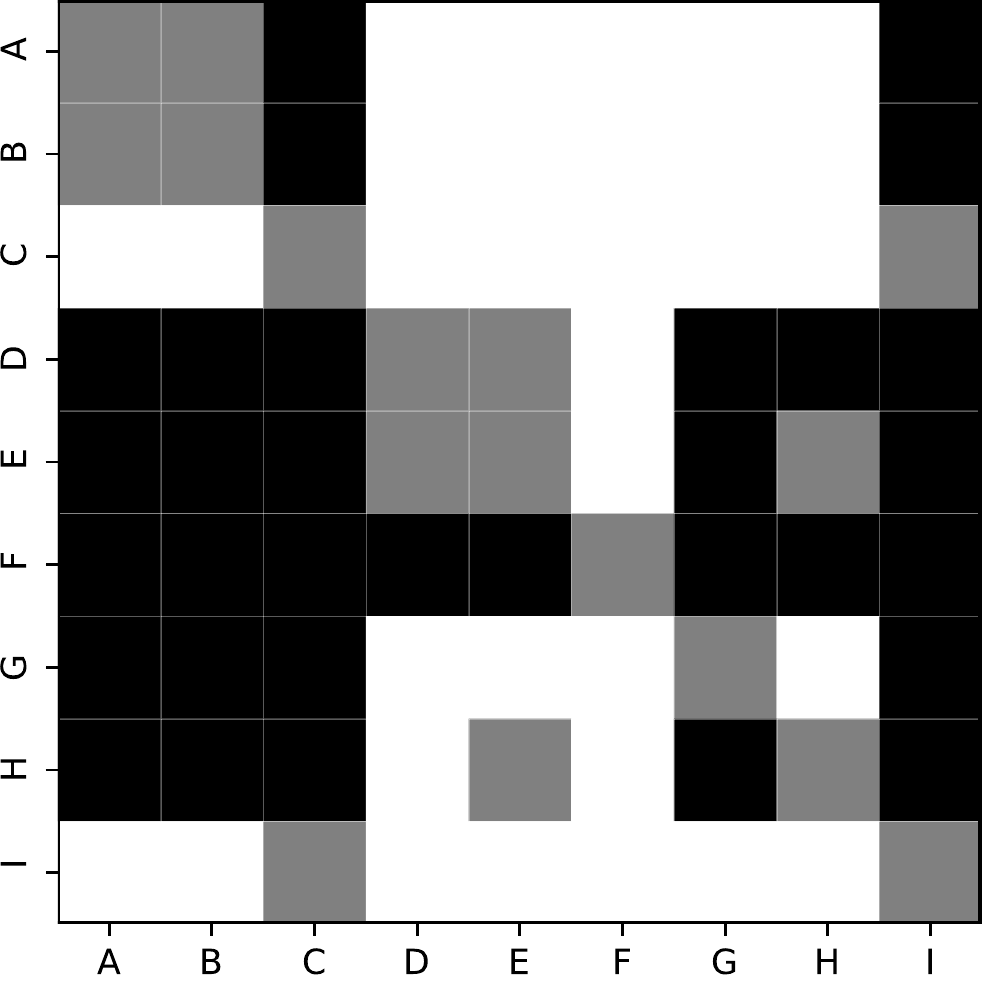}
\end{minipage}%
}%
\caption{Statistical test results of the proposed method and compared metrics on the SJTU-PCQA, WPC, and CMDM databases. A black/white block means the row method is statistically worse/better than the column one. A gray block means the row method and the column method are statistically indistinguishable. The metrics are denoted by the same index as in Table \ref{tab:pcqa} and Table \ref{tab:mqa} respectively.}
\label{heatmap}
\vspace{-0.2cm}
\end{figure*}

\begin{table*}[t]
\renewcommand\arraystretch{1.5}
\setlength{\abovecaptionskip}{-5pt}
  \caption{ Performance of the data-sensitivity experiment. The best performance for each database is marked in bold.}
  
  \label{tab:param}
  \begin{center}
  \begin{tabular}{c|cccc|cccc|cccc}
    \toprule
    \multirow{2}{*}{Criteria}  & \multicolumn{4}{c|}{SJTU-PCQA}  & \multicolumn{4}{c|}{WPC}  &\multicolumn{4}{c}{CMDM} \\
    \cline{2-13}
    & 20\% & 40\% & 60\% & 80\% & 20\% & 40\% & 60\% & 80\% & 20\% & 40\% & 60\% & 80\% \\
    \hline
    PLCC & 0.6009 & 0.6458 & 0.6655 & \textbf{0.7282} & 0.5303 & 0.5575 & 0.5702 & \textbf{0.6514} & 0.6935 & 0.7464 & 0.8178   & \textbf{0.8626} \\
    SRCC & 0.5851 & 0.6260 & 0.6462 & \textbf{0.7144} & 0.5208 & 0.5187 & 0.5542 & \textbf{0.6479} & 0.5807 & 0.6375 & 0.7422   & \textbf{0.8754} \\
    KRCC & 0.4083 & 0.4402 & 0.4574 & \textbf{0.5174} & 0.3611 & 0.3655 & 0.3938 & \textbf{0.4417} & 0.4152 & 0.4682 & 0.5680   & \textbf{0.7222} \\
    RMSE & 2.0708 & 1.9482 & 1.9018 & \textbf{1.7686} & 19.2921 & 18.9968 & 18.1487 & \textbf{16.5176} & 0.9064 & 0.8157 & 0.7208   & \textbf{0.6062} \\
    \bottomrule
  \end{tabular}
  \end{center}
  \vspace{-0.4cm}
\end{table*}

\subsection{Statistical Test}
To further analyze the performance of the proposed method, we conduct the statistical test in this section. We follow the same experiment setup as in \cite{statistic-test} and compare the difference between the predicted quality scores with the subjective ratings. All possible pairs of models are tested and the results are listed in Fig. \ref{heatmap}. 
It can be seen that our method is significantly superior to 7 compared PCQA metrics on the SJTU-PCQA and the WPC databases while our method also significantly outperforms 7 compared MQA metrics on the CMDM database. Specifically, the FR metric PCQM achieves significantly better performance than our method on both SJTU-PCQA and WPC databases, the FR metric CMDM is insignificantly distinguishable from our method on the CMDM database.

\subsection{Data-sensitivity Experiment}
{ To find out the influence of the number of training samples, we conduct the data-sensitivity experiment by changing the proportion of training set as about 20\%, 40\%, 60\%, and 80\% (2, 4, 6, 8 sample groups for training on the SJTU-PCQA database, 4, 8, 12, 16 for training on the WPC database, and 1, 2, 3, 4 samples for training on the CMDM database correspondingly). The experimental results are exhibited in Table \ref{tab:param}, from which we can find that increasing the number of training samples is helpful to improve the performance of the proposed method. The CMDM database employs coarse-grained distortion levels and includes relatively small amounts of mesh samples, thus the increase of training data improves the performance more significantly. For the PCQA database, we think using fewer training samples may exaggerate the impact of noisy labels and result in the over-fit of the model, which causes the performance to drop. In all, increasing the diversity and number of the training samples are beneficial for obtaining more robustness and higher performance. }


\begin{figure}
    \centering
    \includegraphics[width = 8.8cm]{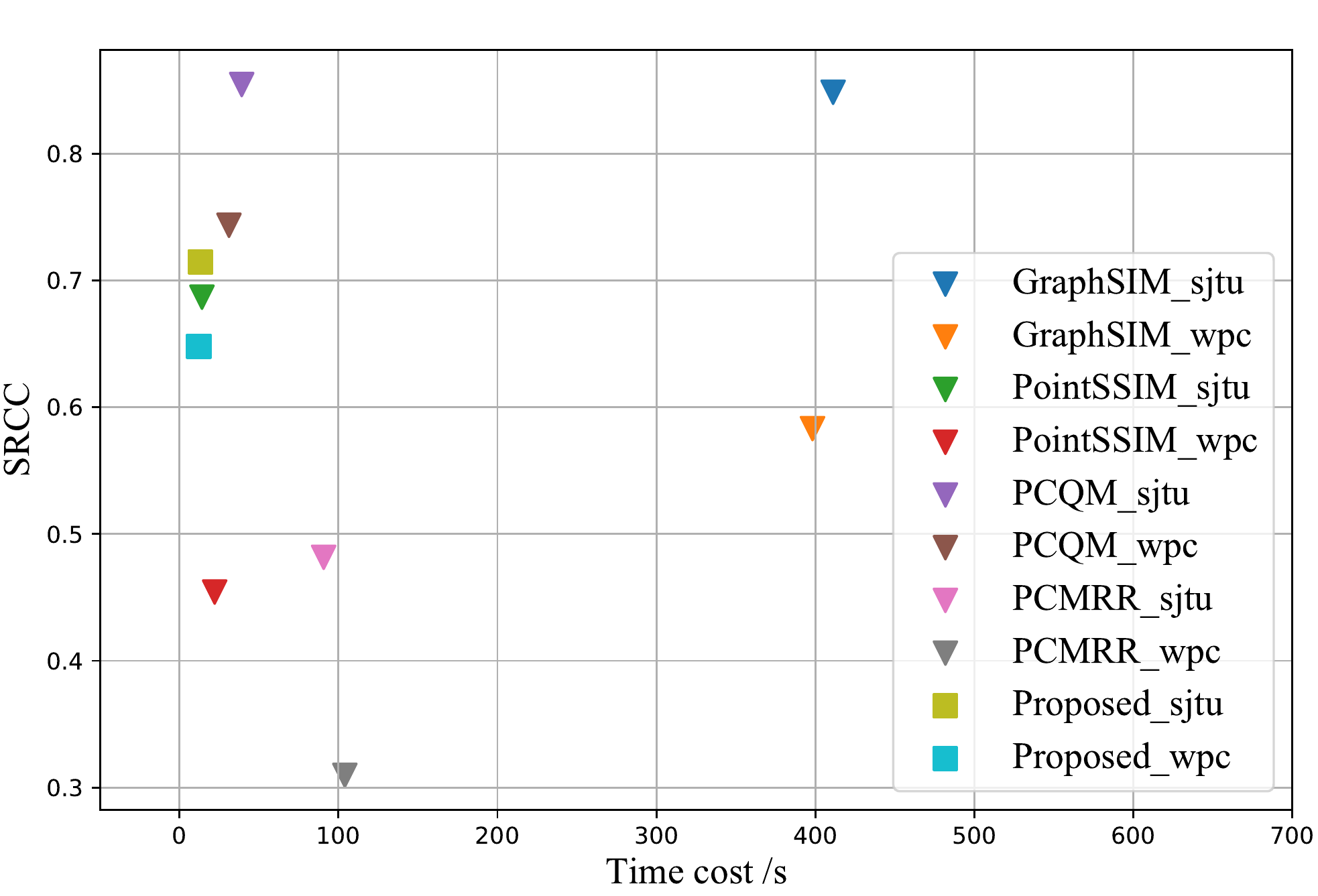}
    \caption{The results of time cost comparison on the SJTU-PCQA and the WPC databases. The time cost refers to the average time consumption per point cloud for each database.}
    \label{fig:time}
    \vspace{-0.3cm}
\end{figure}

\subsection{Computational Efficiency}
Considering that the proposed method operates directly from the 3D model, we also focus on computational efficiency. The image-based metrics usually operate quite fast due to the mature development of IQA metrics. Therefore, to make the comparison meaningful, we select model-based methods GraphSIM, PointSSIM, PCQM, and PCMRR as competitors. We conduct the test on a computer with Intel (R) Core (TM) i5-3470 CPU @ 3.20 GHz and 8 GB RAM on the Windows platform. The corresponding time cost results are shown in Fig. \ref{fig:time}, and we can see clearly that the average time cost of the proposed method is smaller than all the compared model-based methods, which indicates that our method achieves relatively considerable computational efficiency.

\section{Conclusion}
This paper proposes a no-reference colored 3D model quality assessment metric based on entropy and 3D natural scene statistics. The proposed method deals with the quality assessment problems for both colored point cloud and mesh models. We first project the 3D models into corresponding quality-related geometry and color feature domains. Then entropy and various 3D-NSS parameters are estimated to better capture the representative characteristics and quantify the distortions that are more in line with human perception. The proposed method is validated on the colored point cloud quality assessment database (SJTU-PCQA), the Waterloo point cloud assessment database (WPC), and the colored mesh quality assessment database (CMDM). The experimental results show that our method outperforms most compared NR 3D-QA metrics with competitive computational resources and reduces the performance gap with the state-of-the-art FR 3D-QA metrics. The proposed method follows a common NR framework and can be easily modified and expanded to satisfy specific needs, which has great application potential. The code of the proposed method is released for promoting the development of NR 3D-QA.
\label{sec:conclusion}

 




\bibliographystyle{IEEEtran}
\bibliography{output}

\begin{thebibliography}{10}
\providecommand{\url}[1]{#1}
\csname url@samestyle\endcsname
\providecommand{\newblock}{\relax}
\providecommand{\bibinfo}[2]{#2}
\providecommand{\BIBentrySTDinterwordspacing}{\spaceskip=0pt\relax}
\providecommand{\BIBentryALTinterwordstretchfactor}{4}
\providecommand{\BIBentryALTinterwordspacing}{\spaceskip=\fontdimen2\font plus
\BIBentryALTinterwordstretchfactor\fontdimen3\font minus
  \fontdimen4\font\relax}
\providecommand{\BIBforeignlanguage}[2]{{%
\expandafter\ifx\csname l@#1\endcsname\relax
\typeout{** WARNING: IEEEtran.bst: No hyphenation pattern has been}%
\typeout{** loaded for the language `#1'. Using the pattern for}%
\typeout{** the default language instead.}%
\else
\language=\csname l@#1\endcsname
\fi
#2}}
\providecommand{\BIBdecl}{\relax}
\BIBdecl

\bibitem{zhang2021}
Z.~Zhang, W.~Sun, X.~Min, T.~Wang, W.~Lu, W.~Zhu, and G.~Zhai, ``A no-reference
  visual quality metric for 3d color meshes,'' in \emph{2021 IEEE International
  Conference on Multimedia Expo Workshops (ICMEW)}, 2021, pp. 1--6.

\bibitem{application}
H.~{Graf}, S.~P. {Serna}, and A.~{Stork}, ``Adaptive quality meshing for
  "on-the-fly" volumetric mesh manipulations within virtual environments,'' in
  \emph{IEEE Symposium on Virtual Environments, Human-Computer Interfaces and
  Measurement Systems}, 2006, pp. 178--183.

\bibitem{li2020occupancy}
L.~Li, Z.~Li, S.~Liu, and H.~Li, ``Occupancy-map-based rate distortion
  optimization and partition for video-based point cloud compression,''
  \emph{IEEE Transactions on Circuits and Systems for Video Technology},
  vol.~31, no.~1, pp. 326--338, 2020.

\bibitem{mekuria2016design}
R.~Mekuria, K.~Blom, and P.~Cesar, ``Design, implementation, and evaluation of
  a point cloud codec for tele-immersive video,'' \emph{IEEE Transactions on
  Circuits and Systems for Video Technology}, vol.~27, no.~4, pp. 828--842,
  2016.

\bibitem{apple}
\BIBentryALTinterwordspacing
``Apple developer document: Overview of scanning and detecting 3-dimension
  objects,'' May. 1, 2022. [Online]. Available:
  \url{https://wrywhisker.pulpfriction.net/wallcrust/linear-colinear-felinear.html}
\BIBentrySTDinterwordspacing

\bibitem{intel}
\BIBentryALTinterwordspacing
``Intel realsense,'' May. 1, 2022. [Online]. Available:
  \url{https://www.intelrealsense.com/}
\BIBentrySTDinterwordspacing

\bibitem{sjtu-pcqa}
Q.~Yang, H.~Chen, Z.~Ma, Y.~Xu, R.~Tang, and J.~Sun, ``Predicting the
  perceptual quality of point cloud: A 3d-to-2d projection-based exploration,''
  \emph{IEEE Transactions on Multimedia}, pp. 1--1, 2020.

\bibitem{database}
Y.~{Nehmé}, F.~{Dupont}, J.~P. {Farrugia}, P.~{Le Callet}, and G.~{Lavoué},
  ``Visual quality of 3d meshes with diffuse colors in virtual reality:
  Subjective and objective evaluation,'' \emph{IEEE Transactions on
  Visualization and Computer Graphics}, vol.~27, no.~3, pp. 2202--2219, 2021.

\bibitem{pcqa_database1}
E.~Alexiou, I.~Viola, T.~M. Borges, T.~A. Fonseca, R.~L. de~Queiroz, and
  T.~Ebrahimi, ``A comprehensive study of the rate-distortion performance in
  mpeg point cloud compression,'' \emph{APSIPA Transactions on Signal and
  Information Processing}, vol.~8, p. e27, 2019.

\bibitem{pcqa_database2}
E.~M. Torlig, E.~Alexiou, T.~A. Fonseca, R.~L. de~Queiroz, and T.~Ebrahimi,
  ``{A novel methodology for quality assessment of voxelized point clouds},''
  in \emph{Applications of Digital Image Processing XLI}, vol. 10752.\hskip 1em
  plus 0.5em minus 0.4em\relax International Society for Optics and Photonics,
  2018, pp. 174 -- 190.

\bibitem{su2021perceptual}
H.~Su, Q.~Liu, Z.~Duanmu, W.~Liu, and Z.~Wang, ``Perceptual quality assessment
  of colored 3d point clouds,'' \emph{arXiv preprint arXiv:2111.05474}, 2021.

\bibitem{sun2021blind}
W.~Sun, X.~Min, G.~Zhai, and S.~Ma, ``Blind quality assessment for in-the-wild
  images via hierarchical feature fusion and iterative mixed database
  training,'' 2021.

\bibitem{p2point}
P.~Cignoni, C.~Rocchini, and R.~Scopigno, ``Metro: Measuring error on
  simplified surfaces,'' \emph{Computer Graphics Forum}, vol.~17, no.~2, pp.
  167--174, 1998.

\bibitem{p2plane}
R.~Mekuria and P.~Cesar, ``Mp3dg-pcc, open source software framework for
  implementation and evaluation of point cloud compression.''\hskip 1em plus
  0.5em minus 0.4em\relax Association for Computing Machinery, 2016, p.
  1222–1226.

\bibitem{m1}
G.~Lavoué, ``A multiscale metric for 3d mesh visual quality assessment,''
  \emph{Computer Graphics Forum}, vol.~30, no.~5, pp. 1427--1437, 2011.

\bibitem{ff2_roughness}
K.~Wang, F.~Torkhani, and A.~Montanvert, ``A fast roughness-based approach to
  the assessment of 3d mesh visual quality,'' \emph{Computers and Graphics},
  vol.~36, no.~7, pp. 808--818, 2012.

\bibitem{p2mesh}
D.~Tian, H.~Ochimizu, C.~Feng, R.~Cohen, and A.~Vetro, ``Geometric distortion
  metrics for point cloud compression,'' in \emph{IEEE International Conference
  on Image Processing}, 2017, pp. 3460--3464.

\bibitem{angular}
E.~Alexiou and T.~Ebrahimi, ``Point cloud quality assessment metric based on
  angular similarity,'' in \emph{IEEE International Conference on Multimedia
  and Expo}, 2018, pp. 1--6.

\bibitem{pcqa2}
A.~Javaheri, C.~Brites, F.~Pereira, and J.~Ascenso, ``A generalized hausdorff
  distance based quality metric for point cloud geometry,'' in
  \emph{International Conference on Quality of Multimedia Experience}, 2020,
  pp. 1--6.

\bibitem{dame}
L.~Váša and J.~Rus, ``Dihedral angle mesh error: a fast perception correlated
  distortion measure for fixed connectivity triangle meshes,'' \emph{Computer
  Graphics Forum}, vol.~31, no.~5, pp. 1715--1724, 2012.

\bibitem{pcqa3}
E.~Alexiou and T.~Ebrahimi, ``Towards a point cloud structural similarity
  metric,'' in \emph{IEEE International Conference on Multimedia Expo
  Workshops}, 2020, pp. 1--6.

\bibitem{tian-color}
D.~Tian and G.~AlRegib, ``Batex3: Bit allocation for progressive transmission
  of textured 3-d models,'' \emph{IEEE Transactions on Circuits and Systems for
  Video Technology}, vol.~18, no.~1, pp. 23--35, 2008.

\bibitem{guo-color}
J.~Guo, V.~Vidal, I.~Cheng, A.~Basu, A.~Baskurt, and G.~Lavoue, ``Subjective
  and objective visual quality assessment of textured 3d meshes,'' \emph{ACM
  Transactions on Applied Perception (TAP)}, vol.~14, no.~2, 2016.

\bibitem{pcqm}
G.~Meynet, Y.~Nehmé, J.~Digne, and G.~Lavoué, ``Pcqm: A full-reference
  quality metric for colored 3d point clouds,'' in \emph{International
  Conference on Quality of Multimedia Experience}, 2020, pp. 1--6.

\bibitem{liu2021reduced}
Q.~Liu, H.~Yuan, R.~Hamzaoui, H.~Su, J.~Hou, and H.~Yang, ``Reduced reference
  perceptual quality model with application to rate control for video-based
  point cloud compression,'' \emph{IEEE Transactions on Image Processing},
  vol.~30, pp. 6623--6636, 2021.

\bibitem{ssim}
Z.~Wang, A.~Bovik, H.~Sheikh, and E.~Simoncelli, ``Image quality assessment:
  from error visibility to structural similarity,'' \emph{IEEE Transactions on
  Image Processing}, vol.~13, no.~4, pp. 600--612, 2004.

\bibitem{pcqa-large-scale}
Y.~Liu, Q.~Yang, Y.~Xu, and L.~Yang, ``Point cloud quality assessment:
  Large-scale dataset construction and learning-based no-reference approach,''
  2020.

\bibitem{Qi_2017_CVPR}
C.~R. Qi, H.~Su, K.~Mo, and L.~J. Guibas, ``Pointnet: Deep learning on point
  sets for 3d classification and segmentation,'' in \emph{Proceedings of the
  IEEE Conference on Computer Vision and Pattern Recognition}, July 2017.

\bibitem{liu2021pqa}
Q.~Liu, H.~Yuan, H.~Su, H.~Liu, Y.~Wang, H.~Yang, and J.~Hou, ``Pqa-net: Deep
  no reference point cloud quality assessment via multi-view projection,''
  \emph{IEEE Transactions on Circuits and Systems for Video Technology}, 2021.

\bibitem{nr-svr}
I.~Abouelaziz, M.~El~Hassouni, and H.~Cherifi, ``No-reference 3d mesh quality
  assessment based on dihedral angles model and support vector regression,'' in
  \emph{Image and Signal Processing}, 2016, pp. 369--377.

\bibitem{nr-cnn}
I.~{Abouelaziz}, M.~E. {Hassouni}, and H.~{Cherifi}, ``A convolutional neural
  network framework for blind mesh visual quality assessment,'' in \emph{IEEE
  International Conference on Image Processing}, 2017, pp. 755--759.

\bibitem{nr-cnncmp}
I.~Abouelaziz, A.~Chetouani, M.~{El Hassouni}, L.~J. Latecki, and H.~Cherifi,
  ``No-reference mesh visual quality assessment via ensemble of convolutional
  neural networks and compact multi-linear pooling,'' \emph{Pattern
  Recognition}, vol. 100, p. 107174, 2020.

\bibitem{2d1}
S.~Yang, C.-H. Lee, and C.-C.~J. Kuo, ``Optimized mesh and texture multiplexing
  for progressive textured model transmission,'' in \emph{Proceedings of the
  12th annual ACM international conference on Multimedia}, 2004, p. 676–683.

\bibitem{2d2}
F.~Caillaud, V.~Vidal, F.~Dupont, and G.~Lavoué, ``Progressive compression of
  arbitrary textured meshes,'' \emph{Computer Graphics Forum}, vol.~35, no.~7,
  pp. 475--484, 2016.

\bibitem{ms-ssim}
Z.~Wang, E.~Simoncelli, and A.~Bovik, ``Multiscale structural similarity for
  image quality assessment,'' in \emph{Asilomar Conference on Signals, Systems
  Computers}, vol.~2, 2003, pp. 1398--1402 Vol.2.

\bibitem{fsim}
L.~Zhang, L.~Zhang, X.~Mou, and D.~Zhang, ``Fsim: A feature similarity index
  for image quality assessment,'' \emph{IEEE Transactions on Image Processing},
  vol.~20, no.~8, pp. 2378--2386, 2011.

\bibitem{vif}
H.~Sheikh and A.~Bovik, ``Image information and visual quality,'' \emph{IEEE
  Transactions on Image Processing}, vol.~15, no.~2, pp. 430--444, 2006.

\bibitem{brisque}
A.~Mittal, A.~K. Moorthy, and A.~C. Bovik, ``No-reference image quality
  assessment in the spatial domain,'' \emph{IEEE Transactions on Image
  Processing}, vol.~21, no.~12, pp. 4695--4708, 2012.

\bibitem{nss1}
E.~M. Torlig, E.~Alexiou, T.~A. Fonseca, R.~L. de~Queiroz, and T.~Ebrahimi,
  ``{A novel methodology for quality assessment of voxelized point clouds},''
  in \emph{Applications of Digital Image Processing XLI}, vol. 10752.\hskip 1em
  plus 0.5em minus 0.4em\relax International Society for Optics and Photonics,
  pp. 174 -- 190.

\bibitem{nss2}
A.~K. Moorthy and A.~C. Bovik, ``Blind image quality assessment: From natural
  scene statistics to perceptual quality,'' \emph{IEEE Transactions on Image
  Processing}, vol.~20, no.~12, pp. 3350--3364, 2011.

\bibitem{lin2019blind}
Y.~Lin, M.~Yu, K.~Chen, G.~Jiang, Z.~Peng, and F.~Chen, ``Blind mesh quality
  assessment method based on concave, convex and structural features
  analyses,'' in \emph{2019 IEEE International Conference on Multimedia \& Expo
  Workshops (ICMEW)}.\hskip 1em plus 0.5em minus 0.4em\relax IEEE, 2019, pp.
  282--287.

\bibitem{abouelaziz2018blind}
I.~Abouelaziz, M.~El~Hassouni, and H.~Cherifi, ``Blind 3d mesh visual quality
  assessment using support vector regression,'' \emph{Multimedia Tools and
  Applications}, vol.~77, no.~18, pp. 24\,365--24\,386, 2018.

\bibitem{pcqa-curvature2}
Q.~Mérigot, M.~Ovsjanikov, and L.~J. Guibas, ``Voronoi-based curvature and
  feature estimation from point clouds,'' \emph{IEEE Transactions on
  Visualization and Computer Graphics}, vol.~17, no.~6, pp. 743--756, 2011.

\bibitem{pc-eigenvalue1}
H.~Thomas, J.~Deschaud, B.~Marcotegui, F.~Goulette, and Y.~L. Gall, ``Semantic
  classification of 3d point clouds with multiscale spherical neighborhoods,''
  \emph{CoRR}, vol. abs/1808.00495, 2018.

\bibitem{pc-eigenvalue2}
T.~Hackel, J.~D. Wegner, and K.~Schindler, ``Fast semantic segmentation of 3d
  point clouds with strongly varying density,'' \emph{ISPRS annals of the
  photogrammetry, remote sensing and spatial information sciences}, vol.~3, pp.
  177--184, 2016.

\bibitem{gaussian_curvature}
T.~{Surazhsky}, E.~{Magid}, O.~{Soldea}, G.~{Elber}, and E.~{Rivlin}, ``A
  comparison of gaussian and mean curvatures estimation methods on triangular
  meshes,'' in \emph{IEEE International Conference on Robotics and Automation},
  vol.~1, 2003, pp. 1021--1026 vol.1.

\bibitem{average_curvature}
P.~Alliez, D.~Cohen-Steiner, O.~Devillers, B.~L{\'e}vy, and M.~Desbrun,
  ``Anisotropic polygonal remeshing,'' in \emph{ACM SIGGRAPH 2003 Papers},
  2003, pp. 485--493.

\bibitem{dihedral2}
M.~{Corsini}, E.~D. {Gelasca}, T.~{Ebrahimi}, and M.~{Barni}, ``Watermarked 3-d
  mesh quality assessment,'' \emph{IEEE Transactions on Multimedia}, vol.~9,
  no.~2, pp. 247--256, 2007.

\bibitem{angle1}
N.~Mukherjee, ``A hybrid, variational 3d smoother for orphaned shell meshes,''
  in \emph{IMR}, 2002.

\bibitem{angle2}
H.~Lee, P.~Alliez, and M.~Desbrun, ``Angle-analyzer: A triangle-quad mesh
  codec,'' \emph{Computer Graphics Forum}, vol.~21, no.~3, 2002.

\bibitem{sharifi1995estimation}
K.~Sharifi and A.~Leon-Garcia, ``Estimation of shape parameter for generalized
  gaussian distributions in subband decompositions of video,'' \emph{IEEE
  Transactions on Circuits and Systems for Video Technology}, vol.~5, no.~1,
  pp. 52--56, 1995.

\bibitem{sklearn}
\BIBentryALTinterwordspacing
``Scikit-learn for python,'' May. 1, 2022. [Online]. Available:
  \url{https://scikit-learn.org/stable/}
\BIBentrySTDinterwordspacing

\bibitem{niqe}
A.~{Mittal}, R.~{Soundararajan}, and A.~C. {Bovik}, ``Making a “completely
  blind” image quality analyzer,'' \emph{IEEE Signal Processing Letters},
  vol.~20, no.~3, pp. 209--212, 2013.

\bibitem{yang2020inferring}
Q.~Yang, Z.~Ma, Y.~Xu, Z.~Li, and J.~Sun, ``Inferring point cloud quality via
  graph similarity,'' \emph{IEEE Transactions on Pattern Analysis and Machine
  Intelligence}, 2020.

\bibitem{viola2020reduced}
I.~Viola and P.~Cesar, ``A reduced reference metric for visual quality
  evaluation of point cloud contents,'' \emph{IEEE Signal Processing Letters},
  vol.~27, pp. 1660--1664, 2020.

\bibitem{nr-grnn}
I.~{Abouelaziz}, M.~{El Hassouni}, and H.~{Cherifi}, ``A curvature based method
  for blind mesh visual quality assessment using a general regression neural
  network,'' in \emph{International Conference on Signal-Image Technology and
  Internet-Based Systems}, 2016, pp. 793--797.

\bibitem{statistic-test}
H.~Sheikh, M.~Sabir, and A.~Bovik, ``A statistical evaluation of recent full
  reference image quality assessment algorithms,'' \emph{IEEE Transactions on
  Image Processing}, vol.~15, no.~11, pp. 3440--3451, 2006.

\end{thebibliography}
\end{document}